\documentclass{article} 
\usepackage[preprint]{colm2026_conference}
\usepackage{multirow}
\usepackage[most,skins,theorems]{tcolorbox}
\usepackage{microtype}
\usepackage{hyperref}
\usepackage{url}
\usepackage{booktabs}
\usepackage{minitoc}
\mtcsetdepth{parttoc}{2}
\usepackage{amsmath}
\usepackage{amssymb}
\usepackage{mathtools}
\usepackage{amsthm}
\usepackage{pifont}
\usepackage{xcolor}
\usepackage[table]{xcolor}

\definecolor{HeatColor}{HTML}{2ECC71}
\usepackage{microtype}
\usepackage{graphicx}
\usepackage{subcaption}
\usepackage{booktabs} 
\usepackage{hyperref}
\usepackage{url}
\usepackage{tabularx}
\usepackage{widetable}
\usepackage[dvipsnames,table]{xcolor}
\usepackage{tikz}
\usetikzlibrary{shadows}
\usepackage{enumitem}
\usepackage{multirow}
\tcbuselibrary{breakable}
\usepackage{siunitx}
\tcbset{
  aibox/.style={
    width=\linewidth,
    top=8pt,
    bottom=4pt,
    colback=blue!6!white,
    colframe=black,
    colbacktitle=black,
    enhanced,
    center,
    breakable,
    pad at break=4mm,  
    attach boxed title to top left={yshift=-0.1in,xshift=0.15in},
    boxed title style={boxrule=0pt,colframe=white,},
  }
}

\newtcolorbox{AIbox}[2][]{aibox,title=#2,#1}
\usepackage{pifont}
\usepackage{color, colortbl}
\usepackage[boxed]{algorithm2e}
\usepackage[fixed]{fontawesome5}
\usepackage[most]{tcolorbox}
\definecolor{promptbgcolor}{RGB}{245,245,245} 
\definecolor{promptframecolor}{RGB}{100,100,100} 

\usepackage[most]{tcolorbox}
\definecolor{promptbgcolor}{RGB}{245,245,245} 
\definecolor{promptframecolor}{RGB}{100,100,100}
\usepackage{etoc}
\etocsetlevel{xpart}{6}
\etocsetlevel{xchapter}{6}
\etocsetlevel{xsect}{6}

\newtcolorbox{systemprompt}[1][]{
  enhanced,
  breakable,
  title={System Prompt},
  colback=promptbgcolor,
  colframe=promptframecolor,
  coltitle=white,
  fonttitle=\bfseries\sffamily,
  fontupper=\small\ttfamily,
  boxrule=0.8pt,
  arc=3pt,
  left=6pt, right=6pt, top=6pt, bottom=6pt,      
  bottomrule at break=0pt,   
  toprule at break=0pt,      
  pad at break*=0pt,         
  #1
}

\usepackage{lineno}

\definecolor{darkblue}{rgb}{0, 0, 0.5}
\hypersetup{colorlinks=true, citecolor=darkblue, linkcolor=darkblue, urlcolor=darkblue}

\title{\vspace{-0.5cm}Round-Trip Translation Reveals What Frontier Multilingual Benchmarks Miss}

\author{Ronald Skorobogat \hspace{0.5em} Ameya Prabhu$^\dagger$ \hspace{0.5em}  Matthias Bethge$^\dagger$ \\
\vspace{0.2cm}Tübingen AI Center, University of Tübingen 
}

\vspace{-0.6cm}

\begin{document}

\ifcolmsubmission
\linenumbers
\fi

\maketitle
\vspace{-1cm}
\begin{center}
    \begin{tabular}{c@{\hskip 19pt}c}
    \raisebox{-1pt}{\faGlobe} \href{https://bethgelab.github.io/lit-benchmark/}{\texttt{Leaderboard}}
    \hspace{1.0cm}
    \raisebox{-1pt}{\faGithub} \href{https://github.com/bethgelab/lit-benchmark}{\fontsize{8.8pt}{0pt}\texttt{Code}}
    \hspace{1.0cm}
    \raisebox{-1.5pt}{\faDatabase}\href{https://huggingface.co/datasets/bethgelab/lit-benchmark/}{\fontsize{8.8pt}{0pt} \texttt{Dataset}} \\
\end{tabular}
\end{center}
\vspace{0.3cm}
\begin{abstract}
Multilingual benchmarks guide the development of frontier models. Yet multilingual evaluations reported by frontier models are structured similar to popular reasoning and knowledge benchmarks, but across many languages. We show such benchmarks, and consequently multilingual evaluations, measure mathematical reasoning and factual recall, not multilingual proficiency. For example, thinking variants dramatically outperform instruct variants on these benchmarks, yet often perform worse on real-world multilingual tasks, such as LMArena. We propose a simple alternative: evaluate multilingual capability via round-trip translation. Given text in a source language, translate it to a target language and back; semantic gaps between the original and result expose failures in multilingual \textit{generation} capabilities. Round-trip translation correlates almost perfectly ($\rho=0.94$) with user ratings on LMArena with our benchmark, requires no human reference translations, and does not require a more capable multilingual judge than tested models. Lastly, we introduce Lost in Translation (LiT), a challenging round-trip translation benchmark spanning widely spoken languages worldwide, for realistic evaluation of multilingual frontier models.
\end{abstract}
\vspace{-0.2cm}
\vspace{-0.35cm}
\section{Introduction}
\vspace{-0.15cm}

Multilingual benchmarks \citep{son-etal-2025-linguistic, wang2025polymath, romanou2025include, singh-etal-2025-global} shape how frontier models are built. Developers use these evaluations to measure progress, allocate resources, and claim capabilities across languages. Yet a fundamental question remains open: Does progress on current multilingual benchmarks truly reflect progress in multilingual proficiency?

\begin{figure*}[t] 
     \centering
     \vspace*{-0.6cm}
     \begin{subfigure}[b]{0.325\textwidth}
         \centering
         \includegraphics[width=\textwidth]{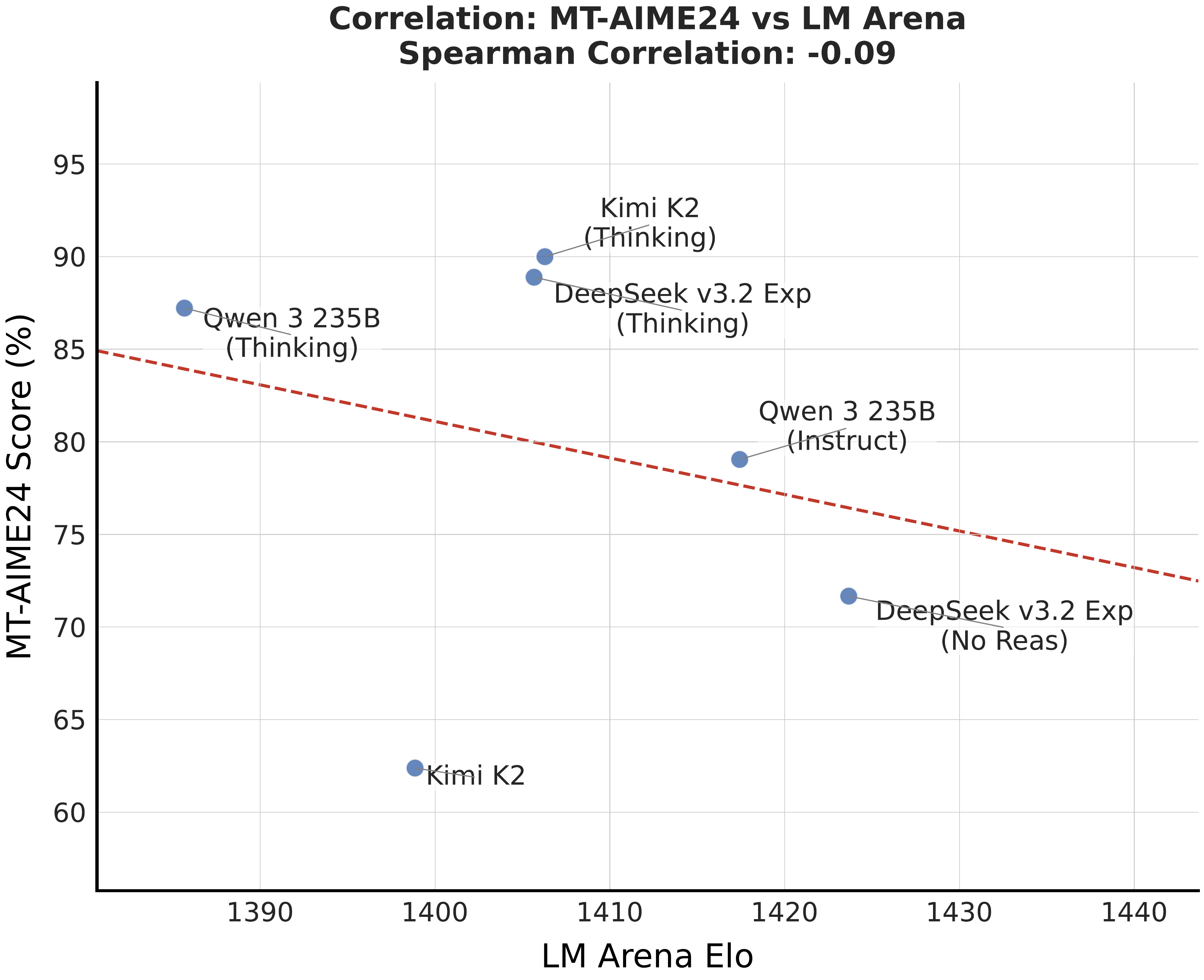}
         \caption{MT-AIME24 vs. LMArena}
         \label{fig:first}
     \end{subfigure}
     \hfill 
     \begin{subfigure}[b]{0.325\textwidth}
         \centering
         \includegraphics[width=\textwidth]{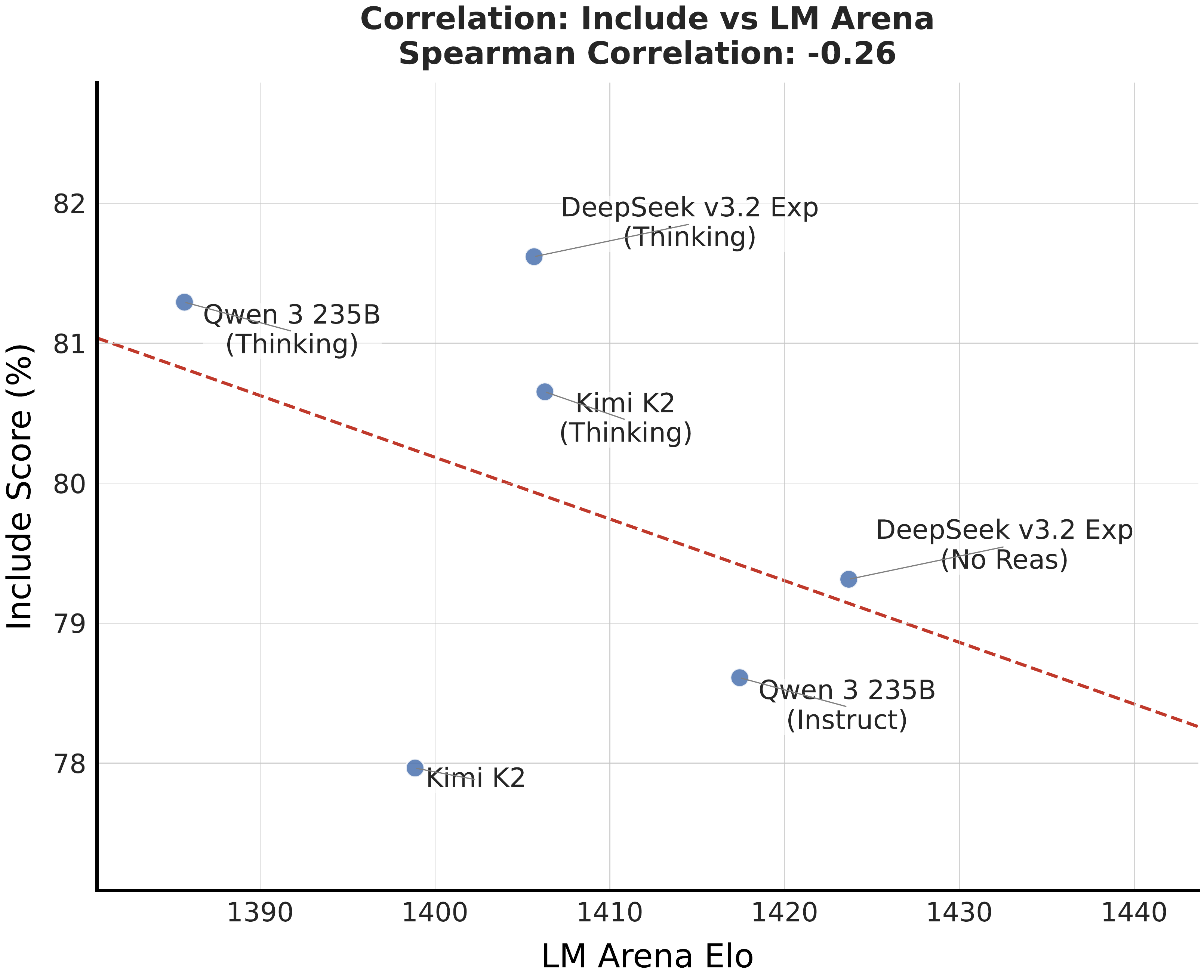}
         \caption{Include vs. LMArena}
         \label{fig:second}
     \end{subfigure}
     \hfill
     \begin{subfigure}[b]{0.325\textwidth}
         \centering
         \includegraphics[width=\textwidth]{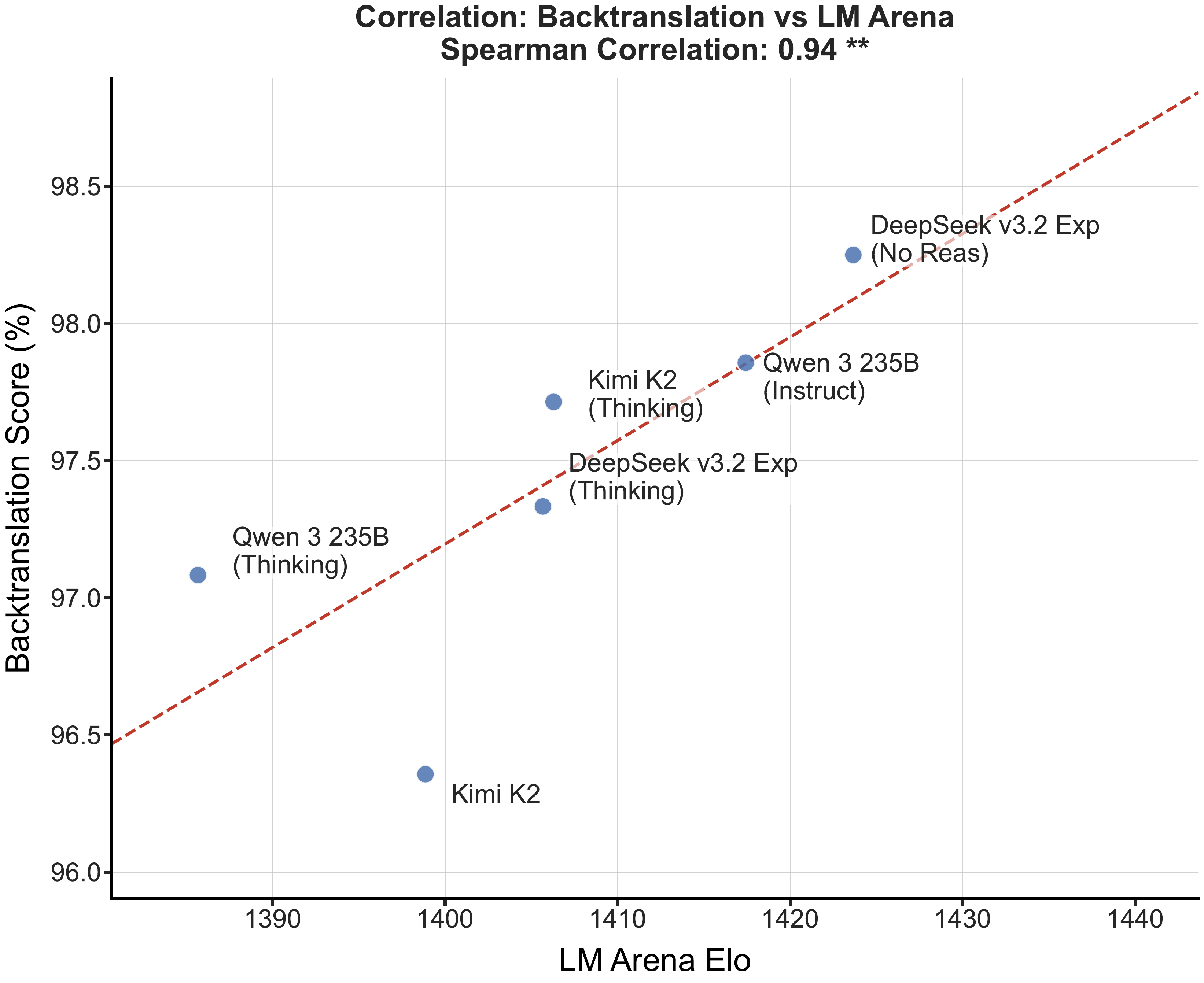}
         \caption{LiT vs. LMArena}
         \label{fig:third}
     \end{subfigure}
     
     \caption{\textbf{Multilingual benchmarks correlate poorly with human preferences.} We show benchmark scores against LMArena Elo ratings for six frontier open-source models, each in Thinking and Non-Thinking variants. (a) MT-AIME24 \citep{son-etal-2025-linguistic} shows near-zero correlation ($\rho=-0.09$), with thinking variants dramatically outperforming on the benchmark but perform no better on LMArena; (b) INCLUDE \citep{romanou2025include} shows moderate negative correlation ($\rho=-0.26$), with similar Thinking vs. Non-Thinking disconnect; (c) Round trip translation using LiT with percentage of MQM scores at least 80 correlates almost perfectly with LMArena ($\rho=0.94$, with one-sided permutation test p=0.008), with minimal gap between thinking and instruct variants of a model.}
     \label{fig:corr_plot}
     \vspace*{-0.5cm}
\end{figure*}

We find it does not. The evaluation of frontier multilingual models is currently dominated by two predominant paradigms: mathematical reasoning tasks, such as MT-AIME24 \citep{son-etal-2025-linguistic} and PolyMath~\citep{wang2025polymath}, and general knowledge multiple-choice question answering (MCQA), such as INCLUDE \citep{romanou2025include} and Global-MMLU \citep{singh-etal-2025-global}. We discover, as shown in Figure \ref{fig:corr_plot} and \ref{fig:new_last}, that performance gaps on such benchmarks primarily reflect differences in mathematical problem-solving ($\rho=0.94$) or factual recall ($\rho=0.83$) respectively -- not multilingual generation capability ($\rho=-0.09$ and $-0.26$ respectively). Intuitively, the original AIME24 and MMLU, similarly, are poor benchmarks for faithful measurement of English comprehension. Consequently, performance gaps between two models on frontier multilingual benchmarks like MT-AIME24 and INCLUDE highlights differences in their mathematical reasoning and factual recall, not their language proficiency. 
Previous works \citep{wu2025bitter} provide support on the failure of multilingual benchmarks to align with human preferences. Overall, we show that frontier multilingual benchmarks are not a faithful measurement of multilingual capabilities. 

\definecolor{goodgreen}{RGB}{34, 139, 34}  
\definecolor{badred}{RGB}{220, 20, 60}     

\definecolor{lightgray}{gray}{0.95}             
\definecolor{highlightblue}{RGB}{180, 220, 255} 

\newcommand{\cmark}{\textcolor{goodgreen}{\ding{51}}} 
\newcommand{\xmark}{\textcolor{badred}{\ding{55}}}    

\begin{table*}[t]
    \centering
    \small
    \vspace*{-0.5cm}
    \setlength{\tabcolsep}{5pt}
    \caption{\textbf{Benchmark comparison across six evaluation criteria.} We compare nine multilingual benchmarks across six evaluation criteria including: contamination-Free, challenging (not saturated yet, difficult for frontier models), linguistic diversity (covers diverse language families and scripts), NLG (tests natural language generation), Efficiency (low computational cost), and ground truth-free (without per-sample human annotation).}
    \vspace*{-0.2cm}
    \resizebox{\linewidth}{!}{
    \begin{tabular}{l|cccccc} 
        \toprule
        \textbf{Benchmark} & 
        \textbf{Contamination-Free} & 
        \textbf{Challenging} &  
        \textbf{Linguistic Diversity} & 
        \textbf{NLG} & 
        \textbf{Efficiency} & 
        \textbf{Ground Truth-Free} \\
        
        \midrule

        \rowcolor{lightgray} \multicolumn{7}{l}{\textit{General Knowledge Multiple-Choice}} \\
        Include       & \cmark & \cmark & \cmark & \xmark & \xmark & \xmark \\
        MMMLU         & \xmark & \xmark & \xmark & \xmark & \xmark & \xmark \\
        Global MMLU   & \xmark & \xmark & \xmark & \xmark & \xmark & \xmark \\
        \midrule

        \rowcolor{lightgray} \multicolumn{7}{l}{\textit{Mathematical Reasoning}} \\
        MT-AIME24     & \xmark & \cmark & \xmark & \cmark & \xmark & \xmark \\
        MGSM          & \xmark & \xmark & \xmark & \cmark & \cmark & \xmark \\
        PolyMath      & \cmark & \cmark & \xmark & \cmark & \xmark & \xmark \\
        \midrule

        \rowcolor{lightgray} \multicolumn{7}{l}{\textit{Machine Translation}} \\
        FLORES-200    & \xmark & \xmark & \cmark & \cmark & \cmark & \xmark \\
        WMT24++         & \cmark & \xmark & \cmark & \cmark & \xmark & \xmark \\
        \midrule

        \rowcolor{highlightblue} \textbf{LiT (Ours)} & \cmark & \cmark & \cmark & \cmark & \cmark & \cmark \\
        \bottomrule
        
    \end{tabular}}
    \vspace{-0.4cm}
    \label{tab:benchmark_tiers}
\end{table*}

We propose a simple solution: evaluate multilingual capability through round-trip translation \citep{brislin1970back, sennrich2016improving}. Given a passage in a source language, round-trip translation involves translating it to a target (sequence of) language(s) and back -- comparing the result to the original passage. Semantic gaps between the original and back-translated text show whether a model can preserve meaning across languages, a critical capability multilingual benchmarks should capture \citep{wu2025bitter}. Unlike MT-AIME or Global-MMLU, multilingual understanding and generation is the challenging task in a round-trip translation benchmark.

We describe its comparative advantages in Table \ref{tab:benchmark_tiers} when compared to current machine translation and current multilingual benchmarks. Round-trip translation has two advantages over classical translation evaluation: scalable sample creation -- it requires no human-written reference translations; and ease of judging -- the judge needs to evaluate two English paragraphs, not compare them in the low-resource language. This implies we do not need a stronger model to evaluate multilingual translation quality, which does not exist when testing frontier models. Current translation benchmarks are too easy for evaluating frontier models. Assuming frontier models keep improving, we argue these advantages --  ease of creating new samples and judging -- should increasingly favor round-trip translation over classical machine translation.

To enable systematic evaluation, we introduce Lost in Translation (LiT), a round-trip translation benchmark of 1600 samples across 8 language sequences spanning high-, medium-, and low-resource languages. LiT covers technical \citep{taguchi-etal-2025-languages}, pragmatic \citep{park-etal-2024-multiprageval}, and informal language \citep{yao-etal-2024-benchmarking} using MQM-based automated judging. We show that round-trip translation using LiT offers two advantages over current benchmarks. First, we show that round-trip translation correlates almost perfectly ($\rho=0.94$) with user ratings on LMArena \citep{chiang2024chatbot}\footnote{LMArena is considered an expensive but gold standard set, as it collects vast array of continuously updated real-world user queries and uses a wisdom of the crowd effect (see \citet{ni2024mixeval} for details).}. Second, it exposes failure modes that reasoning and multiple-choice benchmarks miss. We document systematic hallucinations, content omissions, and semantic drift in multilingual \textit{generation} using MQM scores \citep{lommel2014multidimensional, kocmi-federmann-2023-gemba}, not faithfully captured by current multiple-choice (MCQ) or mathematical reasoning evaluations.

Our results reveal a stark capability gap: open-source frontier models achieve above 88 MQM scores on high-resource sequences but collapse to below 50 on low-resource languages, indicating unusable output\footnote{Translations achieving an MQM score above 80 are generally considered faithful and fit-for-purpose, while a score of $<$80 should trigger mandatory human review \citep{lommel2014multidimensional, lommel-etal-2024-multi}.}. Furthermore, we test model performance on challenging constructions for round-trip translation \citep{somers2005round, zhuo-etal-2023-rethinking}. Model rankings on these challenging cases match general performance closely, validating the robustness of the LiT benchmark.  When instructed to translate faithfully, frontier models prioritize instruction-following over fluency correction. Rather than self-correcting or masking intermediate errors, they reliably reproduce the awkward phrasing. Consequently, this demonstrates that round-trip evaluation is not artificially derailed by complex linguistic phenomena, but instead serves as a stable, highly correlated proxy for genuine cross-lingual generation.

Overall, we hope this work shifts how multilingual frontier models are evaluated. We argue for methods that directly measure user preference \citep{wu2025bitter}: genuine cross-lingual generation competence. We hope the LiT benchmark contributes to guiding the development of models that actually work for the billions of people who do not speak English as their first language.

\section{Related Works}
\label{sec:related}
\textbf{Multilingual Benchmarks.} The standard approach to multilingual LLM evaluation translates English reasoning and knowledge tasks into other languages \citep{hendrycks2021measuring, son-etal-2025-linguistic}. 
However, such translated benchmarks often correlate poorly with human evaluations, performing significantly worse than natively localized benchmarks~\cite{wu2025bitter}. Alternatives like INCLUDE~\cite{romanou2025include} and Global-MMLU~\cite{singh-etal-2025-global} address this by incorporating native content or cultural context, mitigating both ``translationese'' artifacts and Western-centric bias~\cite{wu2025bitter}. However, these localized benchmarks inherit a deeper structural flaw: they evaluate knowledge retrieval via multiple-choice formats rather than actual natural language generation, suffering from selection bias \citep{zheng2024large, balepur-etal-2025-best}. Furthermore, natively sourced benchmarks like INCLUDE \citep{romanou2025include} inherently provide imbalanced cross-lingual comparisons. Evaluating German on the Driver's License topic while testing French on other topics confounds language proficiency with task difficulty. Ultimately, real-world utility depends on coherent, fluent, and semantically faithful generation. We provide empirical evidence demonstrating how current benchmarks fail to capture this, motivating our standardized generative approach.

\textbf{Natural Language Generation} Foundational benchmarks such as MMLU \citep{hendrycks2021measuring, evaluate_nlg} acknowledge that the future of model evaluation lies in Natural Language Generation (NLG), but use multiple-choice formats because NLG remains difficult to evaluate. Consequently, assessing multilingual generation quality remains an open problem.  

Traditional machine translation benchmarks like WMT \citep{kocmi2025findings} and FLORES \citep{nllb2022} directly evaluate translation quality, but they depend on reference translations or per-example human ratings. These benchmarks also draw from Wikipedia and other widely-used sources in training datasets, raising concerns about data contamination into model training \citep{data_cont, karpinska-iyyer-2023-large, vilar-etal-2023-prompting}. Furthermore, most evaluations are  sentence-level, lacking the complexity of real-world translation tasks.

Recent NLG benchmarks such as PolyMath \citep{wang2025polymath}, MGSM \citep{shi2023language}, and MT-AIME24 \citep{son-etal-2025-linguistic} attempt to test multilingual generation by machine-translating English-sourced mathematical questions. However, as shown in Fig~\ref{fig:new_last}, these approaches primarily evaluate mathematical proficiency rather than multilingual capability.

To address these gaps, we curate a benchmark of complex, multi-sentence passages reflecting real-world use cases. We adapt the MQM framework \citep{lommel2014multidimensional}, used in WMT Shared Tasks \citep{freitag-etal-2021-results, lavie-etal-2025-findings} and replace human translators with LLM judges for scalability \citep{lu-etal-2024-error, kocmi-federmann-2023-gemba, kim-2025-rubric}. Separately, LMArena~\cite{chiang2024chatbot} provides a large-scale measure of real-world user preferences. This approach has become the standard for tracking real-world model performance and has inspired work on efficient proxies that maintain high correlation with full arena rankings \citep{li2025from, dubois2024lengthcontrolled, spangher-etal-2025-chatbot}. We use LMArena as a point of comparison for real-world user preferences.


\textbf{Round-Trip Translation.} Round-trip translation has a long history as an evaluation tool. The foundational \citet{brislin1970back} paper established its use for validating translation quality. It later became a widely used data augmentation approach in neural machine translation  \citet{sennrich2016improving}. The core insight is straightforward: if meaning degrades through a round-trip (English $\rightarrow$ target $\rightarrow$ English), the forward translation is usually unreliable. Early statistical models undermined this approach by simply utilizing a "copy mechanism" \citep{somers2005round}. However, modern neural machine translation (NMT) architectures lack this flaw, confirming that round-trip translation is an effective method for reference-free evaluations \citep{moon-etal-2020-revisiting, zhuo-etal-2023-rethinking}. Where prior work mainly uses round-trip translation for MT data augmentation \citep{sennrich2016improving}, we use it to evaluate frontier LLMs \citep{rtt_corr}.

\section{Current Benchmarks Do Not Measure Multilingual Capability}

Frontier model reports \citep{yang2025qwen3technicalreport} evaluate multilingual capabilities using popular multilingual reasoning benchmarks like MT-AIME24 \citep{son-etal-2025-linguistic}, MGSM \citep{shi2023language} and PolyMath~\citep{wang2025polymath}, as well as general knowledge MCQA benchmarks such as MMMLU ~\citep{hendrycks2021measuring} and INCLUDE~\citep{romanou2025include}. In this section, we show that these benchmarks correlate poorly with human preferences because they measure reasoning or factual recall, not multilingual comprehension.

\subsection{Benchmark Scores Diverge from User Preferences}
\label{subsec:user-pref}
Higher scores on a good benchmark should imply better real-world utility. We test whether multilingual benchmarks satisfy this criterion by correlating their scores with user experience in-the-wild.

\textbf{Setup.} We evaluate six frontier open-source models: Kimi K2\footnote{Unlike other models, Kimi K2 Thinking was released after the Non-Thinking variant.}  \citep{team2025kimi}, DeepSeek-V3.2-Exp \citep{liu2025deepseek}, and Qwen3-235B-2507 \citep{yang2025qwen3technicalreport}, each in Thinking and Non-Thinking variants. We constrain the comparison to same-tier models to avoid spurious correlations driven by model size \citep{kaplan2020scalinglawsneurallanguage, ghorbani2022scaling}. We measure zero-shot accuracy on MT-AIME24 \citep{son-etal-2025-linguistic}, a multilingual reasoning benchmark, and INCLUDE \citep{romanou2025include}, a culturally-curated, knowledge-intensive benchmark. We then compute Spearman rank-order correlations between benchmark scores and LMArena \citep{chiang2024chatbot} Elo ratings. For comparison, we also evaluate backtranslation quality on LiT (English → Language → English) across seven LMArena languages\footnote{Chinese, French, German, Japanese, Korean, Russian, Spanish} and report $MQM_{\geq80}$: the percentage of backtranslations whose MQM score is at least 80.

\begin{figure*}[t]
    \centering
    \vspace*{-0.5cm}
        \begin{subfigure}[b]{0.31\textwidth}
        \centering
        \includegraphics[width=\textwidth]{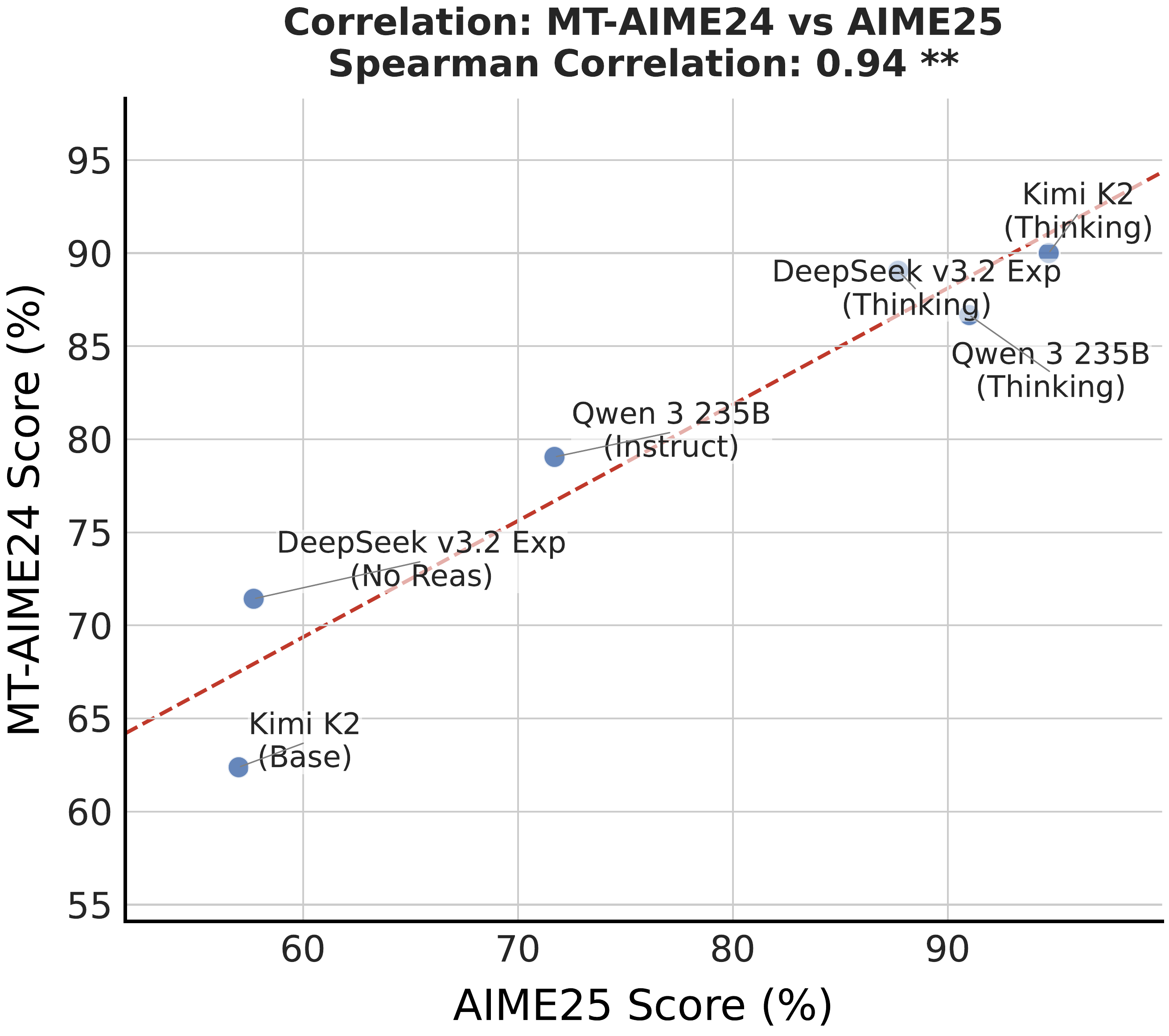}
        \caption{MT-AIME24 vs AIME25}
        \label{fig:mt_aime}
    \end{subfigure}
    \hfill
        \begin{subfigure}[b]{0.31\textwidth}
        \centering
        \includegraphics[width=\textwidth]{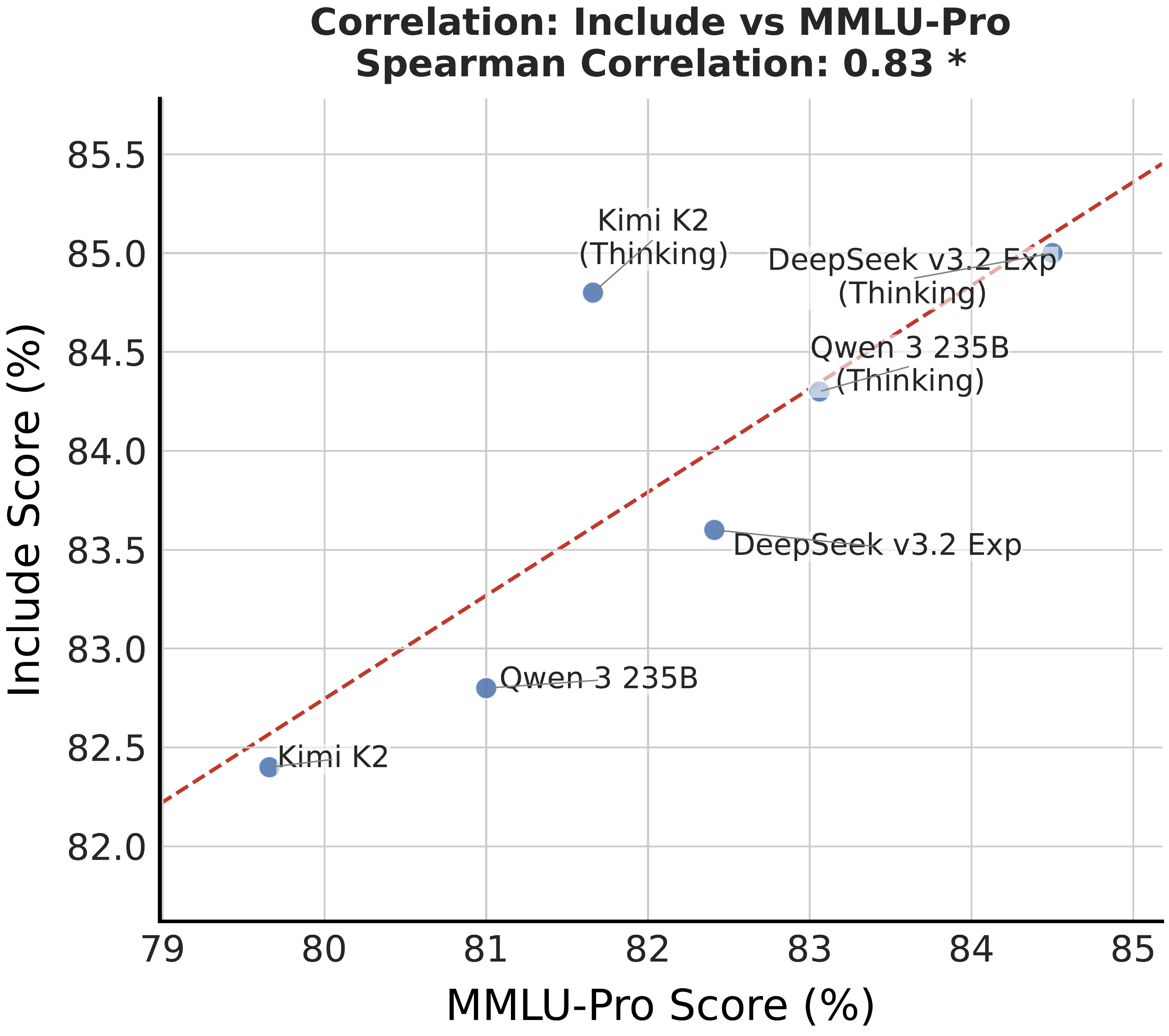}
        \caption{INCLUDE vs MMLU-Pro}
        \label{fig:include-mmlu-pro}
    \end{subfigure}
    \hfill
    \begin{subfigure}[b]{0.31\textwidth}
        \centering
        \includegraphics[width=\textwidth]{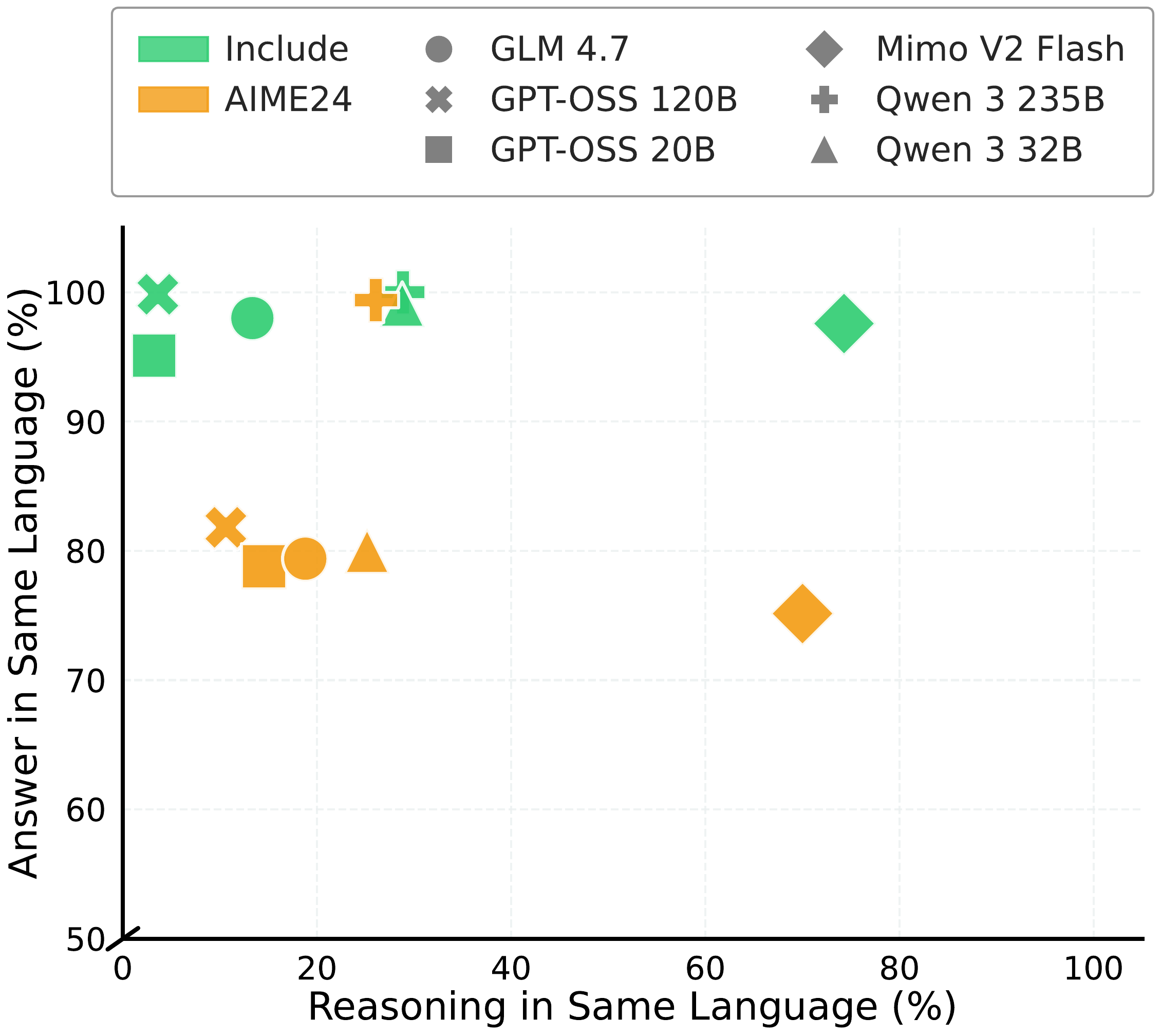}
        \caption{Probing reasoning language}
        \label{fig:reas_lang}
    \end{subfigure}
    \caption{\textbf{Multilingual benchmarks track English reasoning performance, and models reason in English regardless of input language.} (a) MT-AIME24 scores correlate strongly with English AIME25 performance ($\rho = 0.94$, one-sided permutation test p=0.008), indicating the benchmark primarily measures mathematical reasoning ability. (b) INCLUDE scores correlate strongly with English MMLU-Pro ($\rho = 0.83$, one-sided permutation test p=0.029), indicating the benchmark primarily measures factual knowledge. (c) Models overwhelmingly default to English for internal reasoning (y-axis) even when answering in the target language (x-axis). This rules out the possibility that benchmark errors reflect multilingual reasoning failures.}
    \label{fig:new_last}
    \vspace*{-0.3cm}
\end{figure*}

\textbf{Results.} Figure~\ref{fig:corr_plot} shows that benchmark rankings diverge from human preferences. Both MT-AIME24 (left) and INCLUDE (middle) exhibit slight to moderate \textbf{negative} correlation with LMArena Elo scores. The disconnect is strongest between Thinking and Non-Thinking variants of the same model: Thinking models dramatically outperform their counterparts on MT-AIME24 and INCLUDE, yet on LMArena --  where users rate actual multilingual outputs -- they often perform no better. In contrast, round-trip translation scores on LiT correlate almost perfectly ($\rho=0.94$) with LMArena, suggesting closer alignment with real-world multilingual performance. 


\textbf{Analysis.} Why do multilingual benchmarks fail to predict real-world performance? We observe that they conflate two distinct capabilities: reasoning/knowledge and multilingual understanding. Gains on the benchmark may stem from improved reasoning, not improved language proficiency. We  correlate each multilingual benchmark with an English benchmark measuring the same underlying skill: MT-AIME24 with AIME25, and INCLUDE with MMLU-Pro \citep{wang2024mmlupro}. Figures \ref{fig:mt_aime}, \ref{fig:include-mmlu-pro} show near perfect correlations across both benchmark pairs. This suggests that gains on multilingual benchmarks may largely reflect gains on English reasoning and knowledge tasks. Current benchmarks may track reasoning and knowledge gains more than multilingual ones, creating a misleading impression of multilingual progress. Billions who speak languages other than English might see limited benefit from progress on these benchmarks.

\begin{AIbox}{Benchmark Scores Diverge from Real-World Use}
\newcommand{\badge}[2]{%
  \begingroup\setlength{\fboxsep}{1.5pt}\setlength{\fboxrule}{0pt}%
  \colorbox{#1!18}{\textbf{\textcolor{#1!70!black}{#2}}}%
  \endgroup
}

\noindent Multilingual benchmarks conflate two distinct capabilities: reasoning/facts and multilingual understanding, i.e. progress on (MT-)AIME24 primarily stems from better mathematical capability, not improved English (language) proficiency. We need challenging \textit{language-centric} multilingual benchmarks.
\end{AIbox}

\subsection{Errors in Current Benchmarks are not Linguistic}
\label{subsec:err_tax}

We next analyze error types in two common benchmark categories, multilingual reasoning and general-knowledge multiple choice, to test whether these benchmarks reflect reasoning or factual knowledge rather than language proficiency.

\begin{figure*}[t!]
\vspace*{-0.5cm}
    \centering
        \begin{subfigure}[b]{0.48\textwidth}
        \centering
        \includegraphics[width=\textwidth]{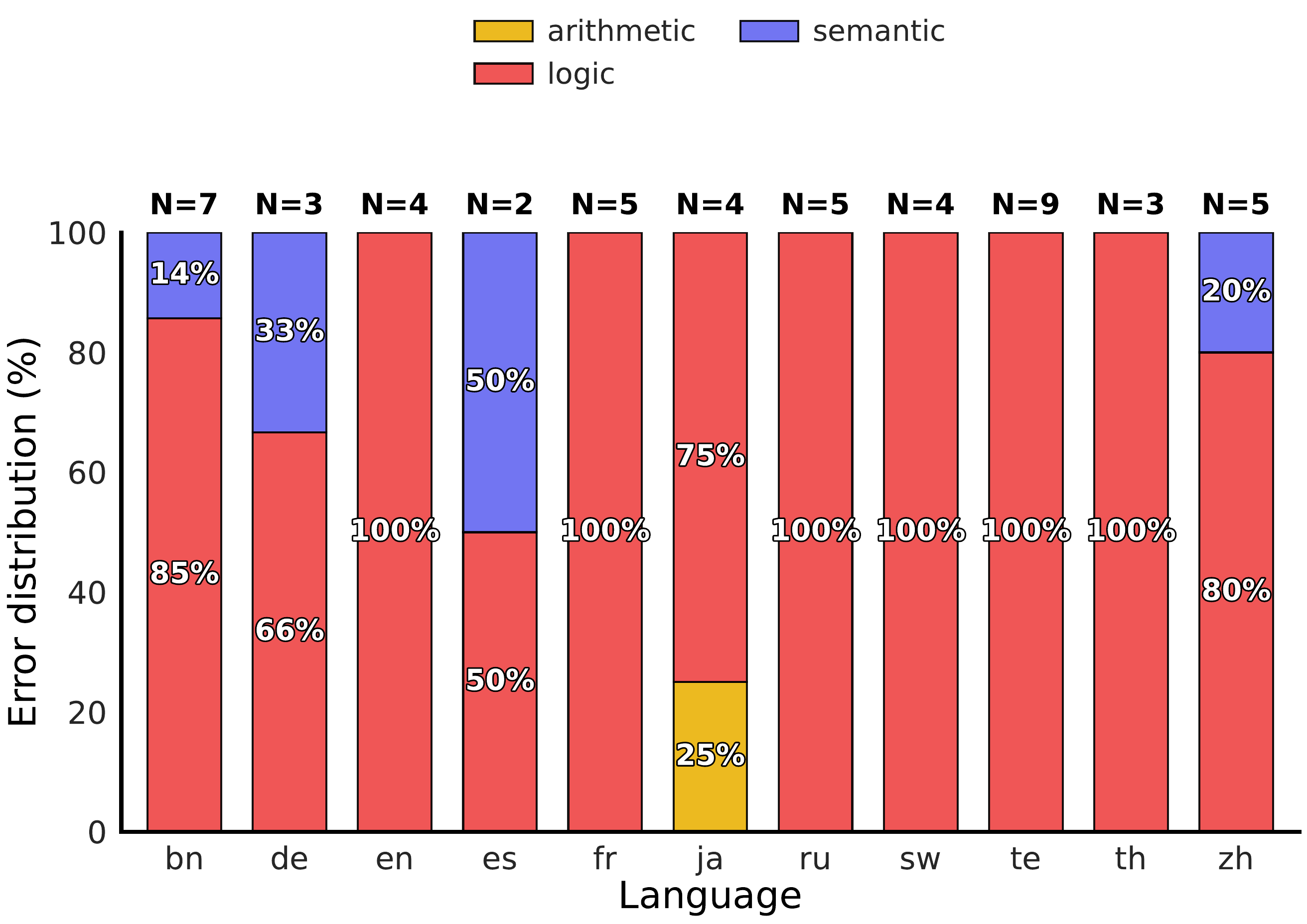}
        \caption{Error distribution for MT-AIME24.}
        \label{fig:err_dist_aime24}
    \end{subfigure}
    \hfill
    \begin{subfigure}[b]{0.48\textwidth}
        \centering
        \includegraphics[width=\textwidth]{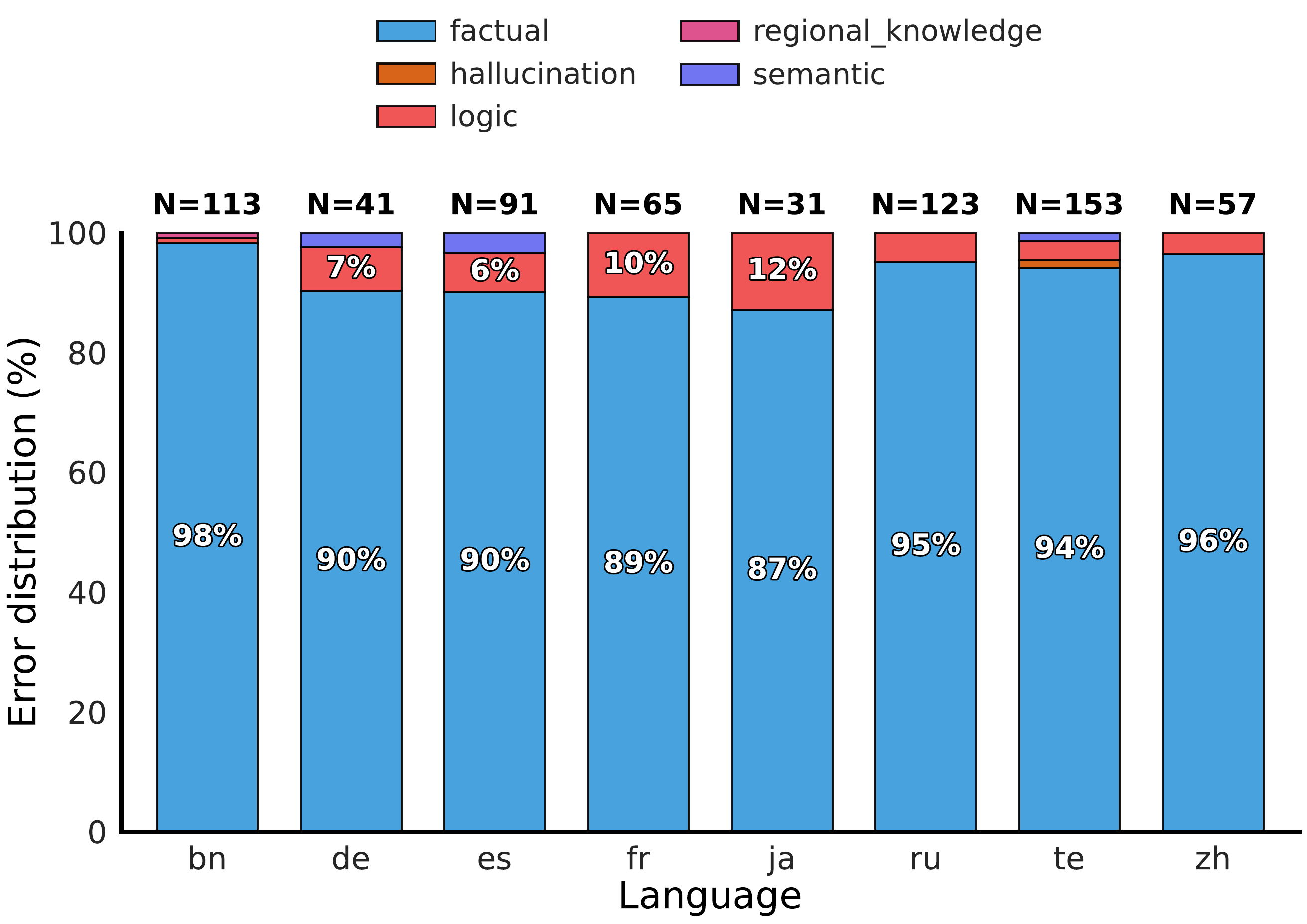}
        \caption{Error distribution for INCLUDE. }
        \label{fig:err_dist_include}
    \end{subfigure}
    \hfill
    \caption{Errors on multilingual benchmarks stem from reasoning and knowledge gaps—not language comprehension failures. We manually categorize all errors made by Qwen3-235B-A22B-Thinking across 11 languages. (a) On MT-AIME24, errors are overwhelmingly logical (arithmetic mistakes, flawed reasoning steps) rather than semantic (misunderstanding the translated question). (b) On INCLUDE, errors are predominantly factual (wrong facts, knowledge gaps) rather than semantic (misunderstanding the translated question). Both patterns confirm that these benchmarks measure reasoning and factual recall -- not multilingual proficiency.}
    \label{fig:err_dist}
    \vspace*{-0.25cm}
\end{figure*}

\textbf{MT-AIME24 errors are logical, not semantic.} We analyze errors for Qwen3-235B-A22B-Thinking-2507 on MT-AIME24 across 11 languages; with similar trends reproducible across a variety of models (detailed in Appendix~\ref{sec:mt_aime_reas}). Figure \ref{fig:err_dist} shows the distribution. We observe that most errors stem from logical mistakes, calculation errors or flawed reasoning steps, and not from poor (multilingual) question comprehension. For French, Russian, and Thai, 100\% of errors reflect failed reasoning, not failed understanding. The general trend is that the model parsed the translated problem correctly; it simply could not solve it. Performance gains on MT-AIME24 reflect better mathematical reasoning, not multilingual capability.

\textbf{INCLUDE errors are factual, not semantic.} One might argue that general knowledge benchmarks such as INCLUDE avoids this problem because it uses native questions rooted in regional knowledge rather than simply translating text from MMLU. However, the same pattern persists, including for other models in Appendix~\ref{sec:include_reas}. Figure \ref{fig:err_dist} shows that most errors stem from incorrect factual knowledge. For Bengali and Chinese, over 96\% of errors reflect knowledge gaps, not misunderstanding of the input question or MCQ options. Performance gains on INCLUDE, therefore, similarly reflect better elicitation and coverage of facts, not better multilingual capability.

\textbf{Reasoning failures occur in English, not the target language.} One could argue that reasoning in the target language is itself worth assessing. We test whether models actually reason in the target language by analyzing traces from multiple language models across both benchmarks. Figure \ref{fig:reas_lang} shows they do not: models almost always default to English for their reasoning traces. The knowledge gaps and logical errors we identify therefore occur in English, not even in the target language. This is further evidence suggesting that reasoning failures do not reflect multilingual failures -- they are reasoning/factual mistakes made in the English language (for extensive results, please refer to the Appendix sections on answer-language and error-distribution analyses).

\textbf{Summary.} Overall, both benchmark types fail to disentangle multilingual capability from orthogonal skills. A model with strong language proficiency, but weak mathematical reasoning (like DeepSeek-V3.2-Exp Non-Thinking) (as seen in Figure \ref{fig:corr_plot}) scores poorly on MT-AIME24. Equivalently, a model with limited factual coverage underperforms on INCLUDE. Because models across newer generations often broadly improve both multilingual and reasoning capabilities at the same time, benchmark scores can appear to track multilingual progress even when they mainly reflect improvement in other areas.

\begin{AIbox}{Errors Don't Reflect Multilingual Failures}
\newcommand{\badge}[2]{%
  \begingroup\setlength{\fboxsep}{1.5pt}\setlength{\fboxrule}{0pt}%
  \colorbox{#1!18}{\textbf{\textcolor{#1!70!black}{#2}}}%
  \endgroup
}

\noindent Errors in current multilingual benchmarks primarily reflect reasoning or knowledge gaps, and these gaps persist in English. Multilingual benchmarks should analyze the errors to understand the gaps in multilingual capabilities.
\end{AIbox}

\section{The Lost In Translation (LiT) Benchmark}

Following our previous analysis of multilingual benchmark shortcomings, we propose Lost In Translation (LiT) -- a natural language generation benchmark centered on backtranslation. LiT addresses the limitations of existing benchmarks summarized in Table~\ref{tab:benchmark_tiers}: limited linguistic diversity, reliance on surface-level metrics, and failure to probe multilingual generation capabilities.

\vspace{-1.0em}
\subsection{Dataset: Description and Setup}
\textbf{Data Curation.} To prevent data contamination from existing corpora, we create a new benchmark of 1600 samples (200 unique texts x 8 language sequences). We manually select samples with diverse linguistic complexities. The dataset spans three major categories, each targeting distinct aspects of translation difficulty, with further details in Appendix \ref{app:additional_details}:  

\textit{\textbf{(a) Abstracts}} (20\%): Split equally between Humanities and STEM. Abstracts are self-contained and semantically dense, making errors easy to detect. Humanities tests argumentative nuance; STEM tests terminological precision and notation.

\textit{\textbf{(b) Pragmatics}} (60\%): Tests meaning beyond literal content across five phenomena: (i) Core Semantics (20\%) i.e. truth conditions and entailment; (ii) Discourse Coherence (17.5\%) i.e. referential chains and logical connectives; (iii) Implicit Content (17.5\%) i.e. presuppositions and implicatures; (iv) Pragmatic Inference (21.7\%) i.e. speech acts and speaker intent; (v) Social Interaction (23.3\%) i.e. politeness, formality, and sociolinguistic appropriateness.

\textit{\textbf{(c) Informal}} (20\%): Tests colloquial language, slang, idioms, and register shifts; preserving tone and social function beyond denotative meaning.

\textbf{Language Sequences.} To rigorously stress-test multilingual capabilities, we evaluate performance across 8 linguistic sequences. We select these sequences to span disjoint language families, distinct scripts (Latin, Cyrillic, Devanagari, Arabic, CJK), and varying levels of pretraining resource availability, We prioritize translation pairs between regions with frequent interactions to reflect real-world translation tasks. Each sequence comprises four languages through which the source text is serially translated before round-trip translation to English:

\begin{itemize}[nosep,leftmargin=*]
    \item \textit{East Asia (E.Asia, High resource)}: Japanese $\rightarrow$ Korean $\rightarrow$ Chinese $\rightarrow$ Russian.
    \item \textit{Central Europe (C.Europe, High resource)}: Romanian $\rightarrow$ Hungarian $\rightarrow$ German $\rightarrow$ Polish.
    \item \textit{Near East (N.East, High resource)}: Bulgarian $\rightarrow$ Greek $\rightarrow$ Turkish $\rightarrow$ Persian.
    \item \textit{North Europe (N.Europe, Medium resource)}: Icelandic $\rightarrow$ Swedish $\rightarrow$ Finnish $\rightarrow$ Lithuanian.
    \item \textit{Southeast Asia (SE.Asia, Medium resource)}: Chinese $\rightarrow$ Thai $\rightarrow$ Vietnamese $\rightarrow$ Tagalog.
    \item \textit{South Asia (S.Asia, Medium resource)}: Hindi $\rightarrow$ Tamil $\rightarrow$ Bengali $\rightarrow$ Punjabi.
    \item \textit{Africa (Africa, Low resource)}: Swahili $\rightarrow$ Arabic $\rightarrow$ Hausa $\rightarrow$ Amharic.
    \item \textit{South America (S.America, Low resource)}: Portuguese $\rightarrow$ Quechua $\rightarrow$ Spanish $\rightarrow$ Guarani.
\end{itemize}

\definecolor{HeatColor}{HTML}{2E8B57}
\begingroup
\newcommand{\avgmqm}[3]{%
  \cellcolor{HeatColor!#1}%
  \shortstack[c]{\rule{0pt}{1.05em}#2\\[0.25em]{\scriptsize $\pm$ #3}}%
}

\begin{table*}[t]
\centering
\caption{LiT benchmark by linguistic category under $MQM_{\geq80}$. We report the percentage of translations with MQM $\geq 80$, aggregated within each category over eight translation sequences. 
Each cell shows the mean and the bootstrap-estimated standard error, computed via sentence-level resampling after averaging across sequences. Higher is better. Informal text is the hardest category overall (30.8 avg), while Core Semantics is the easiest (61.5).}
\resizebox{1.0\textwidth}{!}{
\renewcommand{\arraystretch}{1.55}
\setlength{\tabcolsep}{3.5pt}
\small
\begin{tabular}{>{\raggedright\arraybackslash}m{4.0cm} c cc ccccc c}
\toprule
\multirow{2}{*}{\textbf{Model}} & \multirow{2}{*}{\textbf{Average}} & \multicolumn{2}{c}{\textbf{Abstracts}} & \multicolumn{5}{c}{\textbf{Pragmatics}} & \textbf{Informal} \\
\cmidrule(lr){3-4} \cmidrule(lr){5-9}
 & & \textbf{Humanities} & \textbf{STEM} & \textbf{Core Semantics} & \textbf{Discourse} & \textbf{Implicit} & \textbf{Inference} & \textbf{Social} & \\
\midrule
\shortstack[l]{Gemini-3-Flash\\(No-Thinking)}
 & \avgmqm{44}{87.2}{0.7} & \avgmqm{45}{90.0}{1.9} & \avgmqm{43}{85.6}{1.8} & \avgmqm{44}{87.5}{1.9} & \avgmqm{45}{89.3}{1.5} & \avgmqm{43}{86.9}{2.3} & \avgmqm{44}{88.5}{1.6} & \avgmqm{45}{89.7}{1.6} & \avgmqm{40}{80.3}{2.2} \\
Qwen3.5-397B (Thinking) & \avgmqm{37}{73.8}{1.0} & \avgmqm{35}{70.0}{3.7} & \avgmqm{44}{88.1}{2.7} & \avgmqm{41}{81.2}{2.5} & \avgmqm{38}{75.6}{3.2} & \avgmqm{38}{76.2}{3.1} & \avgmqm{39}{78.4}{2.0} & \avgmqm{38}{75.9}{2.5} & \avgmqm{22}{45.0}{3.5} \\
Gemma-4-31B (Instruct) & \avgmqm{37}{73.0}{1.0} & \avgmqm{36}{71.2}{3.4} & \avgmqm{38}{76.2}{4.4} & \avgmqm{39}{78.1}{2.3} & \avgmqm{38}{75.0}{2.8} & \avgmqm{35}{69.0}{3.1} & \avgmqm{40}{79.8}{1.7} & \avgmqm{39}{77.7}{1.6} & \avgmqm{28}{56.9}{2.7} \\
Gemma-4-31B (Thinking) & \avgmqm{36}{71.9}{1.0} & \avgmqm{35}{70.6}{3.2} & \avgmqm{38}{76.2}{3.4} & \avgmqm{38}{75.5}{2.6} & \avgmqm{37}{74.4}{2.5} & \avgmqm{35}{70.8}{2.9} & \avgmqm{37}{74.0}{2.5} & \avgmqm{39}{77.7}{1.8} & \avgmqm{28}{55.9}{2.9} \\
GLM-5 (Thinking) & \avgmqm{35}{70.3}{1.1} & \avgmqm{33}{66.2}{4.0} & \avgmqm{37}{74.4}{2.9} & \avgmqm{39}{77.1}{3.1} & \avgmqm{39}{77.4}{1.4} & \avgmqm{36}{71.4}{3.3} & \avgmqm{39}{77.4}{1.9} & \avgmqm{36}{72.8}{3.0} & \avgmqm{23}{45.9}{3.2} \\
Qwen3.5-397B (Instruct) & \avgmqm{34}{67.6}{1.2} & \avgmqm{36}{71.2}{4.0} & \avgmqm{35}{69.4}{4.4} & \avgmqm{36}{71.9}{3.8} & \avgmqm{35}{69.6}{2.2} & \avgmqm{33}{66.7}{4.0} & \avgmqm{37}{74.0}{1.9} & \avgmqm{36}{71.4}{2.4} & \avgmqm{23}{46.9}{3.2} \\
Kimi-K2 (Thinking) & \avgmqm{32}{64.7}{0.9} & \avgmqm{32}{63.1}{3.2} & \avgmqm{29}{58.1}{4.0} & \avgmqm{35}{70.6}{1.3} & \avgmqm{35}{70.8}{1.4} & \avgmqm{34}{67.6}{2.6} & \avgmqm{36}{71.6}{1.5} & \avgmqm{34}{68.3}{1.9} & \avgmqm{24}{47.2}{2.4} \\
GLM-4.7 (Thinking) & \avgmqm{31}{61.6}{1.0} & \avgmqm{32}{63.8}{3.2} & \avgmqm{30}{60.6}{3.5} & \avgmqm{36}{72.4}{2.5} & \avgmqm{35}{70.2}{2.0} & \avgmqm{31}{61.3}{3.6} & \avgmqm{34}{67.8}{1.9} & \avgmqm{33}{65.2}{2.3} & \avgmqm{16}{31.6}{2.9} \\
Qwen3-235B (Thinking) & \avgmqm{28}{55.6}{1.0} & \avgmqm{25}{49.4}{3.7} & \avgmqm{26}{51.9}{3.8} & \avgmqm{34}{67.2}{1.9} & \avgmqm{33}{66.7}{2.1} & \avgmqm{30}{60.1}{3.7} & \avgmqm{34}{67.3}{1.8} & \avgmqm{30}{59.8}{2.8} & \avgmqm{11}{22.8}{2.5} \\
DeepSeek-V3.2-Exp & \avgmqm{27}{53.4}{1.1} & \avgmqm{26}{52.5}{3.6} & \avgmqm{21}{42.5}{3.9} & \avgmqm{29}{58.9}{2.9} & \avgmqm{29}{57.1}{3.4} & \avgmqm{29}{58.3}{2.8} & \avgmqm{32}{63.9}{2.7} & \avgmqm{29}{57.6}{2.6} & \avgmqm{18}{36.6}{3.1} \\
GLM-5 (Instruct) & \avgmqm{27}{53.3}{1.2} & \avgmqm{26}{51.2}{4.2} & \avgmqm{22}{45.0}{4.2} & \avgmqm{30}{60.9}{2.2} & \avgmqm{30}{60.7}{3.6} & \avgmqm{27}{54.8}{3.7} & \avgmqm{30}{59.1}{2.7} & \avgmqm{31}{61.2}{2.6} & \avgmqm{17}{33.1}{3.2} \\
\shortstack[l]{DeepSeek-V3.2-Exp\\(Thinking)}
 & \avgmqm{27}{53.2}{1.3} & \avgmqm{28}{55.0}{5.5} & \avgmqm{18}{36.9}{5.1} & \avgmqm{33}{65.6}{3.1} & \avgmqm{31}{61.9}{2.9} & \avgmqm{30}{60.1}{3.4} & \avgmqm{29}{58.7}{3.5} & \avgmqm{30}{59.8}{2.3} & \avgmqm{14}{27.8}{2.8} \\
Qwen3.5-35B (Thinking) & \avgmqm{26}{52.0}{1.2} & \avgmqm{24}{48.1}{4.4} & \avgmqm{26}{52.5}{3.7} & \avgmqm{32}{63.5}{2.5} & \avgmqm{30}{60.1}{3.4} & \avgmqm{27}{54.2}{4.6} & \avgmqm{32}{63.5}{2.5} & \avgmqm{29}{58.5}{2.5} & \avgmqm{8}{15.9}{2.8} \\
Kimi-K2 & \avgmqm{25}{49.7}{1.3} & \avgmqm{24}{47.2}{3.8} & \avgmqm{22}{43.3}{5.3} & \avgmqm{30}{60.9}{2.9} & \avgmqm{28}{55.6}{4.0} & \avgmqm{25}{49.7}{4.0} & \avgmqm{29}{57.4}{2.3} & \avgmqm{26}{51.3}{3.1} & \avgmqm{16}{32.1}{3.3} \\
Gemma-3-27B (Instruct) & \avgmqm{24}{47.8}{1.3} & \avgmqm{22}{43.8}{4.6} & \avgmqm{14}{28.1}{5.3} & \avgmqm{30}{60.4}{2.9} & \avgmqm{28}{55.4}{3.4} & \avgmqm{25}{49.4}{3.7} & \avgmqm{30}{60.1}{1.9} & \avgmqm{26}{52.2}{2.8} & \avgmqm{16}{32.8}{3.4} \\
Qwen3-235B (Instruct) & \avgmqm{22}{44.1}{1.2} & \avgmqm{20}{39.4}{3.8} & \avgmqm{18}{35.0}{4.4} & \avgmqm{27}{53.6}{2.6} & \avgmqm{26}{52.4}{3.2} & \avgmqm{24}{48.2}{3.3} & \avgmqm{25}{49.5}{2.3} & \avgmqm{25}{49.6}{3.4} & \avgmqm{13}{25.3}{3.2} \\
GPT-OSS-120B (High) & \avgmqm{21}{41.9}{1.3} & \avgmqm{18}{35.8}{4.4} & \avgmqm{21}{41.2}{5.5} & \avgmqm{28}{55.7}{2.5} & \avgmqm{27}{54.8}{2.5} & \avgmqm{20}{39.3}{4.6} & \avgmqm{28}{56.2}{2.7} & \avgmqm{22}{44.2}{3.3} & \avgmqm{4}{7.8}{2.1} \\
MiniMax-M2.5 & \avgmqm{20}{39.9}{1.2} & \avgmqm{15}{30.6}{3.2} & \avgmqm{18}{36.2}{4.3} & \avgmqm{29}{57.8}{3.4} & \avgmqm{26}{51.8}{3.6} & \avgmqm{21}{42.9}{4.9} & \avgmqm{25}{49.0}{3.0} & \avgmqm{22}{43.3}{2.9} & \avgmqm{4}{7.2}{1.8} \\
Qwen3.5-35B (Instruct) & \avgmqm{19}{38.2}{1.3} & \avgmqm{16}{32.5}{4.3} & \avgmqm{18}{36.9}{4.1} & \avgmqm{25}{49.5}{3.3} & \avgmqm{21}{41.7}{3.4} & \avgmqm{20}{40.5}{5.0} & \avgmqm{26}{51.9}{2.7} & \avgmqm{20}{40.2}{3.1} & \avgmqm{6}{12.8}{2.2} \\
MiMo-V2-Flash & \avgmqm{15}{30.2}{1.1} & \avgmqm{12}{25.0}{3.8} & \avgmqm{7}{14.4}{3.6} & \avgmqm{22}{44.3}{3.2} & \avgmqm{21}{41.7}{3.3} & \avgmqm{17}{34.5}{3.5} & \avgmqm{20}{40.9}{3.2} & \avgmqm{15}{30.8}{2.8} & \avgmqm{5}{10.3}{2.3} \\
Qwen3-30B (Instruct) & \avgmqm{11}{21.4}{1.0} & \avgmqm{7}{14.4}{2.5} & \avgmqm{10}{19.4}{4.1} & \avgmqm{15}{30.7}{3.0} & \avgmqm{14}{28.0}{3.0} & \avgmqm{12}{23.2}{3.4} & \avgmqm{13}{25.5}{2.3} & \avgmqm{13}{26.8}{2.4} & \avgmqm{2}{3.4}{1.1} \\
Nemotron-3-Nano & \avgmqm{3}{5.5}{0.5} & \avgmqm{3}{5.6}{1.9} & \avgmqm{1}{1.2}{0.8} & \avgmqm{5}{10.4}{1.8} & \avgmqm{4}{7.7}{1.6} & \avgmqm{1}{3.0}{1.4} & \avgmqm{5}{10.6}{1.9} & \avgmqm{3}{5.8}{1.2} & \avgmqm{0}{0.0}{0.0} \\
\midrule
\textbf{Average} & \cellcolor{HeatColor!26}52.6 & \cellcolor{HeatColor!25}49.8 & \cellcolor{HeatColor!24}48.8 & \cellcolor{HeatColor!31}61.5 & \cellcolor{HeatColor!30}59.0 & \cellcolor{HeatColor!27}54.0 & \cellcolor{HeatColor!30}60.2 & \cellcolor{HeatColor!28}56.4 & \cellcolor{HeatColor!15}30.8 \\
\bottomrule
\end{tabular}}
\vspace*{-0.4cm}
\label{tab:mqm_scores_hierarchical_sorted}
\end{table*}
\endgroup

\begin{table}[t!]
\centering
\renewcommand{\arraystretch}{1.2}
\setlength{\tabcolsep}{3pt}
\small
\caption{
\textbf{$MQM_{\geq80}$ collapses on low-resource language sequences.} We report \textbf{$MQM_{\geq80}$} rates, the percentage of subset examples with MQM $\geq 80$, across the same eight language sequences grouped by resource availability. (a) High-resource sequences remain solvable for the strongest frontier models, with Gemini-3-Flash averaging 97.0 and reaching 98.0 on both Central European and Near Eastern chains. (b) Medium-resource sequences remain relatively stable for the best models, though separation widens below the frontier tier. (c) Low-resource sequences collapse sharply: only Gemini-3-Flash sustains a substantial pass rate, averaging 59.2 across African and South American chains; the next-best model drops to 32.0. (d) Global averages across these eight sequences show Gemini-3-Flash far ahead of all competitors.
}

\begin{minipage}[t]{0.555\textwidth}
    \centering
    \textbf{(a) High \& Med-High Resource Sequences} \\
    \vspace{2pt}
    \resizebox{\linewidth}{!}{
    \begin{tabular}{l|ccc|c}
        \toprule
        \textbf{Model} & \textbf{E. Asia} & \textbf{C. Europe} & \textbf{N. East} & \textbf{Avg} \\
        \midrule
        Gemini-3-Flash (No-Think) & 95.0 & \textbf{98.0} & \textbf{98.0} & \cellcolor{gray!15}\textbf{97.0} \\
        GLM-5 (Thinking) & \textbf{93.5} & 93.5 & 82.5 & 89.8 \\
        Qwen3.5-397B (Thinking) & 91.0 & 92.0 & 85.5 & 89.5 \\
        Gemma-4-31B (Instruct) & 90.5 & 90.5 & 86.0 & 89.0 \\
        Gemma-4-31B (Thinking) & \textbf{93.5} & 87.5 & 86.0 & 89.0 \\
        GLM-4.7 (Thinking) & 92.0 & 90.0 & 80.5 & 87.5 \\
        Qwen3.5-397B (Instruct) & 88.5 & 91.5 & 82.0 & 87.3 \\
        Kimi-K2 (Thinking) & 84.0 & 89.0 & 79.0 & 84.0 \\
        DeepSeek-V3.2-Exp & 87.5 & 85.0 & 71.5 & 81.3 \\
        GLM-5 (Instruct) & 87.5 & 86.0 & 69.5 & 81.0 \\
        Qwen3-235B (Thinking) & 82.5 & 86.0 & 71.5 & 80.0 \\
        DeepSeek-V3.2-Exp (Think) & 84.5 & 75.0 & 69.5 & 76.3 \\
        Qwen3.5-35B (Thinking) & 76.0 & 83.0 & 65.5 & 74.8 \\
        Gemma-3-27B (Instruct) & 70.0 & 79.5 & 68.5 & 72.7 \\
        Kimi-K2 & 71.5 & 78.0 & 56.5 & 68.7 \\
        Qwen3-235B (Instruct) & 83.0 & 62.5 & 57.5 & 67.7 \\
        MiniMax-M2.5 & 60.0 & 75.5 & 58.5 & 64.7 \\
        Qwen3.5-35B (Instruct) & 71.5 & 73.5 & 46.5 & 63.8 \\
        GPT-OSS-120B (High) & 68.5 & 65.5 & 49.0 & 61.0 \\
        MiMo-V2-Flash & 57.0 & 61.5 & 30.5 & 49.7 \\
        Qwen3-30B (Instruct) & 56.5 & 49.0 & 21.5 & 42.3 \\
        Nemotron-3-Nano & 29.5 & 4.0 & 0.0 & 11.2 \\
        \bottomrule
    \end{tabular}}

    \vspace{12pt}

    \centering
    \textbf{(b) Medium Resource Sequences} \\
    \vspace{2pt}
    \resizebox{\linewidth}{!}{
    \begin{tabular}{l|ccc|c}
        \toprule
        \textbf{Model} & \textbf{SE. Asia} & \textbf{S. Asia} & \textbf{N. Europe} & \textbf{Avg} \\
        \midrule
        Gemini-3-Flash (No-Think) & \textbf{93.0} & \textbf{95.5} & \textbf{95.5} & \cellcolor{gray!15}\textbf{94.7} \\
        Qwen3.5-397B (Thinking) & 82.0 & 77.0 & 87.0 & 82.0 \\
        Gemma-4-31B (Instruct) & 92.5 & 90.0 & 61.5 & 81.3 \\
        Gemma-4-31B (Thinking) & 88.5 & 93.0 & 62.5 & 81.3 \\
        GLM-5 (Thinking) & 86.5 & 74.0 & 79.5 & 80.0 \\
        Qwen3.5-397B (Instruct) & 83.5 & 71.0 & 75.0 & 76.5 \\
        GLM-4.7 (Thinking) & 70.5 & 66.0 & 63.5 & 66.7 \\
        Kimi-K2 (Thinking) & 71.5 & 58.5 & 65.0 & 65.0 \\
        Qwen3-235B (Thinking) & 74.5 & 55.0 & 53.5 & 61.0 \\
        DeepSeek-V3.2-Exp & 75.0 & 46.5 & 51.5 & 57.7 \\
        GLM-5 (Instruct) & 63.5 & 57.0 & 50.0 & 56.8 \\
        DeepSeek-V3.2-Exp (Think) & 69.5 & 45.5 & 53.0 & 56.0 \\
        Qwen3.5-35B (Thinking) & 60.5 & 48.0 & 56.0 & 54.8 \\
        Gemma-3-27B (Instruct) & 60.5 & 50.0 & 46.5 & 52.3 \\
        Kimi-K2 & 57.0 & 42.0 & 54.0 & 51.0 \\
        Qwen3-235B (Instruct) & 69.0 & 48.0 & 22.5 & 46.5 \\
        GPT-OSS-120B (High) & 49.0 & 37.5 & 42.5 & 43.0 \\
        MiniMax-M2.5 & 45.0 & 16.0 & 43.0 & 34.7 \\
        Qwen3.5-35B (Instruct) & 46.0 & 21.0 & 30.5 & 32.5 \\
        MiMo-V2-Flash & 41.5 & 18.0 & 18.5 & 26.0 \\
        Qwen3-30B (Instruct) & 28.0 & 5.0 & 1.5 & 11.5 \\
        Nemotron-3-Nano & 2.5 & 0.0 & 0.0 & 0.8 \\
        \bottomrule
    \end{tabular}}
\end{minipage}
\hfill
\begin{minipage}[t]{0.435\textwidth}
    \centering
    \textbf{(c) Low Resource / Imbalanced Sequences} \\
    \vspace{2pt}
    \resizebox{\linewidth}{!}{
    \begin{tabular}{l|cc|c}
        \toprule
        \textbf{Model} & \textbf{Africa} & \textbf{S. America} & \textbf{Avg} \\
        \midrule
        Gemini-3-Flash (No-Think) & \textbf{78.5} & \textbf{40.0} & \cellcolor{gray!15}\textbf{59.2} \\
        Gemma-4-31B (Instruct) & 60.0 & 4.0 & 32.0 \\
        Qwen3.5-397B (Thinking) & 41.0 & 15.0 & 28.0 \\
        Gemma-4-31B (Thinking) & 51.5 & 3.0 & 27.2 \\
        GLM-5 (Thinking) & 30.5 & 7.0 & 18.8 \\
        Qwen3.5-397B (Instruct) & 27.5 & 9.0 & 18.2 \\
        DeepSeek-V3.2-Exp (Think) & 8.5 & 6.0 & 7.2 \\
        GLM-4.7 (Thinking) & 8.0 & 3.0 & 5.5 \\
        Nemotron-3-Nano & 1.5 & 4.5 & 3.0 \\
        Kimi-K2 (Thinking) & 5.0 & 0.5 & 2.8 \\
        Qwen3.5-35B (Thinking) & 4.0 & 1.5 & 2.8 \\
        MiMo-V2-Flash & 0.0 & 4.5 & 2.2 \\
        GPT-OSS-120B (High) & 1.5 & 1.0 & 1.3 \\
        Qwen3.5-35B (Instruct) & 1.0 & 1.5 & 1.2 \\
        Qwen3-235B (Thinking) & 0.5 & 2.0 & 1.2 \\
        Gemma-3-27B (Instruct) & 2.0 & 0.0 & 1.0 \\
        DeepSeek-V3.2-Exp & 1.5 & 0.5 & 1.0 \\
        GLM-5 (Instruct) & 2.0 & 0.0 & 1.0 \\
        MiniMax-M2.5 & 1.0 & 0.5 & 0.8 \\
        Kimi-K2 & 1.0 & 0.0 & 0.5 \\
        Qwen3-235B (Instruct) & 0.5 & 0.0 & 0.2 \\
        Qwen3-30B (Instruct) & 0.0 & 0.0 & 0.0 \\
        \bottomrule
    \end{tabular}}

    \vspace{12pt}

    \centering
    \textbf{(d) Overall Performance} \\
    \vspace{2pt}
    \resizebox{\linewidth}{!}{
    \begin{tabular}{l|c}
        \toprule
        \textbf{Model} & \textbf{Global Average} \\
        \midrule
        Gemini-3-Flash (No-Think) & \cellcolor{gray!15}\textbf{86.7} \\
        Gemma-4-31B (Instruct) & 71.9 \\
        Qwen3.5-397B (Thinking) & 71.3 \\
        Gemma-4-31B (Thinking) & 70.7 \\
        GLM-5 (Thinking) & 68.4 \\
        Qwen3.5-397B (Instruct) & 66.0 \\
        GLM-4.7 (Thinking) & 59.2 \\
        Kimi-K2 (Thinking) & 56.6 \\
        Qwen3-235B (Thinking) & 53.2 \\
        DeepSeek-V3.2-Exp & 52.4 \\
        GLM-5 (Instruct) & 51.9 \\
        DeepSeek-V3.2-Exp (Think) & 51.4 \\
        Qwen3.5-35B (Thinking) & 49.3 \\
        Gemma-3-27B (Instruct) & 47.1 \\
        Kimi-K2 & 45.0 \\
        Qwen3-235B (Instruct) & 42.9 \\
        GPT-OSS-120B (High) & 39.3 \\
        MiniMax-M2.5 & 37.4 \\
        Qwen3.5-35B (Instruct) & 36.4 \\
        MiMo-V2-Flash & 28.9 \\
        Qwen3-30B (Instruct) & 20.2 \\
        Nemotron-3-Nano & 5.2 \\
        \bottomrule
    \end{tabular}}
\end{minipage}
\label{tab:mqm_pass80_quadrant_breakdown}
\vspace*{-0.4cm}
\end{table}

The sequences are also grouped into sequences containing exclusively well-resourced languages or including at least one medium- or low-resource language.

\textbf{Evaluation: LLM-as-a-Judge.} We employ Grok 4.1 Fast  as our primary judge model, being an inexpensive, fast but simultaneously very capable model -- consistently one of the highest ranked in multilingual LMArena \citep{chiang2024chatbot}. Following standard practice~\cite{lommel2014multidimensional, kim-2025-rubric}, we adopt the MQM framework, a weighted penalty system that categorizes errors by severity: (i) \textit{Minor} ($-1$): Slight awkwardness or non-critical fluency issues that do not affect comprehension; (ii) \textit{Major} ($-5$): Significant semantic shifts, structural failures, or mistranslations that alter meaning; (iii) \textit{Critical} ($-25$): Complete loss of meaning, hallucinations, or safety-relevant errors. The framework grounds evaluation in interpretable error categories, where low MQM scores indicate severe failure modes dominated by critical errors, with 80 widely considered to be a pass threshold \citep{farinha-etal-2022-findings, lommel-etal-2024-multi}. We additionally show in Appendix \ref{app:robustness_judge} that both the scores and rankings nearly perfectly correlate when using two alternatives: (i) raw MQM scores and (ii) LLM judge providing a score from 0-100, demonstrating robustness to the metrics. The Appendix results additionally show robustness to the judge model, allowing us to fix the metric to MQM$_{\ge80}$ and Grok 4.1 Fast judge model for the rest of the section.

\vspace{-0.2cm}
\subsection{Main Results}

\textbf{A Linguistic Category View.} We present our results in Table~\ref{tab:mqm_scores_hierarchical_sorted}. The trends from the previous section also appear across LiT categories: reasoning effects are model-dependent, helping Qwen3-235B-A22B-2507 much more than DeepSeek-V3.2-Exp, while consistently hurting informal translation. Gemini-3-Flash achieves the highest overall score and the highest score in each subcategory. The high gap between Gemini and other models indicates that our benchmark can measure performance across the "long tail" of scenarios efficiently.
 
\textit{Category-wise breakdown.} To understand failure modes, Table~\ref{tab:mqm_scores_hierarchical_sorted} decomposes performance by linguistic category. STEM Abstracts and Informal registers emerge as the most challenging categories, with average performance of 48.8 and 30.8 respectively. Each entry reports the mean and a bootstrap-estimated standard error, computed from $N=10,000$ sentence-level resamples after averaging across sequences. Technical content tests whether models can communicate domain knowledge fluently.
Informal text tests whether models handle slang and colloquialisms in translation. We observe a striking pattern: reasoning models underperform on informal text. One possible explanation is that reasoning models overcomplicate simple colloquial inputs, producing a mismatch in register.

\textbf{A Language-Centric View.} Table~\ref{tab:mqm_pass80_quadrant_breakdown} presents our results grouped by sequence and by high/medium/low resource classification. Most notably, performance drops sharply from high-resource to low-resource settings. Medium-resource language sequences largely follow high-resource trends.

\textit{High-Resource Stability.} In East Asian, Near Eastern, and Central European sequences, most frontier models achieve strong performance. As observed in Table~\ref{tab:mqm_pass80_quadrant_breakdown}, Gemini-3-Flash leads with an $MQM_{\geq80}$ score of 97.0, followed by GLM-5 (Thinking) at 89.8. Most models perform well on these sequences.

\textit{Low-Resource Collapse.} In contrast, performance collapses in African and South American sequences. Except Gemini-3-Flash which maintains a score of 59.2, Table~\ref{tab:mqm_pass80_quadrant_breakdown} shows that the $MQM_{\geq80}$ score of the next best model -- Gemma-4-31B (Instruct) -- drops to merely 32.0. Several models, including Qwen3-30B (0.0) and Qwen3-235B (0.2), receive near-zero scores, signalling critical errors and unusable translations. This confirms the official result that Qwen3-235B does not officially support some low-resource languages in this subset, such as Quechua or Amharic \citep{qwen3_blog_2025}.

\textit{Impact of Reasoning on Translation.} Comparing thinking variants with instruct counterparts in Table~\ref{tab:mqm_pass80_quadrant_breakdown} suggests that inference-time reasoning helps most in low-resource languages. The thinking variant of DeepSeek-V3.2-Exp scores 7.2 in low-resource settings, outperforming the instruct model's $MQM_{\geq80}$ score of 1.0. These results suggest that reasoning cannot compensate for missing lexical knowledge, but may mitigate some severe errors in unfamiliar linguistic settings. More broadly, however, reasoning does not reliably improve translation quality: in high-resource settings it is often only comparable to instruct models, and even recently released models such as Gemma-4-31B show the instruct variant matching or outperforming the thinking variant overall, with the clearest advantage in low-resource settings.

\begin{AIbox}{Key Findings}
\newcommand{\badge}[2]{%
  \begingroup\setlength{\fboxsep}{1.5pt}\setlength{\fboxrule}{0pt}%
  \colorbox{#1!18}{\textbf{\textcolor{#1!70!black}{#2}}}%
  \endgroup
}

\noindent \begin{itemize}[nosep,leftmargin=*]
    \item Reasoning often hurts translation quality, especially on informal text.
    \item A catastrophic accuracy cliff separates high-resource from low-resource languages.
    \item Reasoning models outperform instruct models primarily in low-resource languages
\end{itemize}
\end{AIbox}
\vspace{-0.1cm}
\subsection{Round-Trip Translation Robustness}
\vspace{-0.05cm}
While LiT uses round-trip translation, the method has several known limitations \citep{somers2005round, 10.1007/11941439_149}. A correct round-trip translation does not necessarily imply correct intermediate translations, which may contain awkward phrasing, inappropriate formality, register mismatches, or missing cultural nuance that disappear in the final back-translation. To assess whether these classical concerns still affect current models, we evaluate a separate 480-sample robustness benchmark. Our results show that modern LLMs are largely robust to these failure modes, and that round-trip translation remains informative on such challenging cases. Additional details are provided in Appendix~\ref{app:robustness_benchmark_details}.

\begin{itemize}[nosep,leftmargin=*]
    \item \textbf{Polysemy} where words carry multiple meanings that only context can resolve.\vspace{0.05cm}
    \item \textbf{Syntactic ambiguity} where sentence structure remains unclear until the final word.\vspace{0.05cm}
    \item \textbf{Idioms and cultural metaphors} to test whether literal translation destroys meaning.\vspace{0.05cm}
    \item \textbf{Register shifts} including formal, colloquial \& technical language within a single example.\vspace{0.05cm}
    \item \textbf{Abstract nuance} where physical vocabulary describes non-physical concepts.
\end{itemize} 


\definecolor{HeatColor}{HTML}{2E8B57}
\begingroup
\newcommand{\avgmqm}[3]{%
  \cellcolor{HeatColor!#1}%
  \shortstack[c]{\rule{0pt}{1.05em}#2\\[0.25em]{\scriptsize $\pm$ #3}}%
}
\newcommand{\modelcell}[1]{%
  \shortstack[l]{\rule{0pt}{1.2em}#1}%
}

\begin{table}[h]
\centering
\vspace*{-0.1cm}
\caption{$MQM_{\geq80}$ on the Backtranslation robustness subset, which targets classical round-trip translation challenges: idioms, polysemy, register shifts, and garden-path syntax. We report the percentage of sentence-sequence translations with MQM $\geq 80$, aggregated within each category over the same eight translation sequences. Each cell shows the mean and bootstrap-estimated standard error, computed via sentence-level resampling after averaging across sequences. Higher is better. Each category contains 12 source sentences. Idioms \& Cultural Metaphors is the hardest category overall (35.3 avg), while Conceptual \& Abstract Nuance is the easiest (54.8).}
\vspace*{-0.2cm}
\resizebox{0.95\textwidth}{!}{
\renewcommand{\arraystretch}{1.55}
\small
\setlength{\tabcolsep}{2pt}
\begin{tabular}{>{\raggedright\arraybackslash}m{4.2cm} c ccccc}
\toprule
\multirow{2}{*}{\textbf{Model}} & \multirow{2}{*}{\textbf{Average}} & \textbf{Conceptual \&} & \textbf{Idioms \&} & \textbf{Polysemy \&} & \textbf{Register \&} & \textbf{Syntactic Complexity} \\
 & & \textbf{Abstract Nuance} & \textbf{Cultural Metaphors} & \textbf{Lexical Ambiguity} & \textbf{Tone Shifts} & \textbf{\& Garden Paths} \\
\midrule

\modelcell{Gemini-3-Flash\\(No-Thinking)} & \avgmqm{38}{71.0}{2.1} & \avgmqm{44}{83.3}{3.4} & \avgmqm{35}{65.6}{5.9} & \avgmqm{37}{68.8}{5.4} & \avgmqm{32}{60.4}{5.3} & \avgmqm{41}{77.1}{2.5} \\
\modelcell{Qwen3.5-397B (Thinking)} & \avgmqm{34}{63.5}{2.0} & \avgmqm{40}{75.0}{3.6} & \avgmqm{33}{62.5}{5.9} & \avgmqm{32}{59.4}{4.2} & \avgmqm{31}{58.3}{3.4} & \avgmqm{33}{62.5}{4.2} \\
\modelcell{GLM-5 (Thinking)} & \avgmqm{32}{59.8}{2.0} & \avgmqm{39}{74.0}{2.7} & \avgmqm{29}{54.2}{4.0} & \avgmqm{27}{50.0}{6.9} & \avgmqm{31}{57.3}{3.5} & \avgmqm{34}{63.5}{4.0} \\
\modelcell{Gemma-4-31B\\(Thinking)} & \avgmqm{30}{55.8}{2.4} & \avgmqm{40}{75.0}{3.6} & \avgmqm{26}{47.9}{6.0} & \avgmqm{29}{54.2}{6.2} & \avgmqm{29}{55.2}{4.7} & \avgmqm{25}{46.9}{5.9} \\
\modelcell{Kimi-K2 (Thinking)} & \avgmqm{29}{54.6}{1.8} & \avgmqm{36}{67.7}{2.3} & \avgmqm{26}{49.0}{4.8} & \avgmqm{29}{55.2}{3.6} & \avgmqm{27}{50.0}{5.1} & \avgmqm{27}{51.0}{3.3} \\
\modelcell{Gemma-4-31B\\(Instruct)} & \avgmqm{28}{52.5}{2.2} & \avgmqm{38}{70.8}{3.7} & \avgmqm{24}{44.8}{4.0} & \avgmqm{26}{49.0}{5.1} & \avgmqm{28}{53.1}{6.1} & \avgmqm{24}{44.8}{5.2} \\
\modelcell{Qwen3.5-397B (Instruct)} & \avgmqm{28}{51.9}{2.1} & \avgmqm{38}{70.8}{3.7} & \avgmqm{21}{38.5}{5.0} & \avgmqm{30}{56.2}{3.5} & \avgmqm{28}{52.1}{5.5} & \avgmqm{22}{41.7}{5.3} \\
\modelcell{GLM-4.7} & \avgmqm{26}{48.8}{2.0} & \avgmqm{33}{61.5}{2.7} & \avgmqm{23}{43.8}{5.6} & \avgmqm{26}{47.9}{4.6} & \avgmqm{23}{43.8}{5.3} & \avgmqm{25}{46.9}{4.2} \\
\modelcell{Qwen3-235B (Thinking)} & \avgmqm{25}{47.5}{2.1} & \avgmqm{32}{60.4}{3.2} & \avgmqm{23}{42.7}{5.6} & \avgmqm{24}{45.8}{5.9} & \avgmqm{20}{37.5}{5.1} & \avgmqm{27}{51.0}{3.4} \\
\modelcell{Qwen3.5-35B (Thinking)} & \avgmqm{23}{43.5}{2.4} & \avgmqm{31}{57.3}{4.5} & \avgmqm{22}{40.6}{5.5} & \avgmqm{21}{38.5}{5.8} & \avgmqm{23}{43.8}{5.9} & \avgmqm{20}{37.5}{5.1} \\
\modelcell{DeepSeek-V3.2-Exp\\(Thinking)} & \avgmqm{22}{41.0}{2.4} & \avgmqm{31}{57.3}{4.6} & \avgmqm{16}{29.2}{5.0} & \avgmqm{23}{43.8}{6.6} & \avgmqm{18}{34.4}{4.2} & \avgmqm{22}{40.6}{6.1} \\
\modelcell{DeepSeek-V3.2-Exp} & \avgmqm{21}{39.2}{2.2} & \avgmqm{28}{53.1}{4.2} & \avgmqm{18}{34.4}{5.6} & \avgmqm{17}{31.2}{5.2} & \avgmqm{17}{32.3}{5.2} & \avgmqm{24}{44.8}{4.6} \\
\modelcell{MiniMax-M2.5} & \avgmqm{21}{38.8}{2.5} & \avgmqm{23}{43.8}{4.8} & \avgmqm{18}{33.3}{5.4} & \avgmqm{22}{41.7}{5.9} & \avgmqm{17}{32.3}{6.8} & \avgmqm{23}{42.7}{4.3} \\
\modelcell{Gemma-3-27B\\(Instruct)} & \avgmqm{20}{37.5}{1.9} & \avgmqm{31}{57.3}{3.1} & \avgmqm{17}{31.2}{4.3} & \avgmqm{18}{34.4}{5.2} & \avgmqm{20}{37.5}{4.4} & \avgmqm{14}{27.1}{4.2} \\
\modelcell{GLM-5 (Instruct)} & \avgmqm{20}{36.9}{2.1} & \avgmqm{27}{50.0}{4.9} & \avgmqm{14}{26.0}{4.5} & \avgmqm{18}{34.4}{5.6} & \avgmqm{16}{30.2}{4.0} & \avgmqm{23}{43.8}{4.0} \\
\modelcell{GPT-OSS-120B\\(High)} & \avgmqm{18}{34.6}{2.4} & \avgmqm{20}{37.5}{5.1} & \avgmqm{20}{37.5}{3.6} & \avgmqm{16}{30.2}{5.9} & \avgmqm{21}{38.5}{5.8} & \avgmqm{16}{29.2}{5.7} \\
\modelcell{Kimi-K2} & \avgmqm{17}{32.1}{2.0} & \avgmqm{27}{51.4}{4.8} & \avgmqm{12}{22.5}{4.1} & \avgmqm{16}{30.5}{5.6} & \avgmqm{13}{24.6}{3.8} & \avgmqm{17}{31.5}{4.3} \\
\modelcell{Qwen3-235B (Instruct)} & \avgmqm{17}{31.9}{2.3} & \avgmqm{22}{41.7}{5.2} & \avgmqm{14}{26.0}{5.6} & \avgmqm{18}{33.3}{5.6} & \avgmqm{17}{32.3}{5.2} & \avgmqm{14}{26.0}{4.1} \\
\modelcell{Qwen3.5-35B\\(Instruct)} & \avgmqm{13}{24.6}{2.0} & \avgmqm{24}{44.8}{6.2} & \avgmqm{9}{17.7}{4.0} & \avgmqm{12}{21.9}{3.9} & \avgmqm{11}{19.8}{3.7} & \avgmqm{10}{18.8}{4.4} \\
\modelcell{MiMo-V2-Flash} & \avgmqm{13}{24.4}{2.2} & \avgmqm{21}{39.6}{5.9} & \avgmqm{8}{15.6}{3.0} & \avgmqm{12}{22.9}{5.6} & \avgmqm{7}{13.5}{4.0} & \avgmqm{16}{30.2}{5.6} \\
\modelcell{Qwen3-30B (Instruct)} & \avgmqm{7}{14.0}{1.3} & \avgmqm{13}{25.0}{2.9} & \avgmqm{4}{7.3}{2.8} & \avgmqm{6}{11.5}{2.7} & \avgmqm{7}{12.5}{2.5} & \avgmqm{7}{13.5}{4.0} \\
\modelcell{Nemotron-3-Nano} & \avgmqm{3}{5.2}{0.8} & \avgmqm{4}{8.3}{1.7} & \avgmqm{3}{5.2}{1.8} & \avgmqm{1}{1.0}{1.0} & \avgmqm{3}{5.2}{2.7} & \avgmqm{3}{6.2}{1.8} \\
\midrule
\textbf{Average} & \cellcolor{HeatColor!22}41.3 & \cellcolor{HeatColor!29}54.8 & \cellcolor{HeatColor!19}35.3 & \cellcolor{HeatColor!21}39.2 & \cellcolor{HeatColor!20}37.5 & \cellcolor{HeatColor!21}39.9 \\
\bottomrule
\end{tabular}}
\label{tab:hard_cases_heatmap}
\vspace*{-0.15cm}
\end{table}
\endgroup

\textbf{Results.} Table~\ref{tab:hard_cases_heatmap} reports category-wise $MQM_{\geq 80}$ scores on the robustness subset. Idioms \& Cultural Metaphors emerge as the most challenging category (35.3\% on average), consistent with the difficulty of preserving figurative meaning across typologically diverse languages, whereas Conceptual \& Abstract Nuance is the easiest (54.8\%), suggesting that frontier models handle metaphorical extension more reliably than fixed idiomatic expressions. Although these items were specifically designed to stress classical failure modes of round-trip translation, model rankings remain highly consistent with the main LiT benchmark: Table~\ref{tab:lit_robustness_category_spearman} in Appendix~\ref{app:robustness_benchmark_details} shows that the Spearman correlation between each LiT category and the robustness benchmark average is significant at $p<0.001$ for all categories, ranging from $\rho=0.78$ for Informal to $\rho=0.96$ for Core Semantics. As in the main benchmark, we report mean performance together with bootstrap-estimated standard errors. 

Manual inspection helps explain this robustness: when an intermediate translation is awkward or overly literal, stronger models often preserve that awkwardness rather than correct it, because they follow the instruction to translate faithfully. In this sense, instruction-following models can faithfully propagate errors instead of repairing them, suggesting that round-trip translation is less vulnerable to some classical critiques and can still probe whether models genuinely understand idiomatic or structurally complex constructions.

\textbf{Implications.} These results suggest that round-trip translation, when applied to instruction-following LLMs, largely  overcome the classical criticisms against it. The benchmark successfully surfaces genuine cross-lingual weaknesses (e.g., idiom handling, register preservation) rather than being artificially derailed by the phenomena it was designed to probe. The high rank-order consistency with the main benchmark (Table~\ref{tab:lit_robustness_category_spearman} in Appendix~\ref{app:robustness_benchmark_details}) further validates that robustness cases do not introduce a separate, orthogonal axis of difficulty, but rather stress-test the same underlying multilingual generation capabilities measured by the main LiT benchmark.
\section{Discussion and Future Work}
We next discuss benchmark-design choices and considerations for future work.
\subsection{Benchmark Design} 
While serial language sequences inherently compound translation errors, we deliberately designed this mechanism to strictly stress-test a model's multilingual capabilities. A capable multilingual model must maintain semantic integrity across multiple translation hops; the catastrophic degradation observed outside high-resource sequences successfully isolates the boundary of this capability. By evaluating 200 highly dense, paragraph-length source texts \citep{laubli-etal-2018-machine}, LiT prioritizes linguistic depth over superficial breadth to prevent benchmark saturation \citep{bowman-dahl-2021-will}. This provides an efficient evaluation setting \citep{wu2025bitter} that probes linguistic phenomena often missed by sentence-level benchmarks.

\subsection{Statistical Methodology and Evidence}
We report rank correlations across six same-tier frontier configurations ($n{=}6$), constrained as described in Section~\ref{subsec:user-pref} to avoid scale-driven confounding. We therefore interpret the correlations together with three additional pieces of evidence: (1) across all three model families, Thinking variants significantly outperform on existing benchmarks yet perform comparably or worse on LMArena, with effect sizes exceeding 20 percentage points on MT-AIME24; (2) our error taxonomy (Section~\ref{subsec:err_tax}) independently shows benchmark failures are logical and factual, not linguistic; and (3) the reasoning-language analysis (Figure~\ref{fig:reas_lang}) confirms models default to English reasoning regardless of input language. We present $\rho$ values for directional contrast, not as standalone hypothesis tests.

Finally, LMArena remains the main large-scale source of multilingual human-preference data \citep{chiang2024chatbot}, which constrains the set of models with reliable cross-lingual ratings. Future work can strengthen the quantitative evidence as multilingual evaluation platforms expand.

\subsection{Automated Evaluation and Validity} 
To keep LiT scalable and reproducible, we use an LLM-as-a-judge framework. While this removes direct human-in-the-loop verification, recent literature demonstrates the effectiveness of this approach \citep{kocmi-federmann-2023-gemba, lu-etal-2024-error}, strongly correlating with human preference. Crucially, we validate the quality of our automated judge by demonstrating its high rank-order correlation with real-world human preference ratings from LMArena \citep{chiang2024chatbot}. This confirms that the semantic degradation penalized by our round-trip translation approach strictly aligns with the cross-lingual generation failures penalized by actual multilingual users.

\subsection{Limitations and Future Directions} While LiT resolves critical limitations in existing multilingual benchmarks, it is merely the first step toward more comprehensive evaluation paradigms. We highlight three high-impact directions for future work which could address current limitations:

\textbf{Scaling to Document-Level Discourse.} Frontier models are increasingly deployed for full-document tasks involving reports, literature, and legal text. While LiT effectively evaluates paragraph-level pragmatics, document-scale translation introduces complex global dependencies. Future benchmarks must evaluate the preservation of long-range referential chains, stylistic consistency, and persistent terminology memory across massive context windows.

\textbf{Disentangling Single Language Performance.} While our serial language sequences bound overall cross-lingual robustness, they also obscure single-language performance. To isolate exact failure modes, future work should separate these sequences into independent, controlled evaluations. This would enable more granular capability profiles for individual low-resource languages without the relatively fuzzy intermediate error cascading.

\textbf{Efficient and Culturally Grounded Generation.} An important direction is to develop efficient proxies that better reflect real-world human utility. While round-trip translation serves as an exceptional proxy for semantic preservation, future evaluation suites must expand beyond translation entirely. Developing highly efficient, native generation tasks that evaluate cultural grounding and target-language fluency, without relying on an English source or pivot, will be critical. Ultimately, combining round-trip evaluation with such native generative tasks will yield a comprehensive multilingual suite, moving the needle for the billions of users who rely on frontier models for global knowledge access and digitization.

\section{Conclusion}

In this work, we asked a simple question: Do current multilingual benchmarks faithfully measure multilingual capability? Benchmarks like MT-AIME24 and INCLUDE, used by frontier models to claim multilingual progress, actually might be confounded by improving reasoning and factual recall performance.
Two findings support this conclusion. First, thinking variants dramatically outperform instruct variants on these benchmarks, yet perform no better (and often worse) on LMArena, where real users rate multilingual outputs. Second, our error analysis reveals that failures are logical and factual, not linguistic: models parse translated questions correctly but fail to solve them. We bring back round-trip translation as an alternative method. Unlike existing benchmarks, performance on our LiT benchmark correlates positively with user preferences on LMArena. It spans high-resource to low-resource languages and shows underperformance of reasoning models in informal settings as well as substantial degradation on low-resource languages. We show that round-trip translation is a robust and scalable reference-free method for evaluating current multilingual capabilities of frontier models. We hope this work contributes to multilingual benchmarks that better measure preservation of meaning, fluency, and cultural context across languages.





\section*{Acknowledgements}

RS acknowledges funding by the Federal Ministry of Research, Technology and Space (BMFTR), FKZ: 16IS24079A. AP and MB acknowledge financial support by Federal Ministry of Research, Technology and Space (BMFTR) FKZ: 16IS24085B and Open Philanthropy Foundation funded by the Good Ventures Foundation.  MB is a member of the Machine Learning Cluster of Excellence, EXC number 2064/1 – Project number 390727645.

\bibliography{colm2026_conference}

@misc{yang2025qwen3technicalreport,
      title={Qwen3 Technical Report}, 
      author={An Yang and Anfeng Li and Baosong Yang and Beichen Zhang and Binyuan Hui and Bo Zheng and Bowen Yu and Chang Gao and Chengen Huang and Chenxu Lv and Chujie Zheng and Dayiheng Liu et al.},
      year={2025},
      eprint={2505.09388},
      archivePrefix={arXiv},
      primaryClass={cs.CL},
      url={https://arxiv.org/abs/2505.09388}, 
}

@inproceedings{
shi2023language,
title={Language models are multilingual chain-of-thought reasoners},
author={Freda Shi and Mirac Suzgun and Markus Freitag and Xuezhi Wang and Suraj Srivats and Soroush Vosoughi and Hyung Won Chung and Yi Tay and Sebastian Ruder and Denny Zhou and Dipanjan Das and Jason Wei},
booktitle={The Eleventh International Conference on Learning Representations },
year={2023},
url={https://openreview.net/forum?id=fR3wGCk-IXp}
}

@inproceedings{
hendrycks2021measuring,
title={Measuring Massive Multitask Language Understanding},
author={Dan Hendrycks and Collin Burns and Steven Basart and Andy Zou and Mantas Mazeika and Dawn Song and Jacob Steinhardt},
booktitle={International Conference on Learning Representations},
year={2021},
url={https://openreview.net/forum?id=d7KBjmI3GmQ}
}

@article{ni2024mixeval,
  title={Mixeval: Deriving wisdom of the crowd from llm benchmark mixtures},
  author={Ni, Jinjie and Xue, Fuzhao and Yue, Xiang and Deng, Yuntian and Shah, Mahir and Jain, Kabir and Neubig, Graham and You, Yang},
  journal={Advances in Neural Information Processing Systems},
  volume={37},
  pages={98180--98212},
  year={2024}
}

@inproceedings{chiang2024chatbot,
  title={Chatbot arena: An open platform for evaluating llms by human preference},
  author={Chiang, Wei-Lin and Zheng, Lianmin and Sheng, Ying and Angelopoulos, Anastasios Nikolas and Li, Tianle and Li, Dacheng and Zhu, Banghua and Zhang, Hao and Jordan, Michael and Gonzalez, Joseph E and others},
  booktitle={Forty-first International Conference on Machine Learning},
  year={2024}
}

@inproceedings{son-etal-2025-linguistic,
    title = "Linguistic Generalizability of Test-Time Scaling in Mathematical Reasoning",
    author = "Son, Guijin  and
      Hong, Jiwoo  and
      Ko, Hyunwoo  and
      Thorne, James",
    booktitle = "Proceedings of the 63rd Annual Meeting of the Association for Computational Linguistics (Volume 1: Long Papers)",
    month = jul,
    year = "2025",
    publisher = "Association for Computational Linguistics",
    url = "https://aclanthology.org/2025.acl-long.699/",
    doi = "10.18653/v1/2025.acl-long.699",
}

@inproceedings{
wang2025polymath,
title={PolyMath: Evaluating Mathematical Reasoning in Multilingual Contexts},
author={Yiming Wang and Pei Zhang and Jialong Tang and Hao-Ran Wei and Baosong Yang and Rui Wang and Chenshu Sun and Feitong Sun and Jiran Zhang and Junxuan Wu and Qiqian Cang and Yichang Zhang and Fei Huang and Junyang Lin and Fei Huang and Jingren Zhou},
booktitle={The Thirty-ninth Annual Conference on Neural Information Processing Systems Datasets and Benchmarks Track},
year={2025},
url={https://openreview.net/forum?id=B1vCImy6yI}
}

@article{wu2025bitter,
  title={The bitter lesson learned from 2,000+ multilingual benchmarks},
  author={Wu, Minghao and Wang, Weixuan and Liu, Sinuo and Yin, Huifeng and Wang, Xintong and Zhao, Yu and Lyu, Chenyang and Wang, Longyue and Luo, Weihua and Zhang, Kaifu},
  journal={arXiv preprint arXiv:2504.15521},
  year={2025}
}

@inproceedings{
romanou2025include,
title={{INCLUDE}: Evaluating Multilingual Language Understanding with Regional Knowledge},
author={Angelika Romanou and Negar Foroutan and Anna Sotnikova and Sree Harsha Nelaturu and Shivalika Singh and Rishabh Maheshwary and Micol Altomare and Zeming Chen and Mohamed A. Haggag and Snegha A and Alfonso Amayuelas and Azril Hafizi et al.},
booktitle={The Thirteenth International Conference on Learning Representations},
year={2025},
url={https://openreview.net/forum?id=k3gCieTXeY}
}

@inproceedings{singh-etal-2025-global,
    title = "Global {MMLU}: Understanding and Addressing Cultural and Linguistic Biases in Multilingual Evaluation",
    author = "Singh, Shivalika  and
      Romanou, Angelika  and
      Fourrier, Cl{\'e}mentine  and
      Adelani, David Ifeoluwa  and
      Ngui, Jian Gang  and
      Vila-Suero, Daniel  and
      Limkonchotiwat, Peerat  and
      Marchisio, Kelly  and
      Leong, Wei Qi  et al.",
    booktitle = "Proceedings of the 63rd Annual Meeting of the Association for Computational Linguistics (Volume 1: Long Papers)",
    year = "2025",
    publisher = "Association for Computational Linguistics",
    url = "https://aclanthology.org/2025.acl-long.919/",
    doi = "10.18653/v1/2025.acl-long.919",
}

@article{lommel2014multidimensional,
  title={Multidimensional quality metrics (MQM): A framework for declaring and describing translation quality metrics},
  author={Lommel, Arle and Uszkoreit, Hans and Burchardt, Aljoscha},
  journal={Tradum{\`a}tica},
  number={12},
  pages={0455--463},
  year={2014}
}

@misc{5team2025glm45agenticreasoningcoding,
      title={GLM-4.5: Agentic, Reasoning, and Coding (ARC) Foundation Models}, 
      author={Z.ai and 5 Team and Aohan Zeng and Xin Lv and Qinkai Zheng and Zhenyu Hou and Bin Chen and Chengxing Xie and Cunxiang Wang and Da Yin and Hao Zeng and Jiajie Zhang and Kedong Wang and Lucen Zhong and Mingdao Liu and Rui Lu and Shulin Cao and Xiaohan Zhang and Xuancheng Huang and Yao Wei and Yean Cheng and Yifan An and Yilin Niu and Yuanhao Wen and Yushi Bai and Zhengxiao Du and Zihan Wang and Zilin Zhu and Bohan Zhang and Bosi Wen and Bowen Wu and Bowen Xu and Can Huang and Casey Zhao and Changpeng Cai and Chao Yu and Chen Li and Chendi Ge and Chenghua Huang and Chenhui Zhang and Chenxi Xu and Chenzheng Zhu and Chuang Li and Congfeng Yin and Daoyan Lin and Dayong Yang and Dazhi Jiang and Ding Ai and Erle Zhu and Fei Wang and Gengzheng Pan and Guo Wang and Hailong Sun and Haitao Li and Haiyang Li and Haiyi Hu and Hanyu Zhang and Hao Peng and Hao Tai and Haoke Zhang and Haoran Wang and Haoyu Yang and He Liu and He Zhao and Hongwei Liu and Hongxi Yan and Huan Liu and Huilong Chen and Ji Li and Jiajing Zhao and Jiamin Ren and Jian Jiao and Jiani Zhao and Jianyang Yan and Jiaqi Wang and Jiayi Gui and Jiayue Zhao and Jie Liu and Jijie Li and Jing Li and Jing Lu and Jingsen Wang and Jingwei Yuan and Jingxuan Li and Jingzhao Du and Jinhua Du and Jinxin Liu and Junkai Zhi and Junli Gao and Ke Wang and Lekang Yang and Liang Xu and Lin Fan and Lindong Wu and Lintao Ding and Lu Wang and Man Zhang and Minghao Li and Minghuan Xu and Mingming Zhao and Mingshu Zhai and Pengfan Du and Qian Dong and Shangde Lei and Shangqing Tu and Shangtong Yang and Shaoyou Lu and Shijie Li and Shuang Li and Shuang-Li and Shuxun Yang and Sibo Yi and Tianshu Yu and Wei Tian and Weihan Wang and Wenbo Yu and Weng Lam Tam and Wenjie Liang and Wentao Liu and Xiao Wang and Xiaohan Jia and Xiaotao Gu and Xiaoying Ling and Xin Wang and Xing Fan and Xingru Pan and Xinyuan Zhang and Xinze Zhang and Xiuqing Fu and Xunkai Zhang and Yabo Xu and Yandong Wu and Yida Lu and Yidong Wang and Yilin Zhou and Yiming Pan and Ying Zhang and Yingli Wang and Yingru Li and Yinpei Su and Yipeng Geng and Yitong Zhu and Yongkun Yang and Yuhang Li and Yuhao Wu and Yujiang Li and Yunan Liu and Yunqing Wang and Yuntao Li and Yuxuan Zhang and Zezhen Liu and Zhen Yang and Zhengda Zhou and Zhongpei Qiao and Zhuoer Feng and Zhuorui Liu and Zichen Zhang and Zihan Wang and Zijun Yao and Zikang Wang and Ziqiang Liu and Ziwei Chai and Zixuan Li and Zuodong Zhao and Wenguang Chen and Jidong Zhai and Bin Xu and Minlie Huang and Hongning Wang and Juanzi Li and Yuxiao Dong and Jie Tang},
      year={2025},
      eprint={2508.06471},
      archivePrefix={arXiv},
      primaryClass={cs.CL},
      url={https://arxiv.org/abs/2508.06471}, 
}

@inproceedings{sennrich2016improving,
  title={Improving neural machine translation models with monolingual data},
  author={Sennrich, Rico and Haddow, Barry and Birch, Alexandra},
  booktitle={Proceedings of the 54th annual meeting of the association for computational linguistics (volume 1: long papers)},
  pages={86--96},
  year={2016}
}

@article{brislin1970back,
  title={Back-translation for cross-cultural research},
  author={Brislin, Richard W},
  journal={Journal of cross-cultural psychology},
  volume={1},
  number={3},
  pages={185--216},
  year={1970},
  publisher={Sage Publications Sage CA: Thousand Oaks, CA}
}

@article{liu2025deepseek,
  title={Deepseek-v3. 2: Pushing the frontier of open large language models},
  author={Liu, Aixin and Mei, Aoxue and Lin, Bangcai and Xue, Bing and Wang, Bingxuan and Xu, Bingzheng and Wu, Bochao and Zhang, Bowei and Lin, Chaofan and Dong, Chen and others},
  journal={arXiv preprint arXiv:2512.02556},
  year={2025}
}

@inproceedings{kocmi2025findings,
  title={Findings of the wmt25 general machine translation shared task: Time to stop evaluating on easy test sets},
  author={Kocmi, Tom and Artemova, Ekaterina and Avramidis, Eleftherios and Bawden, Rachel and Bojar, Ond{\v{r}}ej and Dranch, Konstantin and Dvorkovich, Anton and Dukanov, Sergey and Fishel, Mark and Freitag, Markus and others},
  booktitle={Proceedings of the Tenth Conference on Machine Translation},
  pages={355--413},
  year={2025}
}

@article{team2025kimi,
  title={Kimi k2: Open agentic intelligence},
  author={Team, Kimi and Bai, Yifan and Bao, Yiping and Chen, Guanduo and Chen, Jiahao and Chen, Ningxin and Chen, Ruijue and Chen, Yanru and Chen, Yuankun and Chen, Yutian and others},
  journal={arXiv preprint arXiv:2507.20534},
  year={2025}
}

@inproceedings{somers2005round,
  title={Round-trip translation: What is it good for?},
  author={Somers, Harold},
  booktitle={Proceedings of the Australasian Language Technology Workshop 2005},
  pages={127--133},
  year={2005}
}

@inproceedings{park-etal-2024-multiprageval,
    title = "{M}ulti{P}rag{E}val: Multilingual Pragmatic Evaluation of Large Language Models",
    author = "Park, Dojun  and
      Lee, Jiwoo  and
      Park, Seohyun  and
      Jeong, Hyeyun  and
      Koo, Youngeun  and
      Hwang, Soonha  and
      Park, Seonwoo  and
      Lee, Sungeun",
    editor = "Hupkes, Dieuwke  and
      Dankers, Verna  and
      Batsuren, Khuyagbaatar  and
      Kazemnejad, Amirhossein  and
      Christodoulopoulos, Christos  and
      Giulianelli, Mario  and
      Cotterell, Ryan",
    booktitle = "Proceedings of the 2nd GenBench Workshop on Generalisation (Benchmarking) in NLP",
    month = nov,
    year = "2024",
    address = "Miami, Florida, USA",
    publisher = "Association for Computational Linguistics",
    url = "https://aclanthology.org/2024.genbench-1.7/",
    doi = "10.18653/v1/2024.genbench-1.7",
    pages = "96--119"
}

@inproceedings{yao-etal-2024-benchmarking,
    title = "Benchmarking Machine Translation with Cultural Awareness",
    author = "Yao, Binwei  and
      Jiang, Ming  and
      Bobinac, Tara  and
      Yang, Diyi  and
      Hu, Junjie",
    editor = "Al-Onaizan, Yaser  and
      Bansal, Mohit  and
      Chen, Yun-Nung",
    booktitle = "Findings of the Association for Computational Linguistics: EMNLP 2024",
    month = nov,
    year = "2024",
    address = "Miami, Florida, USA",
    publisher = "Association for Computational Linguistics",
    url = "https://aclanthology.org/2024.findings-emnlp.765/",
    doi = "10.18653/v1/2024.findings-emnlp.765",
    pages = "13078--13096"
}

@inproceedings{taguchi-etal-2025-languages,
    title = "Languages Still Left Behind: Toward a Better Multilingual Machine Translation Benchmark",
    author = "Taguchi, Chihiro  and
      Mai, Seng  and
      Kurabe, Keita  and
      Sakai, Yusuke  and
      Agyei, Georgina  and
      Eslami, Soudabeh  and
      Chiang, David",
    editor = "Christodoulopoulos, Christos  and
      Chakraborty, Tanmoy  and
      Rose, Carolyn  and
      Peng, Violet",
    booktitle = "Proceedings of the 2025 Conference on Empirical Methods in Natural Language Processing",
    month = nov,
    year = "2025",
    address = "Suzhou, China",
    publisher = "Association for Computational Linguistics",
    url = "https://aclanthology.org/2025.emnlp-main.1018/",
    doi = "10.18653/v1/2025.emnlp-main.1018",
    pages = "20131--20143",
    ISBN = "979-8-89176-332-6"
}

@inproceedings{zhuo-etal-2023-rethinking,
    title = "Rethinking Round-Trip Translation for Machine Translation Evaluation",
    author = "Zhuo, Terry Yue  and
      Xu, Qiongkai  and
      He, Xuanli  and
      Cohn, Trevor",
    editor = "Rogers, Anna  and
      Boyd-Graber, Jordan  and
      Okazaki, Naoaki",
    booktitle = "Findings of the Association for Computational Linguistics: ACL 2023",
    month = jul,
    year = "2023",
    address = "Toronto, Canada",
    publisher = "Association for Computational Linguistics",
    url = "https://aclanthology.org/2023.findings-acl.22/",
    doi = "10.18653/v1/2023.findings-acl.22",
    pages = "319--337"
}

@article{evaluate_nlg,
author = {Sai, Ananya B. and Mohankumar, Akash Kumar and Khapra, Mitesh M.},
title = {A Survey of Evaluation Metrics Used for NLG Systems},
year = {2022},
issue_date = {February 2023},
publisher = {Association for Computing Machinery},
address = {New York, NY, USA},
volume = {55},
number = {2},
issn = {0360-0300},
url = {https://doi.org/10.1145/3485766},
doi = {10.1145/3485766},
journal = {ACM Comput. Surv.},
month = jan,
articleno = {26},
numpages = {39}
}

@inproceedings{freitag-etal-2021-results,
    title = "Results of the {WMT}21 Metrics Shared Task: Evaluating Metrics with Expert-based Human Evaluations on {TED} and News Domain",
    author = "Freitag, Markus  and
      Rei, Ricardo  and
      Mathur, Nitika  and
      Lo, Chi-kiu  and
      Stewart, Craig  and
      Foster, George  and
      Lavie, Alon  and
      Bojar, Ond{\v{r}}ej",
    editor = "Barrault, Loic  and
      Bojar, Ondrej  and
      Bougares, Fethi  and
      Chatterjee, Rajen  and
      Costa-jussa, Marta R.  and
      Federmann, Christian  and
      Fishel, Mark  and
      Fraser, Alexander  and
      Freitag, Markus  and
      Graham, Yvette  and
      Grundkiewicz, Roman  and
      Guzman, Paco  and
      Haddow, Barry  and
      Huck, Matthias  and
      Yepes, Antonio Jimeno  and
      Koehn, Philipp  and
      Kocmi, Tom  and
      Martins, Andre  and
      Morishita, Makoto  and
      Monz, Christof",
    booktitle = "Proceedings of the Sixth Conference on Machine Translation",
    month = nov,
    year = "2021",
    address = "Online",
    publisher = "Association for Computational Linguistics",
    url = "https://aclanthology.org/2021.wmt-1.73/",
    pages = "733--774"
}

@inproceedings{lavie-etal-2025-findings,
    title = "Findings of the {WMT}25 Shared Task on Automated Translation Evaluation Systems: Linguistic Diversity is Challenging and References Still Help",
    author = "Lavie, Alon  and
      Hanneman, Greg  and
      Agrawal, Sweta  and
      Kanojia, Diptesh  and
      Lo, Chi-Kiu  and
      Zouhar, Vil{\'e}m  and
      Blain, Frederic  and
      Zerva, Chrysoula  and
      Avramidis, Eleftherios  and
      Deoghare, Sourabh  and
      Sindhujan, Archchana  and
      Wang, Jiayi  and
      Adelani, David Ifeoluwa  and
      Thompson, Brian  and
      Kocmi, Tom  and
      Freitag, Markus  and
      Deutsch, Daniel",
    editor = "Haddow, Barry  and
      Kocmi, Tom  and
      Koehn, Philipp  and
      Monz, Christof",
    booktitle = "Proceedings of the Tenth Conference on Machine Translation",
    month = nov,
    year = "2025",
    address = "Suzhou, China",
    publisher = "Association for Computational Linguistics",
    url = "https://aclanthology.org/2025.wmt-1.24/",
    doi = "10.18653/v1/2025.wmt-1.24",
    pages = "436--483",
    ISBN = "979-8-89176-341-8"
}

@article{data_cont,
  author       = {Oscar Sainz and
                  Iker Garc{\'{\i}}a{-}Ferrero and
                  Alon Jacovi and
                  Jon Ander Campos and
                  Yanai Elazar and
                  Eneko Agirre and
                  Yoav Goldberg and
                  Wei{-}Lin Chen and
                  Jenny Chim and
                  Leshem Choshen and
                  Luca D'Amico{-}Wong and
                  Melissa Dell and
                  Run{-}Ze Fan and
                  Shahriar Golchin and
                  Yucheng Li and
                  Pengfei Liu and
                  Bhavish Pahwa and
                  Ameya Prabhu and
                  Suryansh Sharma and
                  Emily Silcock and
                  Kateryna Solonko and
                  David Stap and
                  Mihai Surdeanu and
                  Yu{-}Min Tseng and
                  Vishaal Udandarao and
                  Zengzhi Wang and
                  Ruijie Xu and
                  Jinglin Yang},
  title        = {Data Contamination Report from the 2024 {CONDA} Shared Task},
  journal      = {CoRR},
  volume       = {abs/2407.21530},
  year         = {2024},
  url          = {https://doi.org/10.48550/arXiv.2407.21530},
  doi          = {10.48550/ARXIV.2407.21530},
  eprinttype    = {arXiv},
  eprint       = {2407.21530},
  bibsource    = {dblp computer science bibliography, https://dblp.org}
}

@inproceedings{vilar-etal-2023-prompting,
    title = "Prompting {P}a{LM} for Translation: Assessing Strategies and Performance",
    author = "Vilar, David  and
      Freitag, Markus  and
      Cherry, Colin  and
      Luo, Jiaming  and
      Ratnakar, Viresh  and
      Foster, George",
    editor = "Rogers, Anna  and
      Boyd-Graber, Jordan  and
      Okazaki, Naoaki",
    booktitle = "Proceedings of the 61st Annual Meeting of the Association for Computational Linguistics (Volume 1: Long Papers)",
    month = jul,
    year = "2023",
    address = "Toronto, Canada",
    publisher = "Association for Computational Linguistics",
    url = "https://aclanthology.org/2023.acl-long.859/",
    doi = "10.18653/v1/2023.acl-long.859",
    pages = "15406--15427"
}

@inproceedings{karpinska-iyyer-2023-large,
    title = "Large Language Models Effectively Leverage Document-level Context for Literary Translation, but Critical Errors Persist",
    author = "Karpinska, Marzena  and
      Iyyer, Mohit",
    editor = "Koehn, Philipp  and
      Haddow, Barry  and
      Kocmi, Tom  and
      Monz, Christof",
    booktitle = "Proceedings of the Eighth Conference on Machine Translation",
    month = dec,
    year = "2023",
    address = "Singapore",
    publisher = "Association for Computational Linguistics",
    url = "https://aclanthology.org/2023.wmt-1.41/",
    doi = "10.18653/v1/2023.wmt-1.41",
    pages = "419--451"
}

@inproceedings{kim-2025-rubric,
    title = "{RUBRIC}-{MQM} : Span-Level {LLM}-as-judge in Machine Translation For High-End Models",
    author = "Kim, Ahrii",
    editor = "Rehm, Georg  and
      Li, Yunyao",
    booktitle = "Proceedings of the 63rd Annual Meeting of the Association for Computational Linguistics (Volume 6: Industry Track)",
    month = jul,
    year = "2025",
    address = "Vienna, Austria",
    publisher = "Association for Computational Linguistics",
    url = "https://aclanthology.org/2025.acl-industry.12/",
    doi = "10.18653/v1/2025.acl-industry.12",
    pages = "147--165",
    ISBN = "979-8-89176-288-6"
}

@inproceedings{lu-etal-2024-error,
    title = "Error Analysis Prompting Enables Human-Like Translation Evaluation in Large Language Models",
    author = "Lu, Qingyu  and
      Qiu, Baopu  and
      Ding, Liang  and
      Zhang, Kanjian  and
      Kocmi, Tom  and
      Tao, Dacheng",
    editor = "Ku, Lun-Wei  and
      Martins, Andre  and
      Srikumar, Vivek",
    booktitle = "Findings of the Association for Computational Linguistics: ACL 2024",
    month = aug,
    year = "2024",
    address = "Bangkok, Thailand",
    publisher = "Association for Computational Linguistics",
    url = "https://aclanthology.org/2024.findings-acl.520/",
    doi = "10.18653/v1/2024.findings-acl.520",
    pages = "8801--8816"
}

@inproceedings{kocmi-federmann-2023-gemba,
    title = "{GEMBA}-{MQM}: Detecting Translation Quality Error Spans with {GPT}-4",
    author = "Kocmi, Tom  and
      Federmann, Christian",
    editor = "Koehn, Philipp  and
      Haddow, Barry  and
      Kocmi, Tom  and
      Monz, Christof",
    booktitle = "Proceedings of the Eighth Conference on Machine Translation",
    month = dec,
    year = "2023",
    address = "Singapore",
    publisher = "Association for Computational Linguistics",
    url = "https://aclanthology.org/2023.wmt-1.64/",
    doi = "10.18653/v1/2023.wmt-1.64",
    pages = "768--775"
}

@inproceedings{
li2025from,
title={From Crowdsourced Data to High-quality Benchmarks: Arena-Hard and Benchbuilder Pipeline},
author={Tianle Li and Wei-Lin Chiang and Evan Frick and Lisa Dunlap and Tianhao Wu and Banghua Zhu and Joseph E. Gonzalez and Ion Stoica},
booktitle={Forty-second International Conference on Machine Learning},
year={2025},
url={https://openreview.net/forum?id=KfTf9vFvSn}
}

@inproceedings{
dubois2024lengthcontrolled,
title={Length-Controlled AlpacaEval: A Simple Debiasing of Automatic Evaluators},
author={Yann Dubois and Percy Liang and Tatsunori Hashimoto},
booktitle={First Conference on Language Modeling},
year={2024},
url={https://openreview.net/forum?id=CybBmzWBX0}
}

@inproceedings{spangher-etal-2025-chatbot,
    title = "Chatbot Arena Estimate: towards a generalized performance benchmark for {LLM} capabilities",
    author = "Spangher, Lucas  and
      Li, Tianle  and
      Arnold, William F.  and
      Masiewicki, Nick  and
      Dotiwalla, Xerxes  and
      Pasumarthi, Rama Kumar  and
      Grabowski, Peter  and
      Ie, Eugene  and
      Gruhl, Daniel",
    editor = "Chen, Weizhu  and
      Yang, Yi  and
      Kachuee, Mohammad  and
      Fu, Xue-Yong",
    booktitle = "Proceedings of the 2025 Conference of the Nations of the Americas Chapter of the Association for Computational Linguistics: Human Language Technologies (Volume 3: Industry Track)",
    month = apr,
    year = "2025",
    address = "Albuquerque, New Mexico",
    publisher = "Association for Computational Linguistics",
    url = "https://aclanthology.org/2025.naacl-industry.77/",
    doi = "10.18653/v1/2025.naacl-industry.77",
    pages = "1016--1025",
    ISBN = "979-8-89176-194-0"
}

@inproceedings{moon-etal-2020-revisiting,
    title = "Revisiting Round-trip Translation for Quality Estimation",
    author = "Moon, Jihyung  and
      Cho, Hyunchang  and
      Park, Eunjeong L.",
    editor = "Martins, Andr{\'e}  and
      Moniz, Helena  and
      Fumega, Sara  and
      Martins, Bruno  and
      Batista, Fernando  and
      Coheur, Luisa  and
      Parra, Carla  and
      Trancoso, Isabel  and
      Turchi, Marco  and
      Bisazza, Arianna  and
      Moorkens, Joss  and
      Guerberof, Ana  and
      Nurminen, Mary  and
      Marg, Lena  and
      Forcada, Mikel L.",
    booktitle = "Proceedings of the 22nd Annual Conference of the European Association for Machine Translation",
    month = nov,
    year = "2020",
    address = "Lisboa, Portugal",
    publisher = "European Association for Machine Translation",
    url = "https://aclanthology.org/2020.eamt-1.11/",
    pages = "91--104"
}

@inproceedings{rtt_corr,
author = {Allamanis, Miltiadis and Panthaplackel, Sheena and Yin, Pengcheng},
title = {Unsupervised evaluation of code LLMs with round-trip correctness},
year = {2024},
publisher = {JMLR.org},
booktitle = {Proceedings of the 41st International Conference on Machine Learning},
articleno = {44},
numpages = {17},
location = {Vienna, Austria},
series = {ICML'24}
}

@inproceedings{
zheng2024large,
title={Large Language Models Are Not Robust Multiple Choice Selectors},
author={Chujie Zheng and Hao Zhou and Fandong Meng and Jie Zhou and Minlie Huang},
booktitle={The Twelfth International Conference on Learning Representations},
year={2024},
url={https://openreview.net/forum?id=shr9PXz7T0}
}

@inproceedings{balepur-etal-2025-best,
    title = "Which of These Best Describes Multiple Choice Evaluation with {LLM}s? A) Forced {B}) Flawed {C}) Fixable {D}) All of the Above",
    author = "Balepur, Nishant  and
      Rudinger, Rachel  and
      Boyd-Graber, Jordan Lee",
    editor = "Che, Wanxiang  and
      Nabende, Joyce  and
      Shutova, Ekaterina  and
      Pilehvar, Mohammad Taher",
    booktitle = "Proceedings of the 63rd Annual Meeting of the Association for Computational Linguistics (Volume 1: Long Papers)",
    month = jul,
    year = "2025",
    address = "Vienna, Austria",
    publisher = "Association for Computational Linguistics",
    url = "https://aclanthology.org/2025.acl-long.169/",
    doi = "10.18653/v1/2025.acl-long.169",
    pages = "3394--3418",
    ISBN = "979-8-89176-251-0"
}

@inproceedings{
ghorbani2022scaling,
title={Scaling Laws for Neural Machine Translation},
author={Behrooz Ghorbani and Orhan Firat and Markus Freitag and Ankur Bapna and Maxim Krikun and Xavier Garcia and Ciprian Chelba and Colin Cherry},
booktitle={International Conference on Learning Representations},
year={2022},
url={https://openreview.net/forum?id=hR_SMu8cxCV}
}

@misc{kaplan2020scalinglawsneurallanguage,
      title={Scaling Laws for Neural Language Models}, 
      author={Jared Kaplan and Sam McCandlish and Tom Henighan and Tom B. Brown and Benjamin Chess and Rewon Child and Scott Gray and Alec Radford and Jeffrey Wu and Dario Amodei},
      year={2020},
      eprint={2001.08361},
      archivePrefix={arXiv},
      primaryClass={cs.LG},
      url={https://arxiv.org/abs/2001.08361}, 
}

@inproceedings{
wang2024mmlupro,
title={{MMLU}-Pro: A More Robust and Challenging Multi-Task Language Understanding Benchmark},
author={Yubo Wang and Xueguang Ma and Ge Zhang and Yuansheng Ni and Abhranil Chandra and Shiguang Guo and Weiming Ren and Aaran Arulraj and Xuan He and Ziyan Jiang and Tianle Li and Max Ku and Kai Wang and Alex Zhuang and Rongqi Fan and Xiang Yue and Wenhu Chen},
booktitle={The Thirty-eight Conference on Neural Information Processing Systems Datasets and Benchmarks Track},
year={2024},
url={https://openreview.net/forum?id=y10DM6R2r3}
}

@article{rte_dagan,
title = "Recognizing Textual Entailment: Models and Applications",
author = "Ido Dagan and Dan Roth and Mark Sammons and Fabio Zanzotto",
note = "Publisher Copyright: {\textcopyright} Morgan and Claypool Publishers. All rights reserved.",
year = "2013",
doi = "10.2200/S00509ED1V01Y201305HLT023",
language = "English (US)",
volume = "6",
pages = "1--222",
journal = "Synthesis Lectures on Human Language Technologies",
issn = "1947-4040",
publisher = "Springer International Publishing",
number = "4",
}

@inproceedings{10.3115/1219840.1219858,
author = {Barzilay, Regina and Lapata, Mirella},
title = {Modeling local coherence: an entity-based approach},
year = {2005},
publisher = {Association for Computational Linguistics},
address = {USA},
url = {https://doi.org/10.3115/1219840.1219858},
doi = {10.3115/1219840.1219858},
booktitle = {Proceedings of the 43rd Annual Meeting on Association for Computational Linguistics},
pages = {141–148},
numpages = {8},
location = {Ann Arbor, Michigan},
series = {ACL '05}
}

@article{grice1975logic,
  title={Logic and Conversation},
  author={Grice, H Paul},
  journal={Syntax and Semantics},
  volume={3},
  pages={41--58},
  year={1975},
  publisher={Academic Press}
}

@book{searle1969speech,
  title={Speech Acts: An Essay in the Philosophy of Language},
  author={Searle, John R},
  publisher={Cambridge University Press},
  year={1969}
}

@book{brown1987politeness,
  title={Politeness: Some Universals in Language Usage},
  author={Brown, Penelope and Levinson, Stephen C},
  publisher={Cambridge University Press},
  year={1987}
}

@inproceedings{isabelle-etal-2017-challenge,
    title = "A Challenge Set Approach to Evaluating Machine Translation",
    author = "Isabelle, Pierre  and
      Cherry, Colin  and
      Foster, George",
    editor = "Palmer, Martha  and
      Hwa, Rebecca  and
      Riedel, Sebastian",
    booktitle = "Proceedings of the 2017 Conference on Empirical Methods in Natural Language Processing",
    month = sep,
    year = "2017",
    address = "Copenhagen, Denmark",
    publisher = "Association for Computational Linguistics",
    url = "https://aclanthology.org/D17-1263/",
    doi = "10.18653/v1/D17-1263",
    pages = "2486--2496"
}

@misc{kleidermacher2025sciencelanguagesassessingllm,
      title={Science Across Languages: Assessing LLM Multilingual Translation of Scientific Papers}, 
      author={Hannah Calzi Kleidermacher and James Zou},
      year={2025},
      eprint={2502.17882},
      archivePrefix={arXiv},
      primaryClass={cs.AI},
      url={https://arxiv.org/abs/2502.17882}, 
}

@inproceedings{koehn-knowles-2017-six,
    title = "Six Challenges for Neural Machine Translation",
    author = "Koehn, Philipp  and
      Knowles, Rebecca",
    editor = "Luong, Thang  and
      Birch, Alexandra  and
      Neubig, Graham  and
      Finch, Andrew",
    booktitle = "Proceedings of the First Workshop on Neural Machine Translation",
    month = aug,
    year = "2017",
    address = "Vancouver",
    publisher = "Association for Computational Linguistics",
    url = "https://aclanthology.org/W17-3204/",
    doi = "10.18653/v1/W17-3204",
    pages = "28--39"
}

@inproceedings{fadaee-etal-2018-examining,
    title = "Examining the Tip of the Iceberg: A Data Set for Idiom Translation",
    author = "Fadaee, Marzieh  and
      Bisazza, Arianna  and
      Monz, Christof",
    editor = "Calzolari, Nicoletta  and
      Choukri, Khalid  and
      Cieri, Christopher  and
      Declerck, Thierry  and
      Goggi, Sara  and
      Hasida, Koiti  and
      Isahara, Hitoshi  and
      Maegaard, Bente  and
      Mariani, Joseph  and
      Mazo, H{\'e}l{\`e}ne  and
      Moreno, Asuncion  and
      Odijk, Jan  and
      Piperidis, Stelios  and
      Tokunaga, Takenobu",
    booktitle = "Proceedings of the Eleventh International Conference on Language Resources and Evaluation ({LREC} 2018)",
    month = may,
    year = "2018",
    address = "Miyazaki, Japan",
    publisher = "European Language Resources Association (ELRA)",
    url = "https://aclanthology.org/L18-1148/"
}

@inproceedings{farinha-etal-2022-findings,
    title = "Findings of the {WMT} 2022 Shared Task on Chat Translation",
    author = "Farinha, Ana C  and
      Farajian, M. Amin  and
      Buchicchio, Marianna  and
      Fernandes, Patrick  and
      C. de Souza, Jos{\'e} G.  and
      Moniz, Helena  and
      Martins, Andr{\'e} F. T.",
    editor = {Koehn, Philipp  and
      Barrault, Lo{\"i}c  and
      Bojar, Ond{\v{r}}ej  and
      Bougares, Fethi  and
      Chatterjee, Rajen  and
      Costa-juss{\`a}, Marta R.  and
      Federmann, Christian  and
      Fishel, Mark  and
      Fraser, Alexander  and
      Freitag, Markus  and
      Graham, Yvette  and
      Grundkiewicz, Roman  and
      Guzman, Paco  and
      Haddow, Barry  and
      Huck, Matthias  and
      Jimeno Yepes, Antonio  and
      Kocmi, Tom  and
      Martins, Andr{\'e}  and
      Morishita, Makoto  and
      Monz, Christof  and
      Nagata, Masaaki  and
      Nakazawa, Toshiaki  and
      Negri, Matteo  and
      N{\'e}v{\'e}ol, Aur{\'e}lie  and
      Neves, Mariana  and
      Popel, Martin  and
      Turchi, Marco  and
      Zampieri, Marcos},
    booktitle = "Proceedings of the Seventh Conference on Machine Translation (WMT)",
    month = dec,
    year = "2022",
    address = "Abu Dhabi, United Arab Emirates (Hybrid)",
    publisher = "Association for Computational Linguistics",
    url = "https://aclanthology.org/2022.wmt-1.70/",
    doi = "10.18653/v1/2022.wmt-1.70",
    pages = "724--743"
}

@inproceedings{lommel-etal-2024-multi,
    title = "The Multi-Range Theory of Translation Quality Measurement: {MQM} scoring models and Statistical Quality Control",
    author = "Lommel, Arle  and
      Gladkoff, Serge  and
      Melby, Alan  and
      Wright, Sue Ellen  and
      Strandvik, Ingemar  and
      Gasova, Katerina  and
      Vaasa, Angelika  and
      Benzo, Andy  and
      Marazzato Sparano, Romina  and
      Foresi, Monica  and
      Innis, Johani  and
      Han, Lifeng  and
      Nenadic, Goran",
    editor = "Martindale, Marianna  and
      Campbell, Janice  and
      Savenkov, Konstantin  and
      Goel, Shivali",
    booktitle = "Proceedings of the 16th Conference of the Association for Machine Translation in the Americas (Volume 2: Presentations)",
    month = sep,
    year = "2024",
    address = "Chicago, USA",
    publisher = "Association for Machine Translation in the Americas",
    url = "https://aclanthology.org/2024.amta-presentations.6/",
    pages = "75--94"
}

@online{qwen3_blog_2025,
  author = {{Qwen Team}},
  title = {Qwen3: Think Deeper, Act Faster},
  year = {2025},
  month = {April},
  url = {https://qwen.ai/blog?id=qwen3},
  urldate = {2026-03-04} 
}

@inproceedings{10.1007/11941439_149,
author = {van Zaanen, Menno and Zwarts, Simon},
title = {Unsupervised measurement of translation quality using multi-engine, bi-directional translation},
year = {2006},
isbn = {3540497870},
publisher = {Springer-Verlag},
address = {Berlin, Heidelberg},
url = {https://doi.org/10.1007/11941439_149},
doi = {10.1007/11941439_149},
booktitle = {Proceedings of the 19th Australian Joint Conference on Artificial Intelligence: Advances in Artificial Intelligence},
pages = {1208–1214},
numpages = {7},
location = {Hobart, Australia},
series = {AI'06}
}

@inproceedings{laubli-etal-2018-machine,
    title = "Has Machine Translation Achieved Human Parity? A Case for Document-level Evaluation",
    author = {L{\"a}ubli, Samuel  and
      Sennrich, Rico  and
      Volk, Martin},
    editor = "Riloff, Ellen  and
      Chiang, David  and
      Hockenmaier, Julia  and
      Tsujii, Jun{'}ichi",
    booktitle = "Proceedings of the 2018 Conference on Empirical Methods in Natural Language Processing",
    month = oct # "-" # nov,
    year = "2018",
    address = "Brussels, Belgium",
    publisher = "Association for Computational Linguistics",
    url = "https://aclanthology.org/D18-1512/",
    doi = "10.18653/v1/D18-1512",
    pages = "4791--4796"
}

@inproceedings{bowman-dahl-2021-will,
    title = "What Will it Take to Fix Benchmarking in Natural Language Understanding?",
    author = "Bowman, Samuel R.  and
      Dahl, George",
    editor = "Toutanova, Kristina  and
      Rumshisky, Anna  and
      Zettlemoyer, Luke  and
      Hakkani-Tur, Dilek  and
      Beltagy, Iz  and
      Bethard, Steven  and
      Cotterell, Ryan  and
      Chakraborty, Tanmoy  and
      Zhou, Yichao",
    booktitle = "Proceedings of the 2021 Conference of the North American Chapter of the Association for Computational Linguistics: Human Language Technologies",
    month = jun,
    year = "2021",
    address = "Online",
    publisher = "Association for Computational Linguistics",
    url = "https://aclanthology.org/2021.naacl-main.385/",
    doi = "10.18653/v1/2021.naacl-main.385",
    pages = "4843--4855"
}

@misc{nllb2022,
      title={No Language Left Behind: Scaling Human-Centered Machine Translation}, 
      author={NLLB Team and Marta R. Costa-jussà and James Cross et al.},
      year={2022},
      eprint={2207.04672},
      archivePrefix={arXiv},
      primaryClass={cs.CL},
      url={https://arxiv.org/abs/2207.04672}, 
}
\bibliographystyle{colm2026_conference}
\clearpage

\onecolumn
\appendix

\part{Appendix}
\etocsetnexttocdepth{2}
\localtableofcontents

\clearpage

\section{Additional Robustness Benchmark Details}
\label{app:robustness_benchmark_details}
\textbf{Dataset Construction.} We construct a dedicated robustness extension of our LiT benchmark consisting of 480 samples (60 source sentences × 8 language sequences), targeting five categories of classically challenging phenomena:

\textit{\textbf{(i) Conceptual and Abstract Nuance:}} Passages where physical or spatial vocabulary is used metaphorically to describe abstract concepts (e.g., "a heavy decision"), testing whether models preserve figurative meaning across languages.

\textit{\textbf{(ii) Idioms and Cultural Metaphors:}} Figurative expressions whose meaning cannot be recovered from literal word-by-word translation, testing whether models preserve communicative intent rather than surface form.

\textit{\textbf{(iii) Polysemy and Lexical Ambiguity:}} Words carrying multiple distinct meanings where only surrounding context disambiguates the intended sense (e.g., "bank" as financial institution vs. riverbank).

\textit{\textbf{(iv) Register and Tone Shifts:}} Passages that shift between formal, colloquial, and technical registers within a single example, testing sensitivity to sociolinguistic appropriateness.

\textit{\textbf{(v) Syntactic Complexity and Garden Paths:}} Sentences whose grammatical structure remains ambiguous until late in the sentence, forcing re-parsing and testing whether models maintain structural fidelity through the translation chain.

\subsection{Results}
Table~\ref{tab:lit_robustness_category_spearman} showcases the high correlation between the main LiT benchmark and the robustness benchmark. This indicates that round-trip translation with state-of-the-art frontier models is robust to classical backtranslation weaknesses.

\begin{table}[!htbp]
\centering
\small
\setlength{\tabcolsep}{5pt}
\caption{Rank correlation between LiT categories and the robustness benchmark under $MQM_{\geq 80}$. We report Spearman rank correlations between model performance on each LiT category and the robustness benchmark average, using two-sided tests over all models.}
\begin{tabular}{lccc}
\toprule
\textbf{LiT Category} & \textbf{Spearman $\rho$} & \textbf{$p$-value} & \textbf{Significance} \\
\midrule
Average & 0.946 & $3.11\times10^{-10}$ & *** \\
Humanities & 0.917 & $1.27\times10^{-8}$ & *** \\
STEM & 0.901 & $5.98\times10^{-8}$ & *** \\
Core Semantics & 0.959 & $2.46\times10^{-11}$ & *** \\
Discourse Coherence & 0.953 & $8.98\times10^{-11}$ & *** \\
Implicit Content & 0.954 & $7.78\times10^{-11}$ & *** \\
Pragmatic Inference & 0.949 & $1.88\times10^{-10}$ & *** \\
Social Interaction & 0.940 & $7.46\times10^{-10}$ & *** \\
Informal & 0.777 & $5.49\times10^{-5}$ & *** \\
\bottomrule
\end{tabular}
\label{tab:lit_robustness_category_spearman}
\vspace*{-0.2cm}
\end{table}
\clearpage

\section{Robustness of Metric and Judge Choice}
\label{app:robustness_judge}
A potential concern with any LLM-as-a-judge evaluation is sensitivity to the choice of metric formulation and judge model. If model rankings shifted substantially under alternative scoring rubrics, the benchmark's conclusions would be fragile. We address this concern by evaluating all models under two complementary scoring metrics and verifying that rankings remain stable across them.

\textbf{Metrics.} In the main paper, we report MQM$_{\ge80}$: the percentage of translations whose MQM score meets or exceeds the 80-point threshold widely considered to indicate fit-for-purpose translation quality. We additionally evaluate two alternative metrics:

\textit{\textbf{(i) Raw MQM scores}} (Table 6 and 8): Rather than binarizing at the 80-point threshold, we report the average continuous MQM score for each model–sequence combination. This preserves the full distribution of translation quality and avoids potential artifacts introduced by a fixed cutoff.

\textit{\textbf{(ii) Direct judge scores}} (Tables 7 and 9): We prompt the judge model to assign a holistic quality score on a continuous 0–100 scale, without reference to the MQM error taxonomy. This tests whether the structured MQM framework and a simple judge-based scoring rubric converge on the same model ordering.

\subsection{Results}

Comparing global averages across the three scoring paradigms reveals highly stable rankings. The top tier is unchanged across all three metrics: Gemini-3-Flash leads decisively, followed by Gemma-4-31B (Instruct) and Qwen3.5-397B (Thinking). The bottom tier is equally stable, with Nemotron-3-Nano and Qwen3-30B (Instruct) consistently occupying the lowest positions. Mid-table models exhibit only modest reshuffling (typically within 1–2 rank positions), which is expected given that these models perform similarly and minor scoring differences can reorder near-tied entries.
 
The language-sequence breakdown further confirms robustness. All three metrics agree on the central finding: performance collapses catastrophically from high-resource to low-resource language sequences. Under raw MQM (Table 8), only Gemini-3-Flash maintains a clearly usable average score (75.4) on low-resource sequences, while the next-best model drops to 43.7. Under judge scores (Table 9), the same pattern holds, with Gemini-3-Flash at 79.4 and the runner-up at 64.4. The qualitative conclusion -- that a steep accuracy cliff separates high-resource from low-resource performance -- is invariant to the metric.

\textbf{Summary.} Across three metrics (MQM$_{\ge80}$, Raw MQM, and direct judge scores) and validated against an external human-preference benchmark (LMArena), model rankings on LiT remain highly consistent. This stability indicates that our findings -- including the disconnect between reasoning benchmarks and multilingual proficiency, the low-resource performance collapse, and the underperformance of reasoning models on informal text -- are robust properties of the models themselves, not artifacts of a particular evaluation configuration.

\definecolor{HeatColor}{HTML}{2E8B57}
\begingroup
\newcommand{\avgmqm}[3]{%
  \cellcolor{HeatColor!#1}%
  \shortstack[c]{\rule{0pt}{1.05em}#2\\[0.25em]{\scriptsize $\pm$ #3}}%
}

\begin{table*}[t]
\centering
\caption{LiT benchmark by linguistic category under MQM. We report mean MQM scores (higher is better), aggregated within each category over the same eight translation sequences. Each cell shows the mean and the bootstrap-estimated standard error, computed via sentence-level resampling after averaging across sequences. Informal text is the hardest category overall (36.7 avg), while Core Semantics is the easiest (64.4).}
\resizebox{1.0\textwidth}{!}{
\renewcommand{\arraystretch}{1.55}
\setlength{\tabcolsep}{3.5pt}
\small
\begin{tabular}{>{\raggedright\arraybackslash}m{4.0cm} c cc ccccc c}
\toprule
\multirow{2}{*}{\textbf{Model}} & \multirow{2}{*}{\textbf{Average}} & \multicolumn{2}{c}{\textbf{Abstracts}} & \multicolumn{5}{c}{\textbf{Pragmatics}} & \textbf{Informal} \\
\cmidrule(lr){3-4} \cmidrule(lr){5-9}
 & & \textbf{Humanities} & \textbf{STEM} & \textbf{Core Semantics} & \textbf{Discourse} & \textbf{Implicit} & \textbf{Inference} & \textbf{Social} & \\
\midrule
\shortstack[l]{Gemini-3-Flash\\(No-Thinking)}
 & \avgmqm{43}{86.9}{0.3} & \avgmqm{44}{88.6}{0.8} & \avgmqm{43}{86.4}{0.9} & \avgmqm{44}{88.8}{0.8} & \avgmqm{44}{87.7}{0.8} & \avgmqm{43}{85.2}{1.0} & \avgmqm{44}{87.5}{0.8} & \avgmqm{44}{87.5}{0.7} & \avgmqm{42}{83.8}{0.8} \\
Qwen3.5-397B (Thinking) & \avgmqm{39}{78.1}{0.6} & \avgmqm{38}{76.9}{2.1} & \avgmqm{44}{87.5}{1.1} & \avgmqm{43}{85.2}{1.1} & \avgmqm{40}{80.8}{1.4} & \avgmqm{40}{80.0}{1.8} & \avgmqm{40}{79.3}{1.5} & \avgmqm{39}{79.0}{1.2} & \avgmqm{28}{56.5}{2.2} \\
Gemma-4-31B (Instruct) & \avgmqm{38}{75.2}{0.5} & \avgmqm{39}{77.4}{1.5} & \avgmqm{41}{82.0}{1.8} & \avgmqm{38}{75.6}{1.6} & \avgmqm{39}{77.8}{1.6} & \avgmqm{36}{72.3}{1.8} & \avgmqm{38}{76.8}{1.2} & \avgmqm{38}{75.3}{1.1} & \avgmqm{32}{64.7}{1.6} \\
Gemma-4-31B (Thinking) & \avgmqm{37}{73.7}{0.6} & \avgmqm{39}{77.2}{1.5} & \avgmqm{41}{81.6}{1.7} & \avgmqm{37}{74.9}{1.8} & \avgmqm{37}{75.0}{1.5} & \avgmqm{36}{71.3}{2.3} & \avgmqm{37}{74.0}{1.5} & \avgmqm{37}{73.6}{1.3} & \avgmqm{31}{62.4}{1.9} \\
GLM-5 (Thinking) & \avgmqm{37}{73.5}{0.7} & \avgmqm{38}{76.1}{2.1} & \avgmqm{40}{79.9}{2.1} & \avgmqm{40}{79.9}{1.5} & \avgmqm{39}{77.8}{1.3} & \avgmqm{37}{74.0}{2.3} & \avgmqm{37}{74.1}{1.7} & \avgmqm{37}{74.6}{1.8} & \avgmqm{26}{51.4}{2.3} \\
Qwen3.5-397B (Instruct) & \avgmqm{36}{72.5}{0.8} & \avgmqm{38}{76.1}{2.4} & \avgmqm{39}{78.6}{2.9} & \avgmqm{38}{76.6}{2.1} & \avgmqm{37}{73.2}{1.7} & \avgmqm{37}{73.5}{2.4} & \avgmqm{37}{74.6}{1.4} & \avgmqm{36}{72.8}{1.4} & \avgmqm{27}{54.4}{2.9} \\
GLM-4.7 (Thinking) & \avgmqm{32}{63.2}{0.8} & \avgmqm{34}{67.1}{2.6} & \avgmqm{35}{69.4}{2.4} & \avgmqm{34}{68.5}{2.3} & \avgmqm{34}{67.5}{1.8} & \avgmqm{32}{64.4}{2.3} & \avgmqm{33}{65.8}{1.6} & \avgmqm{31}{62.6}{1.8} & \avgmqm{20}{40.1}{2.1} \\
\shortstack[l]{DeepSeek-V3.2-Exp\\(Thinking)}
 & \avgmqm{31}{62.9}{0.9} & \avgmqm{32}{64.8}{3.0} & \avgmqm{26}{51.3}{4.7} & \avgmqm{36}{72.4}{1.7} & \avgmqm{35}{69.2}{1.6} & \avgmqm{34}{67.6}{2.4} & \avgmqm{33}{66.0}{2.1} & \avgmqm{34}{69.0}{1.4} & \avgmqm{22}{43.1}{2.7} \\
Kimi-K2 (Thinking) & \avgmqm{31}{61.5}{0.8} & \avgmqm{29}{58.4}{3.2} & \avgmqm{30}{59.9}{3.9} & \avgmqm{35}{70.0}{1.3} & \avgmqm{34}{67.8}{1.8} & \avgmqm{32}{63.2}{2.0} & \avgmqm{33}{66.8}{1.5} & \avgmqm{32}{63.5}{1.4} & \avgmqm{21}{42.5}{2.5} \\
DeepSeek-V3.2-Exp & \avgmqm{30}{59.2}{1.0} & \avgmqm{32}{63.1}{3.0} & \avgmqm{25}{49.7}{4.9} & \avgmqm{32}{64.2}{1.8} & \avgmqm{31}{61.7}{2.3} & \avgmqm{31}{62.9}{2.6} & \avgmqm{33}{65.1}{1.5} & \avgmqm{31}{62.1}{1.6} & \avgmqm{22}{44.6}{2.4} \\
Gemma-3-27B (Instruct) & \avgmqm{29}{58.8}{1.1} & \avgmqm{30}{59.2}{2.9} & \avgmqm{17}{33.7}{6.9} & \avgmqm{35}{69.1}{1.3} & \avgmqm{34}{67.5}{1.2} & \avgmqm{31}{61.5}{2.0} & \avgmqm{35}{69.3}{1.3} & \avgmqm{32}{63.4}{1.9} & \avgmqm{24}{47.1}{2.0} \\
Qwen3.5-35B (Thinking) & \avgmqm{29}{57.8}{1.0} & \avgmqm{27}{55.0}{3.3} & \avgmqm{34}{67.0}{3.0} & \avgmqm{33}{66.9}{2.0} & \avgmqm{32}{64.5}{1.7} & \avgmqm{30}{60.2}{3.9} & \avgmqm{33}{65.2}{1.2} & \avgmqm{30}{59.7}{1.9} & \avgmqm{12}{24.3}{3.5} \\
GLM-5 (Instruct) & \avgmqm{29}{57.6}{1.0} & \avgmqm{28}{56.9}{3.7} & \avgmqm{28}{56.3}{4.9} & \avgmqm{33}{65.9}{1.8} & \avgmqm{30}{60.1}{2.3} & \avgmqm{30}{60.7}{2.4} & \avgmqm{30}{60.6}{1.9} & \avgmqm{29}{58.1}{1.8} & \avgmqm{21}{41.9}{2.2} \\
Qwen3-235B (Thinking) & \avgmqm{27}{54.9}{0.8} & \avgmqm{27}{53.2}{2.4} & \avgmqm{28}{56.4}{3.2} & \avgmqm{31}{62.8}{1.4} & \avgmqm{30}{60.8}{1.8} & \avgmqm{28}{56.8}{2.8} & \avgmqm{31}{62.0}{1.0} & \avgmqm{28}{55.8}{1.5} & \avgmqm{16}{31.5}{2.5} \\
Kimi-K2 & \avgmqm{26}{52.1}{1.1} & \avgmqm{24}{47.6}{4.0} & \avgmqm{19}{38.5}{5.8} & \avgmqm{30}{61.0}{1.6} & \avgmqm{28}{56.8}{2.2} & \avgmqm{28}{55.5}{2.5} & \avgmqm{30}{59.3}{1.6} & \avgmqm{28}{56.4}{1.8} & \avgmqm{21}{41.9}{2.8} \\
GPT-OSS-120B (High) & \avgmqm{26}{52.1}{0.9} & \avgmqm{24}{48.7}{3.1} & \avgmqm{31}{61.2}{2.5} & \avgmqm{31}{62.4}{1.5} & \avgmqm{30}{60.0}{2.6} & \avgmqm{27}{53.7}{4.1} & \avgmqm{29}{58.7}{1.4} & \avgmqm{26}{52.8}{2.0} & \avgmqm{10}{19.3}{3.3} \\
Qwen3-235B (Instruct) & \avgmqm{25}{49.4}{0.9} & \avgmqm{23}{45.5}{3.0} & \avgmqm{21}{42.4}{4.4} & \avgmqm{29}{58.9}{1.5} & \avgmqm{26}{52.9}{1.9} & \avgmqm{27}{53.6}{1.7} & \avgmqm{27}{53.1}{1.7} & \avgmqm{27}{54.4}{1.8} & \avgmqm{17}{34.3}{2.5} \\
Qwen3.5-35B (Instruct) & \avgmqm{25}{49.4}{1.0} & \avgmqm{23}{45.1}{3.2} & \avgmqm{26}{51.9}{4.3} & \avgmqm{28}{56.3}{1.7} & \avgmqm{28}{55.6}{1.9} & \avgmqm{26}{51.1}{3.4} & \avgmqm{30}{60.5}{1.5} & \avgmqm{26}{52.9}{1.8} & \avgmqm{11}{21.4}{3.6} \\
MiniMax-M2.5 & \avgmqm{24}{47.2}{1.0} & \avgmqm{22}{43.4}{3.6} & \avgmqm{25}{50.0}{3.6} & \avgmqm{31}{61.3}{1.9} & \avgmqm{29}{58.7}{1.9} & \avgmqm{26}{52.1}{3.8} & \avgmqm{27}{54.2}{1.9} & \avgmqm{26}{51.7}{1.9} & \avgmqm{3}{6.5}{3.6} \\
MiMo-V2-Flash & \avgmqm{19}{38.0}{1.0} & \avgmqm{17}{33.8}{2.3} & \avgmqm{6}{12.2}{5.4} & \avgmqm{26}{51.7}{1.9} & \avgmqm{26}{52.1}{2.7} & \avgmqm{22}{44.8}{2.5} & \avgmqm{24}{48.9}{1.8} & \avgmqm{22}{43.8}{1.3} & \avgmqm{8}{16.6}{3.0} \\
Qwen3-30B (Instruct) & \avgmqm{16}{31.1}{1.1} & \avgmqm{12}{24.8}{3.7} & \avgmqm{10}{19.3}{5.4} & \avgmqm{21}{42.3}{2.2} & \avgmqm{20}{40.2}{3.0} & \avgmqm{20}{39.4}{2.2} & \avgmqm{20}{39.7}{1.5} & \avgmqm{19}{37.8}{2.1} & \avgmqm{3}{5.1}{2.5} \\
Nemotron-3-Nano & \avgmqm{0}{-6.3}{1.2} & \avgmqm{0}{-11.9}{3.2} & \avgmqm{0}{-7.2}{5.0} & \avgmqm{1}{1.2}{3.0} & \avgmqm{0}{0.2}{2.8} & \avgmqm{0}{-5.0}{3.9} & \avgmqm{0}{0.4}{2.9} & \avgmqm{0}{-1.4}{2.5} & \avgmqm{0}{-26.6}{3.4} \\
\midrule
\textbf{Average} & \cellcolor{HeatColor!28}56.8 & \cellcolor{HeatColor!28}55.8 & \cellcolor{HeatColor!27}54.9 & \cellcolor{HeatColor!32}64.4 & \cellcolor{HeatColor!31}62.2 & \cellcolor{HeatColor!30}59.0 & \cellcolor{HeatColor!31}61.9 & \cellcolor{HeatColor!30}59.3 & \cellcolor{HeatColor!18}36.7 \\
\bottomrule
\end{tabular}}
\vspace*{-0.4cm}
\label{tab:mqm_scores_raw_hierarchical_sorted}
\end{table*}
\endgroup
\definecolor{HeatColor}{HTML}{2E8B57}
\begingroup
\newcommand{\heatbg}[1]{%
  \cellcolor{HeatColor!\number\numexpr(#1*4+2)/5\relax}%
}
\newcommand{\avgmqm}[3]{%
  \heatbg{#1}%
  \shortstack[c]{\rule{0pt}{1.05em}#2\\[0.25em]{\scriptsize $\pm$ #3}}%
}

\begin{table*}[t]
\centering
\caption{LiT benchmark by linguistic category under raw Judge Score. We report mean judge scores (higher is better), aggregated within each category over the same eight translation sequences. Each cell shows the mean and the bootstrap-estimated standard error, computed via sentence-level resampling after averaging across sequences. Informal text is the hardest category overall (63.3 avg), while STEM is the easiest (71.3).}
\resizebox{1.0\textwidth}{!}{
\renewcommand{\arraystretch}{1.55}
\setlength{\tabcolsep}{3.5pt}
\small
\begin{tabular}{>{\raggedright\arraybackslash}m{4.0cm} c cc ccccc c}
\toprule
\multirow{2}{*}{\textbf{Model}} & \multirow{2}{*}{\textbf{Average}} & \multicolumn{2}{c}{\textbf{Abstracts}} & \multicolumn{5}{c}{\textbf{Pragmatics}} & \textbf{Informal} \\
\cmidrule(lr){3-4} \cmidrule(lr){5-9}
 & & \textbf{Humanities} & \textbf{STEM} & \textbf{Core Semantics} & \textbf{Discourse} & \textbf{Implicit} & \textbf{Inference} & \textbf{Social} & \\
\midrule
\shortstack[l]{Gemini-3-Flash\\(No-Thinking)}
 & \avgmqm{45}{89.1}{0.2} & \avgmqm{46}{91.2}{0.6} & \avgmqm{45}{90.5}{0.7} & \avgmqm{45}{89.6}{0.8} & \avgmqm{44}{88.9}{0.9} & \avgmqm{43}{86.2}{0.9} & \avgmqm{44}{88.5}{0.6} & \avgmqm{44}{88.2}{0.5} & \avgmqm{45}{89.5}{0.5} \\
Qwen3.5-397B (Thinking) & \avgmqm{42}{83.9}{0.4} & \avgmqm{43}{85.2}{1.2} & \avgmqm{46}{91.1}{0.7} & \avgmqm{43}{86.8}{0.8} & \avgmqm{42}{84.9}{0.9} & \avgmqm{41}{81.9}{1.8} & \avgmqm{41}{82.5}{0.9} & \avgmqm{42}{83.1}{0.7} & \avgmqm{38}{75.7}{1.5} \\
Gemma-4-31B (Instruct) & \avgmqm{40}{80.9}{0.3} & \avgmqm{42}{83.1}{1.0} & \avgmqm{44}{87.6}{1.0} & \avgmqm{40}{79.9}{0.9} & \avgmqm{40}{80.7}{1.1} & \avgmqm{38}{76.3}{1.2} & \avgmqm{40}{80.1}{0.7} & \avgmqm{40}{79.4}{0.6} & \avgmqm{40}{80.1}{0.6} \\
GLM-5 (Thinking) & \avgmqm{40}{80.7}{0.4} & \avgmqm{41}{82.7}{1.0} & \avgmqm{43}{86.4}{1.0} & \avgmqm{41}{82.9}{1.1} & \avgmqm{41}{82.6}{0.8} & \avgmqm{39}{77.8}{1.7} & \avgmqm{40}{80.0}{0.8} & \avgmqm{40}{79.5}{0.9} & \avgmqm{37}{73.4}{1.5} \\
Qwen3.5-397B (Instruct) & \avgmqm{40}{80.1}{0.4} & \avgmqm{42}{83.6}{1.6} & \avgmqm{42}{85.0}{1.7} & \avgmqm{40}{80.4}{1.5} & \avgmqm{39}{78.9}{1.1} & \avgmqm{39}{77.2}{1.6} & \avgmqm{40}{79.7}{0.9} & \avgmqm{39}{78.6}{0.7} & \avgmqm{39}{77.4}{0.8} \\
Gemma-4-31B (Thinking) & \avgmqm{40}{79.7}{0.4} & \avgmqm{41}{82.2}{0.8} & \avgmqm{42}{84.0}{1.4} & \avgmqm{40}{79.6}{1.2} & \avgmqm{40}{79.3}{1.2} & \avgmqm{38}{76.0}{1.4} & \avgmqm{40}{79.3}{0.9} & \avgmqm{39}{78.6}{0.8} & \avgmqm{39}{79.0}{0.7} \\
GLM-4.7 (Thinking) & \avgmqm{38}{75.2}{0.4} & \avgmqm{39}{77.7}{1.1} & \avgmqm{41}{81.3}{1.2} & \avgmqm{39}{77.7}{1.1} & \avgmqm{39}{77.7}{0.7} & \avgmqm{36}{72.5}{1.7} & \avgmqm{37}{74.4}{0.7} & \avgmqm{37}{73.5}{0.9} & \avgmqm{33}{67.0}{1.3} \\
\shortstack[l]{DeepSeek-V3.2-Exp\\(Thinking)}
 & \avgmqm{37}{73.7}{0.5} & \avgmqm{39}{77.2}{1.7} & \avgmqm{36}{72.1}{2.2} & \avgmqm{38}{76.3}{1.1} & \avgmqm{38}{75.2}{1.2} & \avgmqm{36}{71.7}{1.5} & \avgmqm{37}{73.6}{0.9} & \avgmqm{37}{73.4}{0.9} & \avgmqm{35}{70.0}{1.6} \\
Kimi-K2 (Thinking) & \avgmqm{36}{72.3}{0.5} & \avgmqm{36}{72.7}{1.4} & \avgmqm{38}{75.3}{2.0} & \avgmqm{37}{74.1}{0.8} & \avgmqm{37}{74.1}{0.9} & \avgmqm{35}{69.8}{1.9} & \avgmqm{36}{72.7}{0.9} & \avgmqm{36}{72.1}{1.0} & \avgmqm{34}{67.9}{1.6} \\
GLM-5 (Instruct) & \avgmqm{36}{71.9}{0.5} & \avgmqm{36}{72.8}{1.7} & \avgmqm{37}{74.4}{2.2} & \avgmqm{37}{73.6}{0.9} & \avgmqm{36}{72.5}{1.1} & \avgmqm{35}{70.1}{1.5} & \avgmqm{36}{71.4}{0.8} & \avgmqm{35}{70.7}{0.9} & \avgmqm{35}{69.6}{0.8} \\
DeepSeek-V3.2-Exp & \avgmqm{36}{71.5}{0.5} & \avgmqm{38}{75.2}{1.6} & \avgmqm{34}{68.7}{2.4} & \avgmqm{36}{72.1}{1.3} & \avgmqm{36}{72.1}{1.2} & \avgmqm{35}{70.2}{1.3} & \avgmqm{36}{72.3}{0.9} & \avgmqm{35}{70.0}{0.9} & \avgmqm{36}{71.3}{0.7} \\
Qwen3.5-35B (Thinking) & \avgmqm{35}{69.8}{0.5} & \avgmqm{35}{69.4}{1.4} & \avgmqm{39}{78.7}{1.4} & \avgmqm{37}{74.1}{1.3} & \avgmqm{36}{71.9}{1.4} & \avgmqm{34}{67.8}{2.1} & \avgmqm{36}{71.3}{0.8} & \avgmqm{34}{67.4}{1.0} & \avgmqm{29}{57.9}{1.2} \\
Qwen3-235B (Thinking) & \avgmqm{34}{67.3}{0.4} & \avgmqm{34}{68.6}{1.1} & \avgmqm{36}{71.9}{1.4} & \avgmqm{35}{69.8}{0.9} & \avgmqm{35}{70.2}{0.9} & \avgmqm{33}{65.2}{2.0} & \avgmqm{34}{67.6}{0.6} & \avgmqm{33}{65.6}{0.9} & \avgmqm{30}{59.6}{1.1} \\
Gemma-3-27B (Instruct) & \avgmqm{34}{67.1}{0.5} & \avgmqm{35}{69.5}{1.2} & \avgmqm{30}{60.2}{2.5} & \avgmqm{35}{70.0}{1.0} & \avgmqm{35}{70.1}{1.0} & \avgmqm{32}{64.5}{1.6} & \avgmqm{35}{69.6}{0.8} & \avgmqm{33}{66.1}{0.9} & \avgmqm{34}{67.2}{0.8} \\
GPT-OSS-120B (High) & \avgmqm{33}{65.4}{0.5} & \avgmqm{34}{68.0}{1.4} & \avgmqm{36}{71.9}{1.7} & \avgmqm{35}{69.7}{0.8} & \avgmqm{35}{70.0}{1.4} & \avgmqm{32}{64.4}{2.0} & \avgmqm{34}{67.5}{0.9} & \avgmqm{32}{64.2}{1.1} & \avgmqm{24}{47.7}{1.8} \\
MiniMax-M2.5 & \avgmqm{32}{63.6}{0.5} & \avgmqm{31}{62.1}{1.6} & \avgmqm{35}{70.4}{1.6} & \avgmqm{34}{68.9}{1.1} & \avgmqm{34}{68.8}{1.2} & \avgmqm{32}{63.2}{2.1} & \avgmqm{32}{64.3}{0.9} & \avgmqm{31}{61.2}{1.1} & \avgmqm{25}{50.1}{1.2} \\
Qwen3.5-35B (Instruct) & \avgmqm{32}{63.0}{0.5} & \avgmqm{31}{62.6}{1.4} & \avgmqm{35}{69.8}{1.9} & \avgmqm{33}{65.1}{1.0} & \avgmqm{32}{63.2}{1.2} & \avgmqm{30}{60.7}{1.9} & \avgmqm{33}{65.1}{0.9} & \avgmqm{30}{60.4}{1.1} & \avgmqm{29}{57.4}{1.1} \\
Kimi-K2 & \avgmqm{31}{62.5}{0.6} & \avgmqm{32}{64.3}{1.6} & \avgmqm{30}{60.5}{3.0} & \avgmqm{33}{65.3}{1.2} & \avgmqm{32}{63.4}{1.3} & \avgmqm{30}{59.8}{1.9} & \avgmqm{32}{63.1}{0.8} & \avgmqm{31}{62.2}{1.0} & \avgmqm{31}{61.6}{1.0} \\
Qwen3-235B (Instruct) & \avgmqm{28}{56.6}{0.6} & \avgmqm{27}{54.6}{2.4} & \avgmqm{29}{57.1}{2.4} & \avgmqm{31}{61.0}{1.4} & \avgmqm{29}{57.1}{1.8} & \avgmqm{29}{57.3}{1.8} & \avgmqm{28}{56.7}{1.4} & \avgmqm{28}{56.0}{1.6} & \avgmqm{27}{53.3}{1.6} \\
MiMo-V2-Flash & \avgmqm{28}{56.2}{0.5} & \avgmqm{28}{55.5}{1.4} & \avgmqm{25}{49.1}{2.6} & \avgmqm{30}{60.6}{0.8} & \avgmqm{30}{60.3}{1.2} & \avgmqm{28}{56.2}{1.7} & \avgmqm{29}{58.2}{1.1} & \avgmqm{28}{56.1}{1.1} & \avgmqm{27}{53.8}{0.9} \\
Qwen3-30B (Instruct) & \avgmqm{24}{47.9}{0.5} & \avgmqm{23}{47.0}{1.2} & \avgmqm{24}{47.8}{2.1} & \avgmqm{25}{50.4}{1.2} & \avgmqm{26}{51.4}{1.2} & \avgmqm{24}{47.1}{1.5} & \avgmqm{25}{49.2}{0.6} & \avgmqm{24}{48.6}{1.1} & \avgmqm{21}{41.3}{0.9} \\
Nemotron-3-Nano & \avgmqm{15}{29.2}{0.5} & \avgmqm{15}{30.1}{1.4} & \avgmqm{18}{35.6}{2.0} & \avgmqm{16}{31.4}{1.2} & \avgmqm{16}{31.0}{1.0} & \avgmqm{13}{27.0}{1.4} & \avgmqm{14}{28.9}{1.3} & \avgmqm{14}{28.9}{1.3} & \avgmqm{10}{21.0}{0.9} \\
\midrule
\textbf{Average} & \heatbg{34}68.5 & \heatbg{35}69.8 & \heatbg{36}71.3 & \heatbg{35}70.9 & \heatbg{35}70.2 & \heatbg{33}66.5 & \heatbg{34}68.9 & \heatbg{34}67.4 & \heatbg{32}63.3 \\
\bottomrule
\end{tabular}}
\vspace*{-0.4cm}
\label{tab:judge_scores_raw_hierarchical_sorted}
\end{table*}
\endgroup

\begin{table}[t!]
\centering
\renewcommand{\arraystretch}{1.2}
\label{tab:lang_breakdown}
\setlength{\tabcolsep}{3pt}
\small
\caption{
\textbf{Raw MQM scores similarly collapse on low-resource languages.} We report the raw MQM scores across eight language sequences grouped by resource availability. (a) High-resource sequences (East Asian, Central European, Near Eastern): top frontier models remain strong, with Gemini-3-Flash leading at 91.0 average and GLM-5 (Thinking) taking the East Asian column at 90.4. (b) Medium-resource sequences (Southeast Asian, South Asian, North European): performance remains stable for the strongest models, with Gemini-3-Flash averaging 89.9. (c) Low-resource sequences (African, South American): catastrophic collapse remains, with only Gemini-3-Flash (75.4) maintaining clearly usable performance; the next-best model (Qwen3.5-397B (Thinking)) drops to 43.7. (d) Global averages across these eight sequences show Gemini-3-Flash holding roughly a 10.5-point lead over the next competitor.
}

\begin{minipage}[t]{0.555\textwidth}
    \centering
    \textbf{(a) High \& Med-High Resource Sequences} \\
    \vspace{2pt}
    \resizebox{\linewidth}{!}{
    \begin{tabular}{l|ccc|c}
        \toprule
        \textbf{Model} & \textbf{E. Asia} & \textbf{C. Europe} & \textbf{N. East} & \textbf{Avg} \\
        \midrule
        Gemini-3-Flash (No-Think) & 90.0 & \textbf{91.6} & \textbf{91.6} & \cellcolor{gray!15}\textbf{91.0} \\
        Qwen3.5-397B (Thinking) & 89.3 & 90.3 & 87.1 & 88.9 \\
        GLM-5 (Thinking) & \textbf{90.4} & 89.7 & 85.4 & 88.5 \\
        Gemma-4-31B (Thinking) & 89.6 & 88.0 & 86.8 & 88.1 \\
        Gemma-4-31B (Instruct) & 88.6 & 88.2 & 86.0 & 87.6 \\
        Qwen3.5-397B (Instruct) & 88.1 & 89.3 & 85.2 & 87.5 \\
        GLM-4.7 (Thinking) & 89.3 & 89.0 & 82.9 & 87.1 \\
        Kimi-K2 (Thinking) & 86.1 & 87.2 & 82.8 & 85.4 \\
        DeepSeek-V3.2-Exp & 87.6 & 86.2 & 79.9 & 84.6 \\
        Qwen3-235B (Thinking) & 85.0 & 87.3 & 80.0 & 84.1 \\
        GLM-5 (Instruct) & 86.3 & 85.9 & 77.8 & 83.3 \\
        DeepSeek-V3.2-Exp (Think) & 85.8 & 82.9 & 79.7 & 82.8 \\
        Qwen3.5-35B (Thinking) & 81.2 & 85.3 & 75.7 & 80.8 \\
        Gemma-3-27B (Instruct) & 78.6 & 83.1 & 77.8 & 79.8 \\
        Kimi-K2 & 79.4 & 82.8 & 70.5 & 77.6 \\
        Qwen3.5-35B (Instruct) & 79.3 & 80.7 & 64.1 & 74.7 \\
        Qwen3-235B (Instruct) & 83.7 & 68.3 & 72.0 & 74.7 \\
        MiniMax-M2.5 & 71.6 & 80.2 & 68.7 & 73.5 \\
        GPT-OSS-120B (High) & 78.3 & 76.8 & 63.9 & 73.0 \\
        MiMo-V2-Flash & 70.2 & 74.5 & 54.3 & 66.4 \\
        Qwen3-30B (Instruct) & 71.3 & 67.5 & 44.4 & 61.0 \\
        Nemotron-3-Nano & 50.9 & 11.5 & -17.1 & 15.1 \\
        \bottomrule
    \end{tabular}}

    \vspace{12pt}

    \centering
    \textbf{(b) Medium Resource Sequences} \\
    \vspace{2pt}
    \resizebox{\linewidth}{!}{
    \begin{tabular}{l|ccc|c}
        \toprule
        \textbf{Model} & \textbf{SE. Asia} & \textbf{S. Asia} & \textbf{N. Europe} & \textbf{Avg} \\
        \midrule
        Gemini-3-Flash (No-Think) & \textbf{89.3} & \textbf{90.3} & \textbf{90.2} & \cellcolor{gray!15}\textbf{89.9} \\
        Qwen3.5-397B (Thinking) & 86.0 & 83.1 & 86.3 & 85.1 \\
        GLM-5 (Thinking) & 86.3 & 83.3 & 83.7 & 84.5 \\
        Gemma-4-31B (Instruct) & 88.6 & 88.2 & 76.0 & 84.2 \\
        Gemma-4-31B (Thinking) & 86.5 & 89.0 & 74.7 & 83.4 \\
        Qwen3.5-397B (Instruct) & 85.8 & 79.7 & 81.9 & 82.5 \\
        GLM-4.7 (Thinking) & 80.9 & 78.1 & 74.6 & 77.9 \\
        Kimi-K2 (Thinking) & 79.6 & 73.2 & 77.6 & 76.8 \\
        GLM-5 (Instruct) & 75.3 & 75.5 & 68.0 & 72.9 \\
        DeepSeek-V3.2-Exp & 81.4 & 66.4 & 70.8 & 72.9 \\
        Qwen3-235B (Thinking) & 81.1 & 71.5 & 63.5 & 72.0 \\
        DeepSeek-V3.2-Exp (Think) & 79.9 & 65.2 & 68.8 & 71.3 \\
        Gemma-3-27B (Instruct) & 76.5 & 66.9 & 66.8 & 70.1 \\
        Qwen3.5-35B (Thinking) & 74.4 & 62.7 & 66.6 & 67.9 \\
        Kimi-K2 & 72.7 & 58.3 & 67.0 & 66.0 \\
        Qwen3-235B (Instruct) & 77.2 & 65.6 & 43.7 & 62.2 \\
        GPT-OSS-120B (High) & 67.7 & 58.0 & 59.4 & 61.7 \\
        Qwen3.5-35B (Instruct) & 68.0 & 46.6 & 53.6 & 56.1 \\
        MiniMax-M2.5 & 61.6 & 34.8 & 57.9 & 51.4 \\
        MiMo-V2-Flash & 62.3 & 38.7 & 40.1 & 47.0 \\
        Qwen3-30B (Instruct) & 53.3 & 18.1 & 1.7 & 24.4 \\
        Nemotron-3-Nano & -8.1 & -41.8 & -46.0 & -31.9 \\
        \bottomrule
    \end{tabular}}
\end{minipage}
\hfill
\begin{minipage}[t]{0.435\textwidth}
    \centering
    \textbf{(c) Low Resource / Imbalanced Sequences} \\
    \vspace{2pt}
    \resizebox{\linewidth}{!}{
    \begin{tabular}{l|cc|c}
        \toprule
        \textbf{Model} & \textbf{Africa} & \textbf{S. America} & \textbf{Avg} \\
        \midrule
        Gemini-3-Flash (No-Think) & \textbf{84.4} & \textbf{66.4} & \cellcolor{gray!15}\textbf{75.4} \\
        Qwen3.5-397B (Thinking) & 58.5 & 28.9 & 43.7 \\
        Gemma-4-31B (Instruct) & 74.8 & 3.5 & 39.2 \\
        Gemma-4-31B (Thinking) & 72.6 & -6.2 & 33.2 \\
        Qwen3.5-397B (Instruct) & 45.9 & 10.7 & 28.3 \\
        GLM-5 (Thinking) & 49.1 & 3.9 & 26.5 \\
        DeepSeek-V3.2-Exp (Think) & 29.6 & 0.4 & 15.0 \\
        Gemma-3-27B (Instruct) & 16.8 & 1.0 & 8.9 \\
        Kimi-K2 (Thinking) & 11.7 & -17.5 & -2.9 \\
        Qwen3.5-35B (Thinking) & 13.8 & -19.7 & -3.0 \\
        GLM-4.7 (Thinking) & 18.4 & -24.7 & -3.2 \\
        DeepSeek-V3.2-Exp & 13.6 & -21.0 & -3.7 \\
        GPT-OSS-120B (High) & 3.4 & -13.2 & -4.9 \\
        Nemotron-3-Nano & -12.8 & 1.5 & -5.6 \\
        Qwen3.5-35B (Instruct) & -11.4 & -3.2 & -7.3 \\
        Kimi-K2 & -16.2 & -1.0 & -8.6 \\
        GLM-5 (Instruct) & 9.2 & -27.6 & -9.2 \\
        Qwen3-30B (Instruct) & -12.7 & -8.9 & -10.8 \\
        Qwen3-235B (Instruct) & -17.3 & -5.9 & -11.6 \\
        MiniMax-M2.5 & -15.1 & -8.4 & -11.8 \\
        Qwen3-235B (Thinking) & -35.8 & -8.4 & -22.1 \\
        MiMo-V2-Flash & -34.6 & -11.1 & -22.9 \\
        \bottomrule
    \end{tabular}}

    \vspace{12pt}

    \centering
    \textbf{(d) Overall Performance} \\
    \vspace{2pt}
    \resizebox{\linewidth}{!}{
    \begin{tabular}{l|c}
        \toprule
        \textbf{Model} & \textbf{Global Average} \\
        \midrule
        Gemini-3-Flash (No-Think) & \cellcolor{gray!15}\textbf{86.7} \\
        Qwen3.5-397B (Thinking) & 76.2 \\
        Gemma-4-31B (Instruct) & 74.2 \\
        Gemma-4-31B (Thinking) & 72.6 \\
        GLM-5 (Thinking) & 71.5 \\
        Qwen3.5-397B (Instruct) & 70.8 \\
        DeepSeek-V3.2-Exp (Think) & 61.5 \\
        GLM-4.7 (Thinking) & 61.1 \\
        Kimi-K2 (Thinking) & 60.1 \\
        Gemma-3-27B (Instruct) & 58.4 \\
        DeepSeek-V3.2-Exp & 58.1 \\
        GLM-5 (Instruct) & 56.3 \\
        Qwen3.5-35B (Thinking) & 55.0 \\
        Qwen3-235B (Thinking) & 53.0 \\
        Kimi-K2 & 51.7 \\
        GPT-OSS-120B (High) & 49.3 \\
        Qwen3-235B (Instruct) & 48.4 \\
        Qwen3.5-35B (Instruct) & 47.2 \\
        MiniMax-M2.5 & 43.9 \\
        MiMo-V2-Flash & 36.8 \\
        Qwen3-30B (Instruct) & 29.3 \\
        Nemotron-3-Nano & -7.7 \\
        \bottomrule
    \end{tabular}}
\end{minipage}
\label{tab:mqm_quadrant_breakdown}
\vspace*{-0.4cm}
\end{table}

\begin{table}[t!]
\centering
\renewcommand{\arraystretch}{1.2}
\setlength{\tabcolsep}{3pt}
\small
\caption{
\textbf{Judge scores also drop on low-resource languages.} We report judge scores across the same eight language sequences on the 200-example subset, grouped by resource availability. (a) High-resource sequences remain strong for frontier models, with Gemini-3-Flash averaging 93.0 and taking both Central Europe (94.2) and Near East (93.5), while GLM-5 (Thinking) narrowly leads East Asia (91.5). (b) Medium-resource sequences remain comparatively stable for the strongest models, with Gemini-3-Flash averaging 91.6. (c) Low-resource sequences show a substantial drop, with Gemini-3-Flash still clearly first at 79.4 average, while the next-best model, Qwen3.5-397B (Thinking), falls to 64.4. (d) Global averages across these eight sequences preserve the same ranking pattern, with Gemini-3-Flash leading at 89.1 and holding a 6.0-point lead over the next competitor.
}

\begin{minipage}[t]{0.555\textwidth}
    \centering
    \textbf{(a) High \& Med-High Resource Sequences} \\
    \vspace{2pt}
    \resizebox{\linewidth}{!}{
    \begin{tabular}{l|ccc|c}
        \toprule
        \textbf{Model} & \textbf{E. Asia} & \textbf{C. Europe} & \textbf{N. East} & \textbf{Avg} \\
        \midrule
        Gemini-3-Flash (No-Think) & 91.2 & \textbf{94.2} & \textbf{93.5} & \cellcolor{gray!15}\textbf{93.0} \\
        Gemma-4-31B (Thinking) & 91.1 & 91.3 & 90.3 & 90.9 \\
        Qwen3.5-397B (Thinking) & 91.0 & 91.5 & 90.0 & 90.8 \\
        Qwen3.5-397B (Instruct) & 89.7 & 91.9 & 90.1 & 90.6 \\
        GLM-5 (Thinking) & \textbf{91.5} & 91.3 & 88.7 & 90.5 \\
        Gemma-4-31B (Instruct) & 90.5 & 90.9 & 89.9 & 90.4 \\
        GLM-4.7 (Thinking) & 90.2 & 90.3 & 87.3 & 89.3 \\
        Kimi-K2 (Thinking) & 88.0 & 88.6 & 87.8 & 88.1 \\
        DeepSeek-V3.2-Exp & 88.6 & 89.0 & 85.3 & 87.6 \\
        GLM-5 (Instruct) & 88.6 & 89.6 & 83.9 & 87.3 \\
        Qwen3-235B (Thinking) & 87.4 & 89.3 & 85.3 & 87.3 \\
        Qwen3.5-35B (Thinking) & 83.8 & 88.3 & 82.8 & 85.0 \\
        DeepSeek-V3.2-Exp (Think) & 87.5 & 83.0 & 84.1 & 84.8 \\
        Gemma-3-27B (Instruct) & 82.7 & 87.2 & 84.4 & 84.8 \\
        Kimi-K2 & 82.4 & 86.9 & 78.9 & 82.8 \\
        Qwen3.5-35B (Instruct) & 83.6 & 85.6 & 77.0 & 82.1 \\
        MiniMax-M2.5 & 77.4 & 84.5 & 79.0 & 80.3 \\
        GPT-OSS-120B (High) & 81.3 & 79.5 & 76.1 & 79.0 \\
        Qwen3-235B (Instruct) & 86.3 & 70.2 & 78.6 & 78.4 \\
        MiMo-V2-Flash & 77.6 & 81.7 & 70.1 & 76.5 \\
        Qwen3-30B (Instruct) & 79.1 & 76.4 & 65.2 & 73.6 \\
        Nemotron-3-Nano & 66.4 & 48.5 & 29.9 & 48.3 \\
        \bottomrule
    \end{tabular}}

    \vspace{12pt}

    \centering
    \textbf{(b) Medium Resource Sequences} \\
    \vspace{2pt}
    \resizebox{\linewidth}{!}{
    \begin{tabular}{l|ccc|c}
        \toprule
        \textbf{Model} & \textbf{SE. Asia} & \textbf{S. Asia} & \textbf{N. Europe} & \textbf{Avg} \\
        \midrule
        Gemini-3-Flash (No-Think) & \textbf{90.3} & \textbf{91.7} & \textbf{92.8} & \cellcolor{gray!15}\textbf{91.6} \\
        Qwen3.5-397B (Thinking) & 86.9 & 86.7 & 89.8 & 87.8 \\
        Gemma-4-31B (Instruct) & 89.9 & 89.4 & 82.7 & 87.3 \\
        GLM-5 (Thinking) & 86.7 & 84.7 & 88.2 & 86.5 \\
        Qwen3.5-397B (Instruct) & 88.1 & 84.2 & 86.9 & 86.4 \\
        Gemma-4-31B (Thinking) & 86.0 & 90.2 & 81.7 & 86.0 \\
        GLM-4.7 (Thinking) & 84.4 & 83.0 & 83.7 & 83.7 \\
        Kimi-K2 (Thinking) & 82.7 & 78.6 & 84.3 & 81.9 \\
        Qwen3-235B (Thinking) & 83.3 & 80.3 & 78.5 & 80.7 \\
        DeepSeek-V3.2-Exp & 84.3 & 77.3 & 80.1 & 80.6 \\
        GLM-5 (Instruct) & 80.4 & 80.3 & 78.4 & 79.7 \\
        DeepSeek-V3.2-Exp (Think) & 81.4 & 74.1 & 79.4 & 78.3 \\
        Gemma-3-27B (Instruct) & 81.1 & 76.6 & 77.2 & 78.3 \\
        Qwen3.5-35B (Thinking) & 78.5 & 74.3 & 79.6 & 77.5 \\
        Kimi-K2 & 77.3 & 71.6 & 78.0 & 75.6 \\
        GPT-OSS-120B (High) & 71.8 & 72.8 & 72.8 & 72.5 \\
        Qwen3.5-35B (Instruct) & 76.2 & 65.8 & 71.6 & 71.2 \\
        Qwen3-235B (Instruct) & 79.1 & 73.6 & 53.8 & 68.8 \\
        MiniMax-M2.5 & 72.9 & 58.0 & 74.2 & 68.4 \\
        MiMo-V2-Flash & 72.6 & 63.7 & 65.8 & 67.4 \\
        Qwen3-30B (Instruct) & 66.5 & 50.0 & 41.3 & 52.6 \\
        Nemotron-3-Nano & 33.2 & 15.9 & 17.0 & 22.0 \\
        \bottomrule
    \end{tabular}}
\end{minipage}
\hfill
\begin{minipage}[t]{0.435\textwidth}
    \centering
    \textbf{(c) Low Resource / Imbalanced Sequences} \\
    \vspace{2pt}
    \resizebox{\linewidth}{!}{
    \begin{tabular}{l|cc|c}
        \toprule
        \textbf{Model} & \textbf{Africa} & \textbf{S. America} & \textbf{Avg} \\
        \midrule
        Gemini-3-Flash (No-Think) & \textbf{86.2} & \textbf{72.6} & \cellcolor{gray!15}\textbf{79.4} \\
        Qwen3.5-397B (Thinking) & 72.4 & 56.3 & 64.4 \\
        Gemma-4-31B (Instruct) & 79.4 & 32.7 & 56.1 \\
        GLM-5 (Thinking) & 66.4 & 41.9 & 54.1 \\
        Qwen3.5-397B (Instruct) & 66.2 & 41.0 & 53.6 \\
        Gemma-4-31B (Thinking) & 74.2 & 31.9 & 53.1 \\
        DeepSeek-V3.2-Exp (Think) & 57.7 & 39.7 & 48.7 \\
        GLM-4.7 (Thinking) & 52.1 & 23.9 & 38.0 \\
        GLM-5 (Instruct) & 48.4 & 23.6 & 36.0 \\
        DeepSeek-V3.2-Exp & 46.6 & 20.4 & 33.5 \\
        Kimi-K2 (Thinking) & 46.8 & 18.5 & 32.6 \\
        Qwen3.5-35B (Thinking) & 47.1 & 14.8 & 31.0 \\
        GPT-OSS-120B (High) & 36.7 & 18.8 & 27.8 \\
        MiniMax-M2.5 & 33.4 & 19.4 & 26.4 \\
        Gemma-3-27B (Instruct) & 46.3 & 2.6 & 24.5 \\
        Qwen3.5-35B (Instruct) & 33.2 & 6.8 & 20.0 \\
        Qwen3-235B (Thinking) & 16.3 & 12.0 & 14.2 \\
        Kimi-K2 & 19.8 & 5.0 & 12.4 \\
        MiMo-V2-Flash & 7.9 & 9.4 & 8.7 \\
        Nemotron-3-Nano & 5.7 & 10.8 & 8.3 \\
        Qwen3-235B (Instruct) & 7.2 & 2.2 & 4.7 \\
        Qwen3-30B (Instruct) & 0.0 & 0.1 & 0.1 \\
        \bottomrule
    \end{tabular}}

    \vspace{12pt}

    \centering
    \textbf{(d) Overall Performance} \\
    \vspace{2pt}
    \resizebox{\linewidth}{!}{
    \begin{tabular}{l|c}
        \toprule
        \textbf{Model} & \textbf{Global Average} \\
        \midrule
        Gemini-3-Flash (No-Think) & \cellcolor{gray!15}\textbf{89.1} \\
        Qwen3.5-397B (Thinking) & 83.1 \\
        Gemma-4-31B (Instruct) & 80.7 \\
        GLM-5 (Thinking) & 79.9 \\
        Qwen3.5-397B (Instruct) & 79.8 \\
        Gemma-4-31B (Thinking) & 79.6 \\
        GLM-4.7 (Thinking) & 74.3 \\
        DeepSeek-V3.2-Exp (Think) & 73.4 \\
        Kimi-K2 (Thinking) & 71.9 \\
        GLM-5 (Instruct) & 71.6 \\
        DeepSeek-V3.2-Exp & 71.4 \\
        Qwen3.5-35B (Thinking) & 68.7 \\
        Gemma-3-27B (Instruct) & 67.3 \\
        Qwen3-235B (Thinking) & 66.5 \\
        GPT-OSS-120B (High) & 63.7 \\
        Kimi-K2 & 62.5 \\
        Qwen3.5-35B (Instruct) & 62.5 \\
        MiniMax-M2.5 & 62.3 \\
        Qwen3-235B (Instruct) & 56.4 \\
        MiMo-V2-Flash & 56.1 \\
        Qwen3-30B (Instruct) & 47.3 \\
        Nemotron-3-Nano & 28.4 \\
        \bottomrule
    \end{tabular}}
\end{minipage}
\label{tab:judge_score_quadrant_breakdown}
\vspace*{-0.4cm}
\end{table}

\clearpage
\section{Robustness of Answer Language Analysis Across Models} 
\label{sec:mt_aime_reas}
\begin{figure*}[htbp]
    \centering
    \begin{subfigure}[b]{0.48\textwidth}
        \centering
        \includegraphics[width=\textwidth]{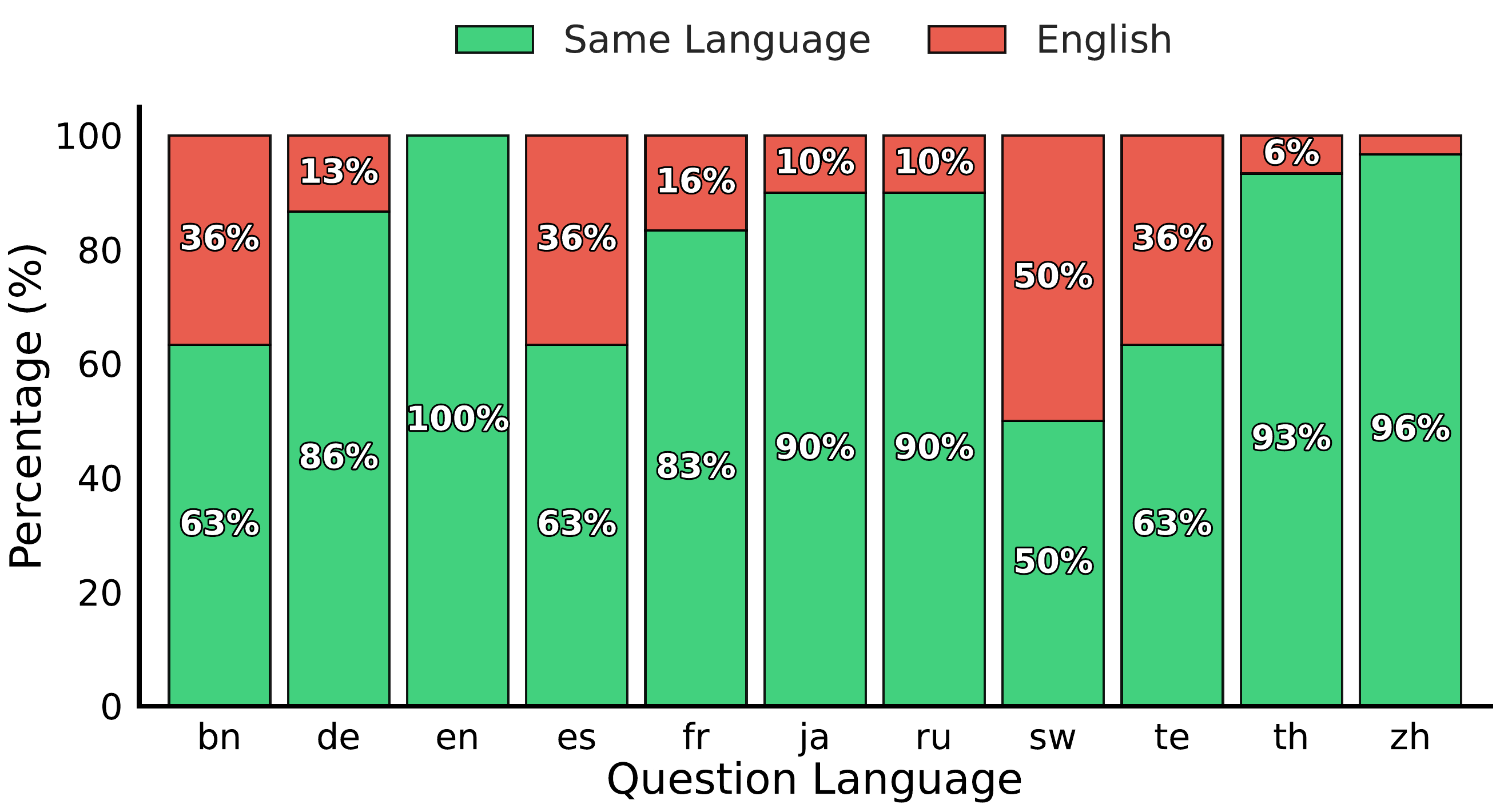}
        \caption{\textbf{Answer language}: the language the model Qwen-3-32B answers in.}
    \end{subfigure}
    \hfill
    \begin{subfigure}[b]{0.48\textwidth}
        \centering
        \includegraphics[width=\textwidth]{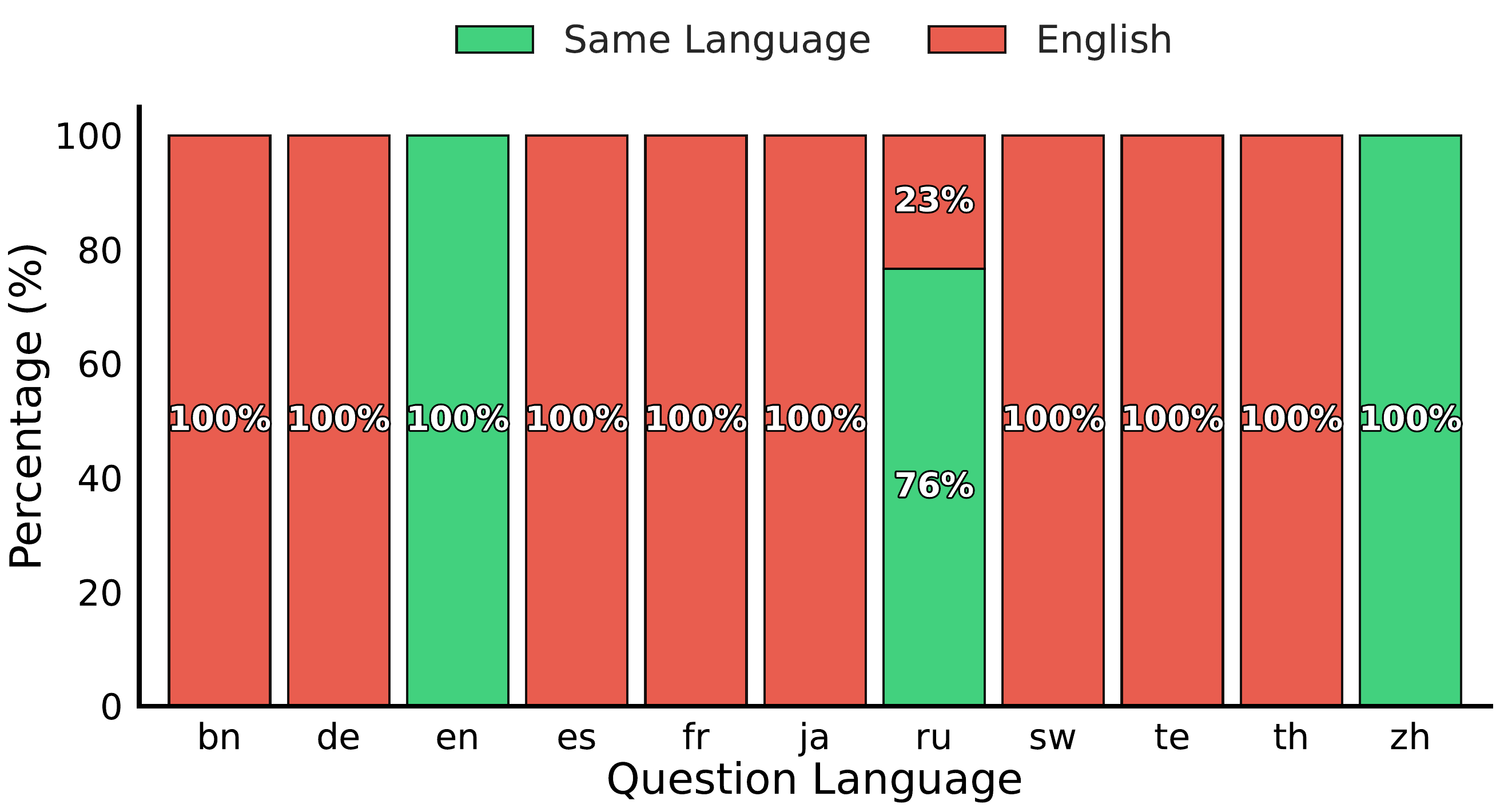}
        \caption{\textbf{Reasoning language}: the language the model Qwen-3-32B reasons in.}
    \end{subfigure}

    \vspace{0.5em} 

    \begin{subfigure}[b]{0.48\textwidth}
        \centering
        \includegraphics[width=\textwidth]{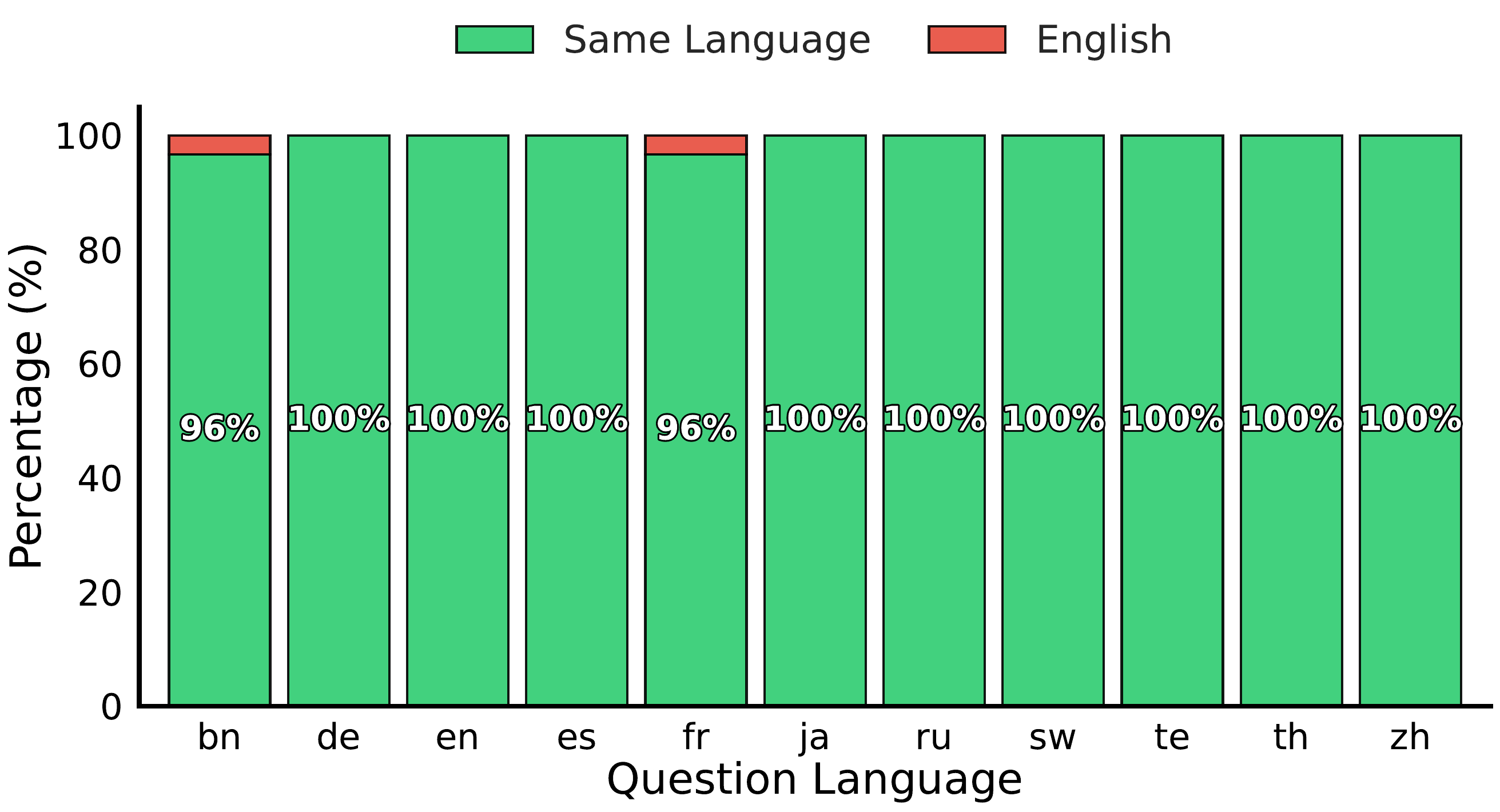}
        \caption{\textbf{Answer language}: the language the model Qwen-3-235B-A22B-Thinking-2507 answers in.}
    \end{subfigure}
    \hfill
    \begin{subfigure}[b]{0.48\textwidth}
        \centering
        \includegraphics[width=\textwidth]{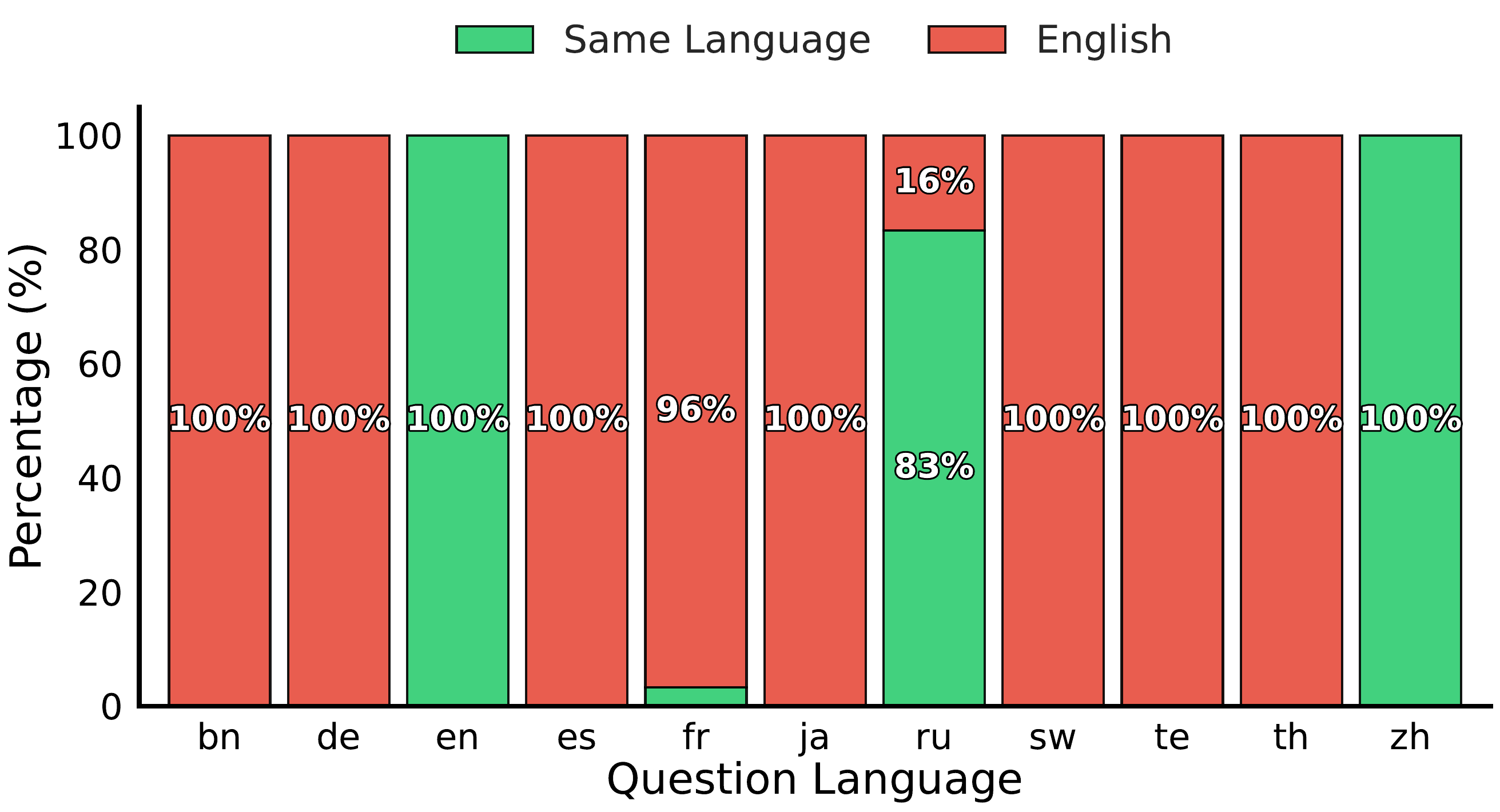}
        \caption{\textbf{Reasoning language}: the language the model Qwen-3-235B-A22B-Thinking-2507 reasons in.}
    \end{subfigure}

    \begin{subfigure}[b]{0.48\textwidth}
        \centering
        \includegraphics[width=\textwidth]{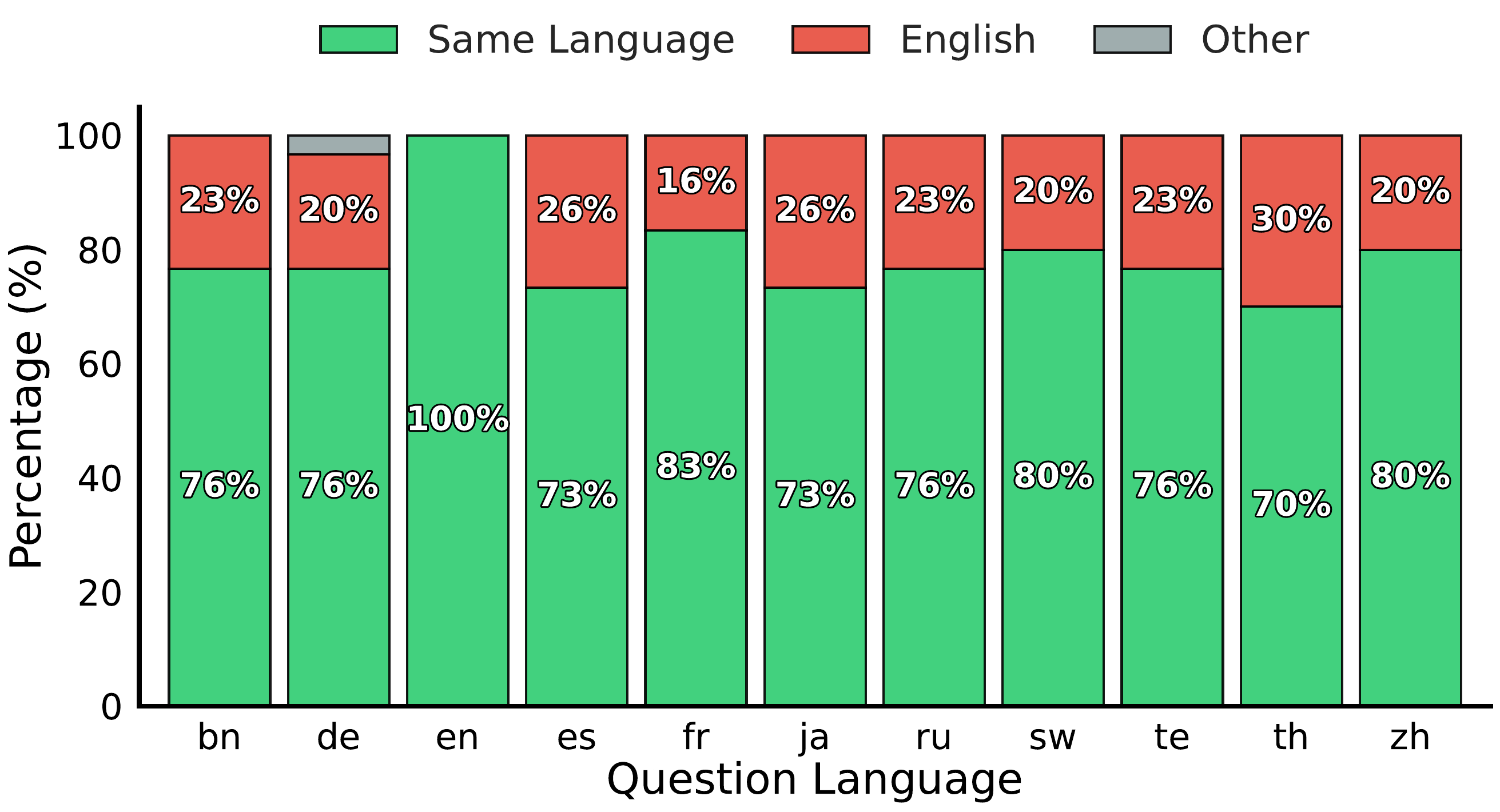}
        \caption{\textbf{Answer language}: the language the model GPT-OSS-20B answers in.}
    \end{subfigure}
    \hfill
    \begin{subfigure}[b]{0.48\textwidth}
        \centering
        \includegraphics[width=\textwidth]{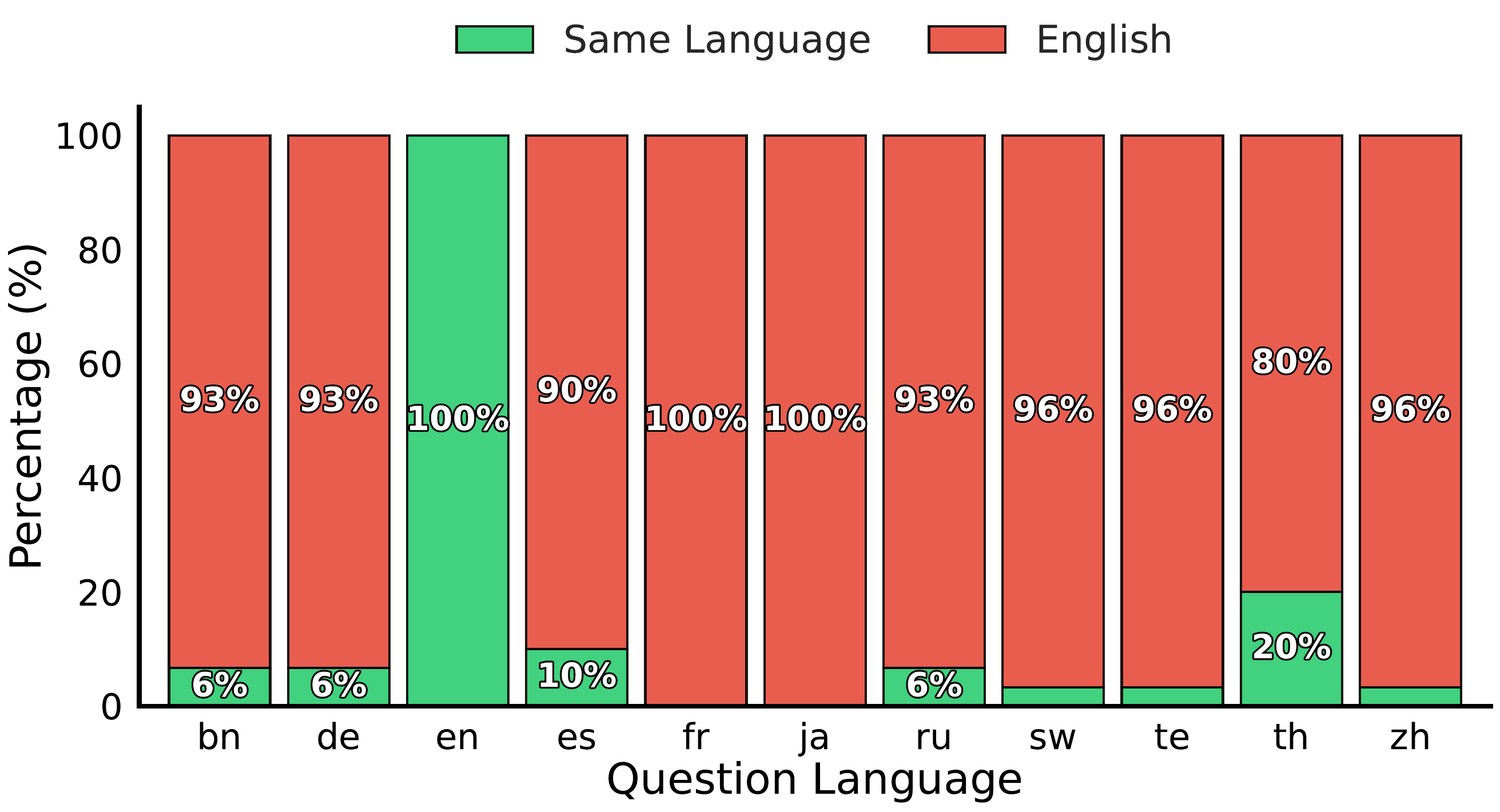}
        \caption{\textbf{Reasoning language}: the language the model GPT-OSS-20B reasons in.}
    \end{subfigure}

        \begin{subfigure}[b]{0.48\textwidth}
        \centering
        \includegraphics[width=\textwidth]{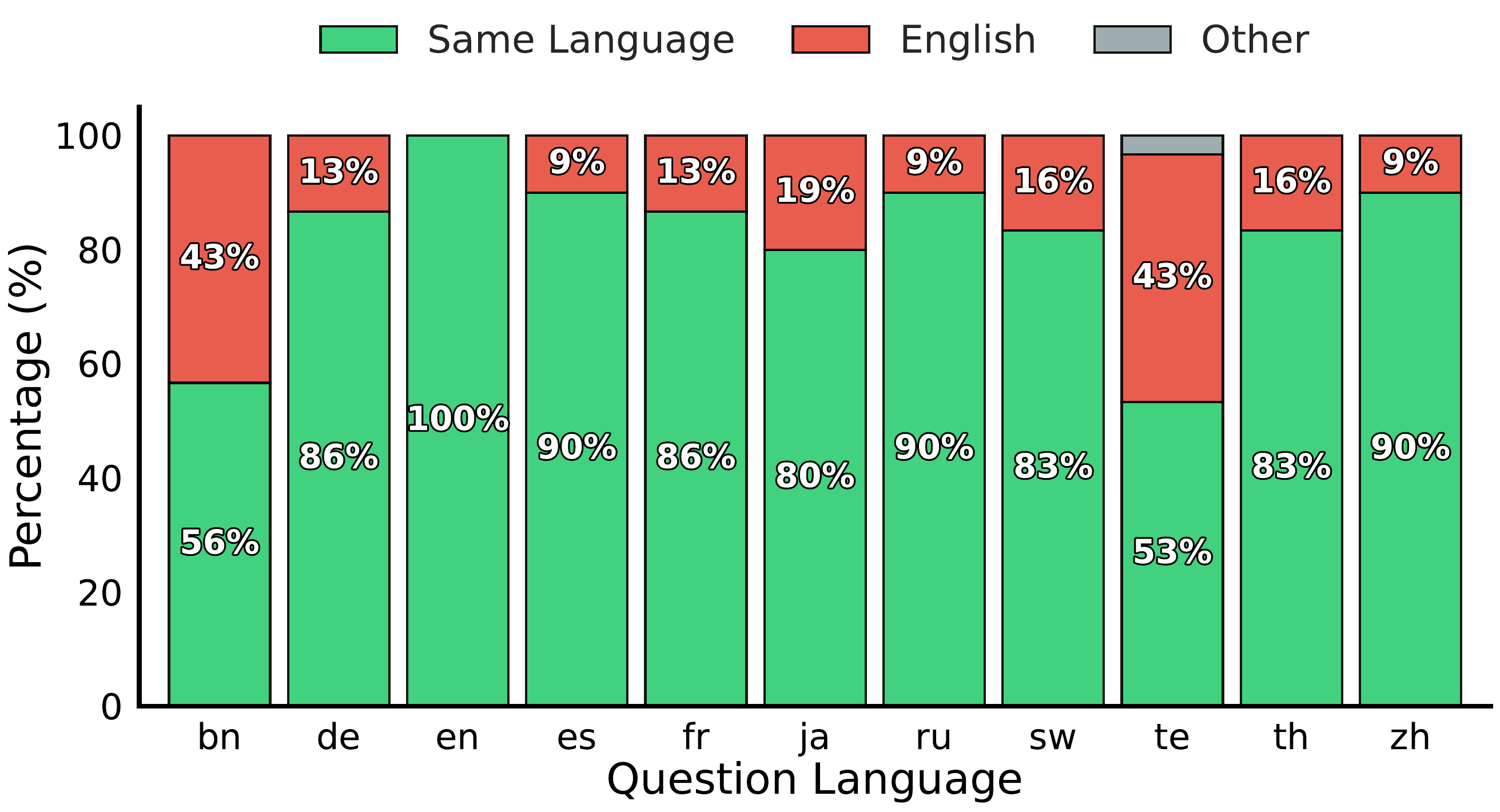}
        \caption{\textbf{Answer language}: the language the model GPT-OSS-120B answers in.}
    \end{subfigure}
    \hfill
    \begin{subfigure}[b]{0.48\textwidth}
        \centering
        \includegraphics[width=\textwidth]{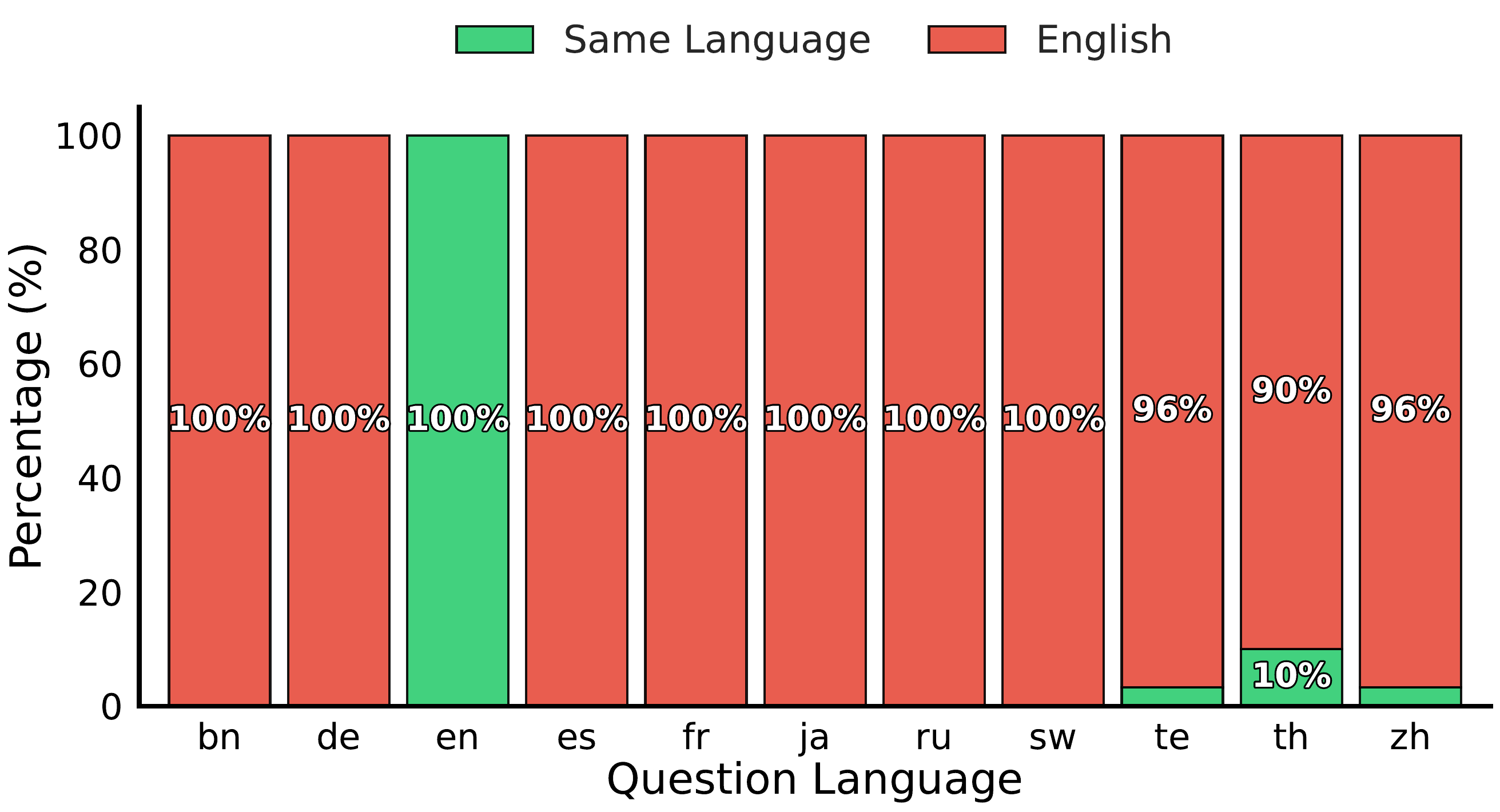}
        \caption{\textbf{Reasoning language}: the language the model GPT-OSS-120B reasons in.}
    \end{subfigure}

    \caption{Qwen-3 and GPT-OSS models default to English reasoning on MT-AIME24. We analyze the language used for answers (left column) and reasoning traces (right column) across 11 languages. (a-b) Qwen-3-32B answers in the target language inconsistently but reasons almost entirely in English. (c-d) Qwen-3-235B-Thinking answers consistently in the target language (100\%) but still reasons predominantly in English, especially for Swahili. (e-f) GPT-OSS-20B shows mixed answering behavior but reasons 93–100\% in English. (g-h) GPT-OSS-120B answers mostly in the target language but reasons almost entirely in English. These patterns confirm that mathematical reasoning occurs in English regardless of input language.}
    \label{fig:qwen_gpt}
\end{figure*}

\begin{figure*}[t!]
    \centering
    \begin{subfigure}[b]{0.48\textwidth}
        \centering
        \includegraphics[width=\textwidth]{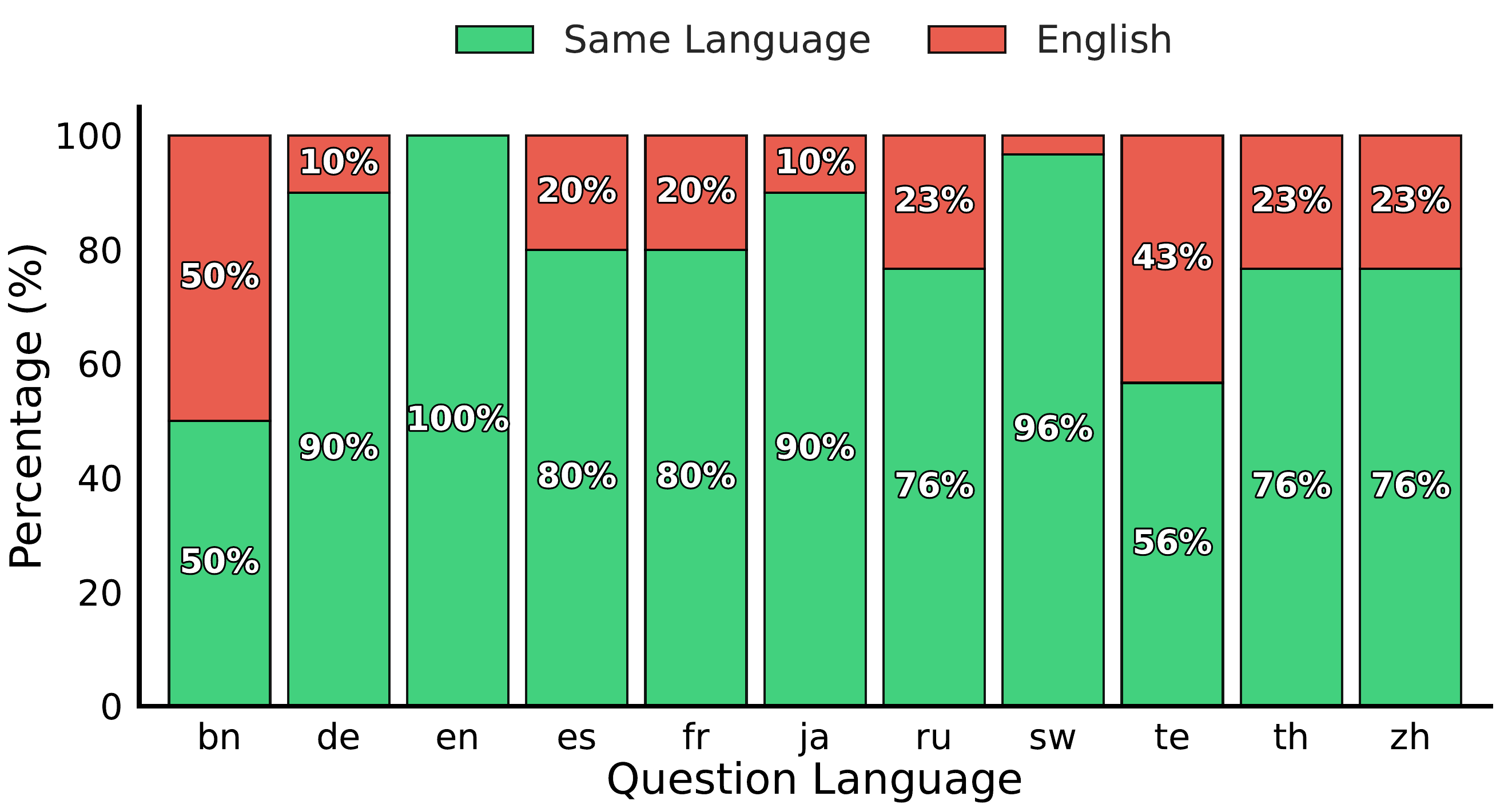}
        \caption{\textbf{Answer language}: the language the model GLM 4.7~\cite{5team2025glm45agenticreasoningcoding} answers in.}
    \end{subfigure}
    \hfill
    \begin{subfigure}[b]{0.48\textwidth}
        \centering
        \includegraphics[width=\textwidth]{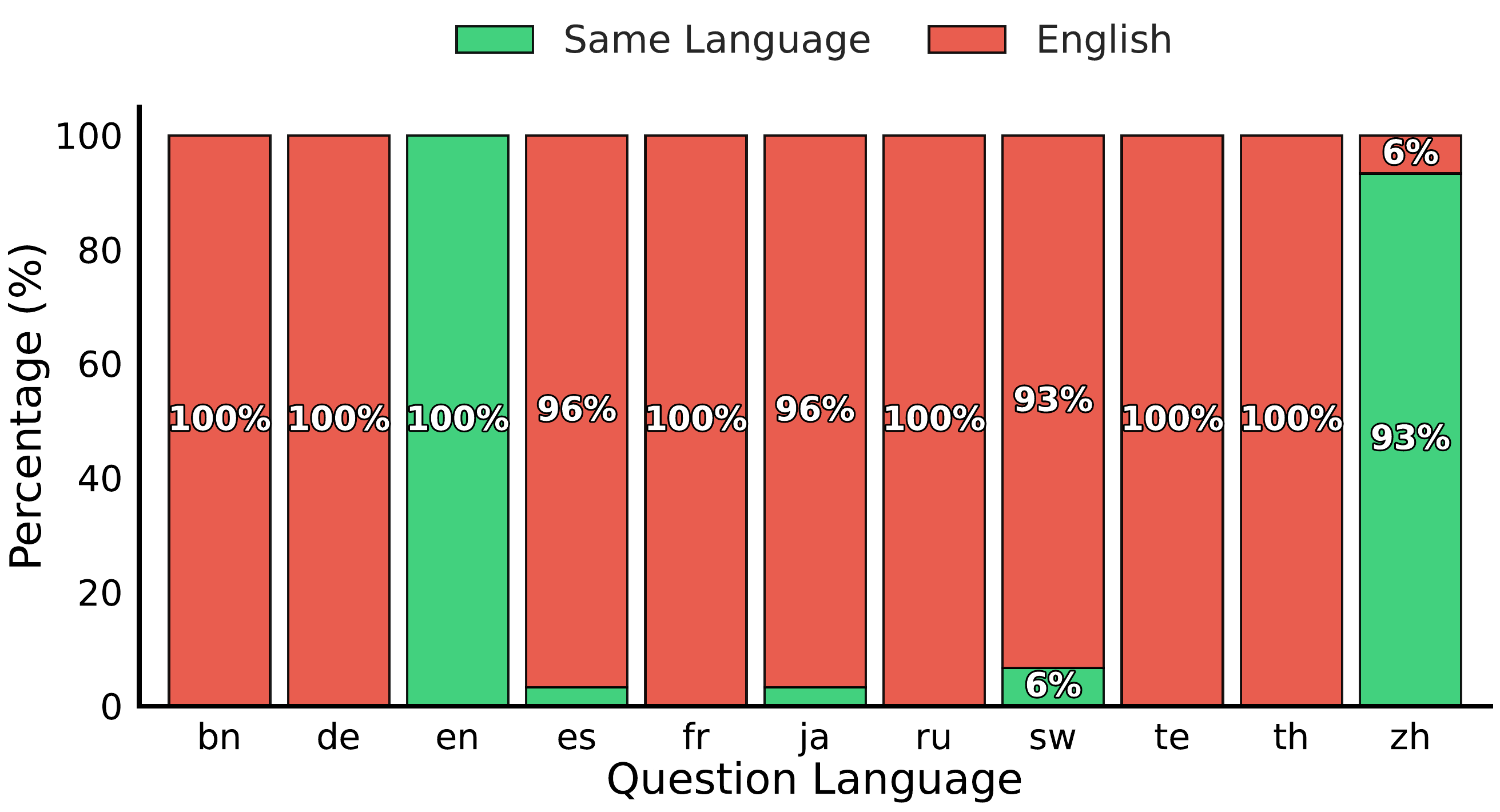}
        \caption{\textbf{Reasoning language}: the language the model GLM 4.7~\cite{5team2025glm45agenticreasoningcoding} reasons in.}
    \end{subfigure}

    \vspace{0.5em} 

    \begin{subfigure}[b]{0.48\textwidth}
        \centering
        \includegraphics[width=\textwidth]{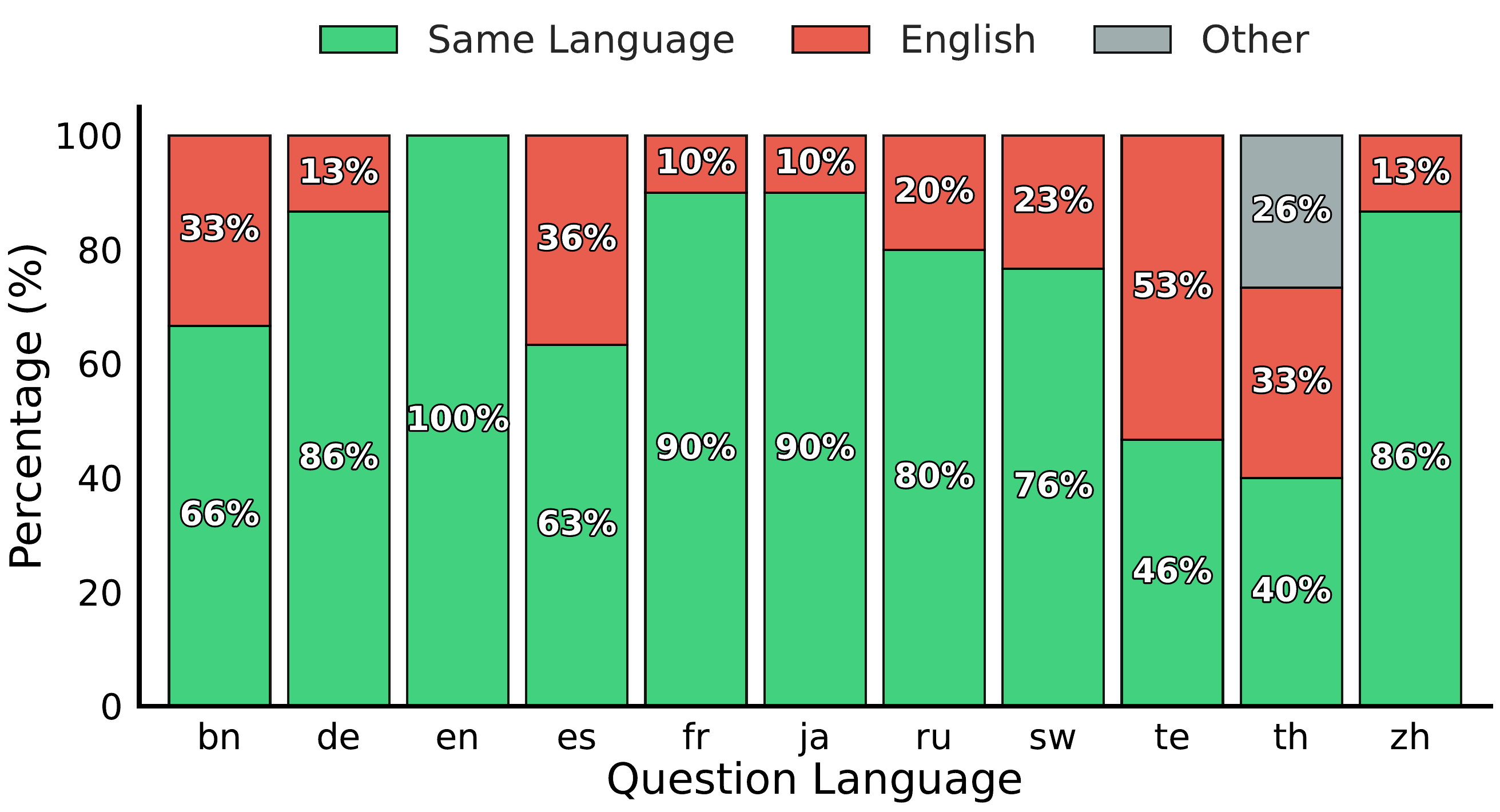}
        \caption{\textbf{Answer language}: the language the model Mimo-V2-Flash answers in.}
    \end{subfigure}
    \hfill
    \begin{subfigure}[b]{0.48\textwidth}
        \centering
        \includegraphics[width=\textwidth]{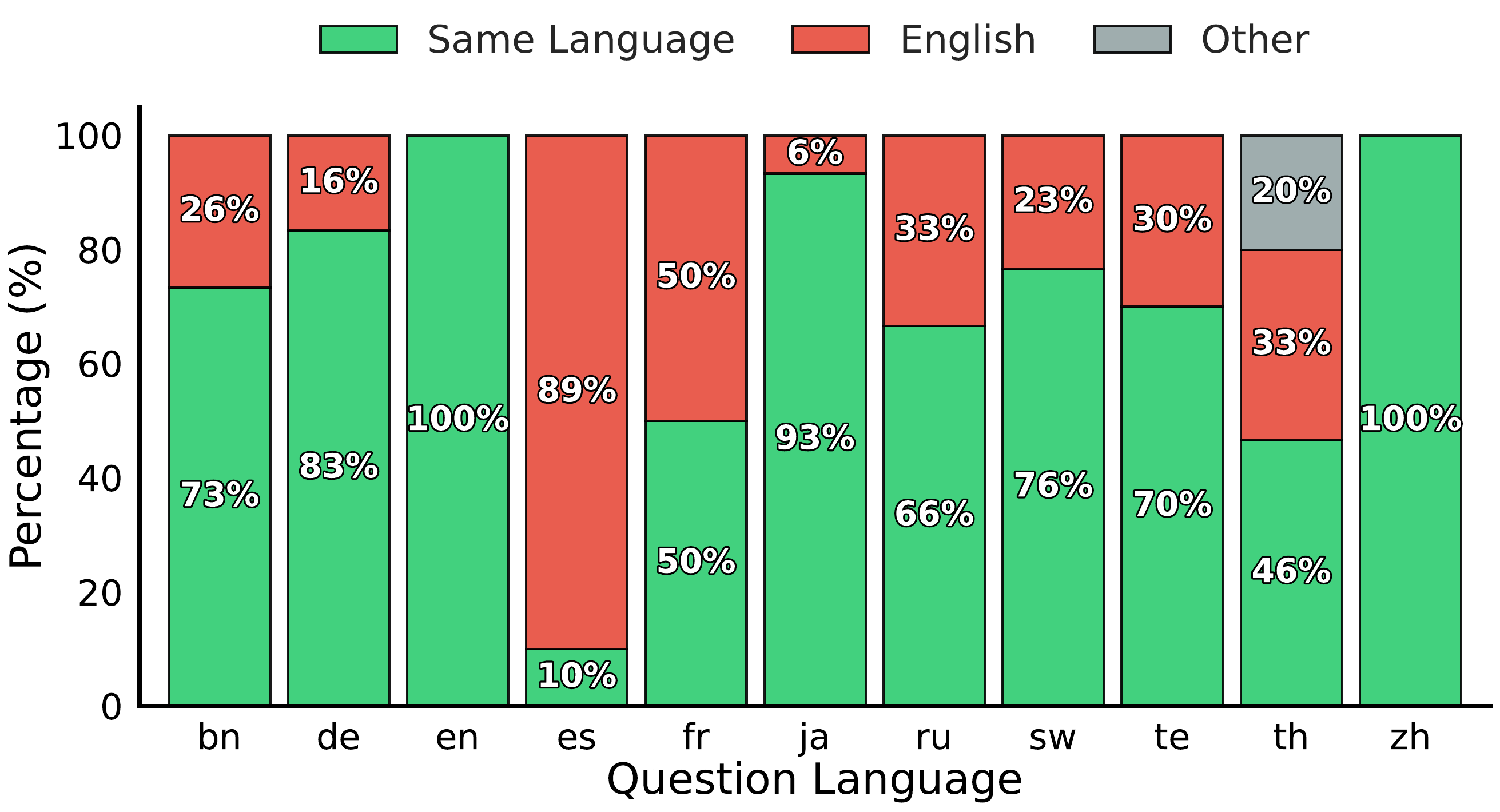}
        \caption{\textbf{Reasoning language}: the language the model Mimo-V2-Flash reasons in.}
    \end{subfigure}

    \caption{GLM-4.7 and MiMo-V2-Flash show contrasting reasoning language patterns on MT-AIME24. (a-b) GLM-4.7 answers in mixed languages across different inputs but reasons overwhelmingly in English (93–100\% for most languages), with slight exceptions for Swahili and Chinese. (c-d) MiMo-V2-Flash displays the most diverse reasoning behavior: it reasons natively 33–100\% of the time depending on the language, making it an outlier among tested models. However, this native reasoning does not translate to better benchmark performance, further suggesting that MT-AIME24 measures reasoning ability rather than multilingual proficiency.}
    \label{fig:glm_mimo}
\end{figure*}

\begin{figure*}[htbp]
    \centering
    \begin{subfigure}[b]{0.45\textwidth}
        \centering
        \includegraphics[width=\textwidth]{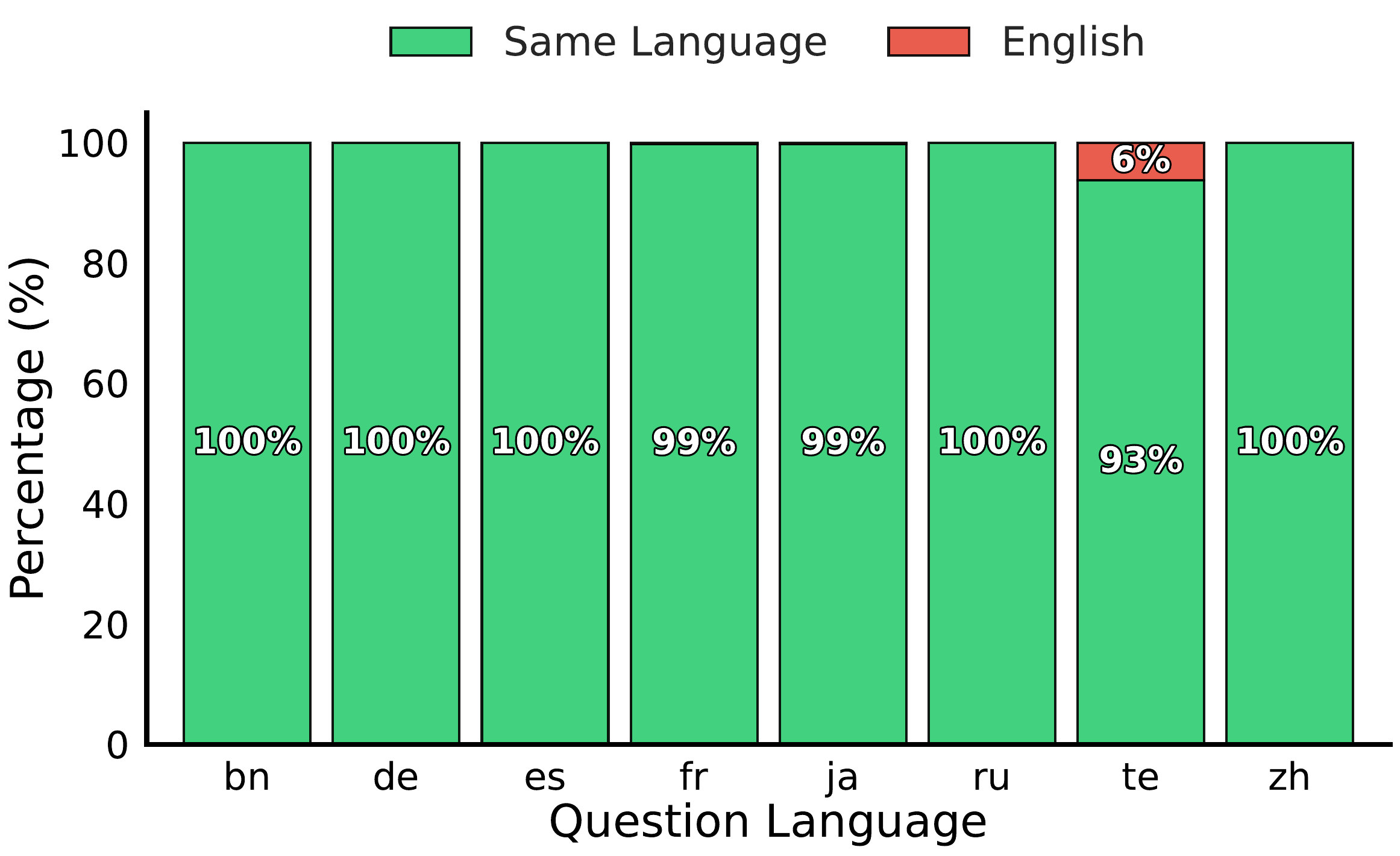}
        \caption{\textbf{Answer language}: the language the 8Qwen-3-32B answers in.}
    \end{subfigure}
    \hfill
    \begin{subfigure}[b]{0.45\textwidth}
        \centering
        \includegraphics[width=\textwidth]{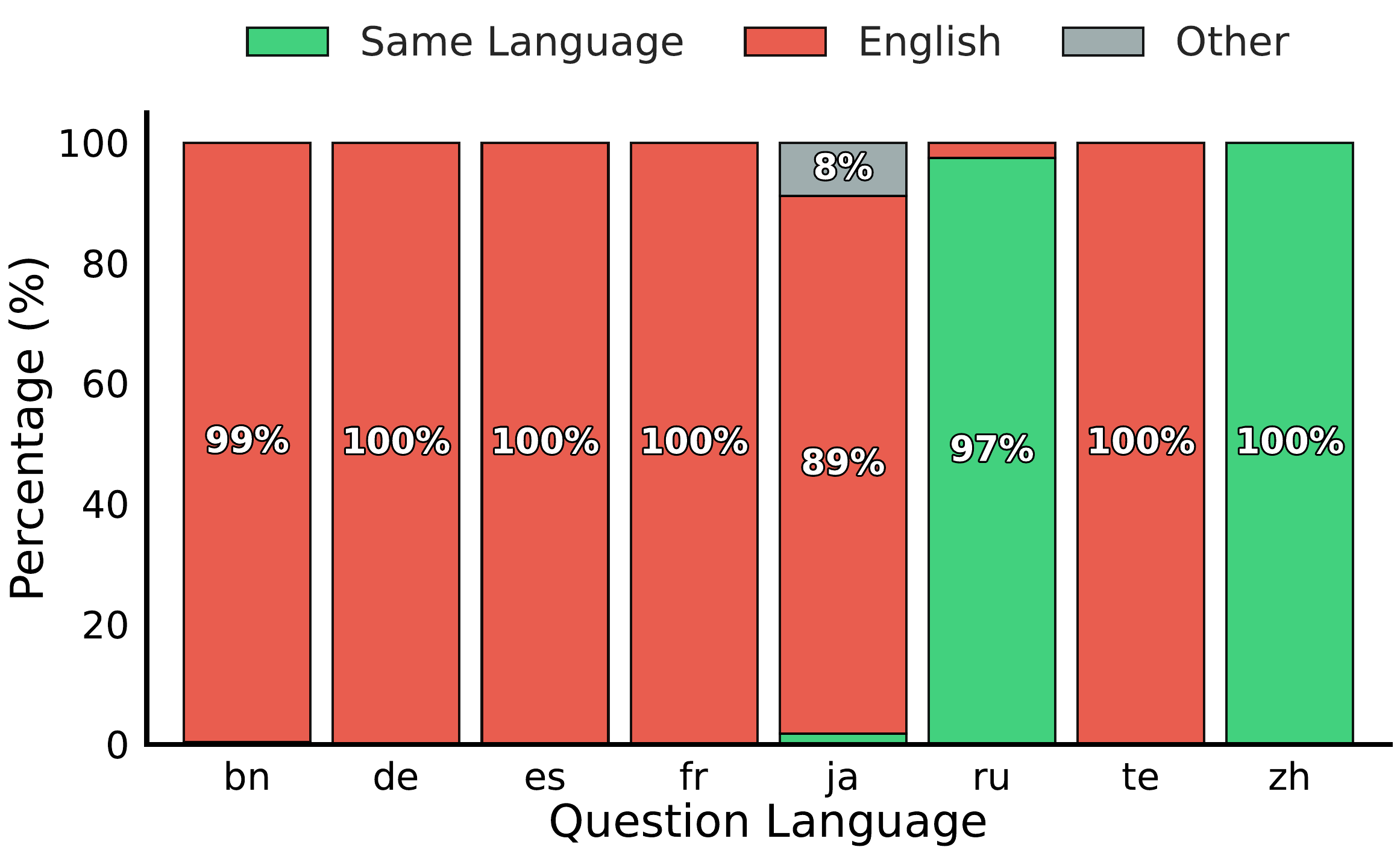}
        \caption{\textbf{Reasoning language}: the language the model Qwen-3-32B reasons in.}
    \end{subfigure}

    \vspace{0.5em} 

    \begin{subfigure}[b]{0.45\textwidth}
        \centering
        \includegraphics[width=\textwidth]{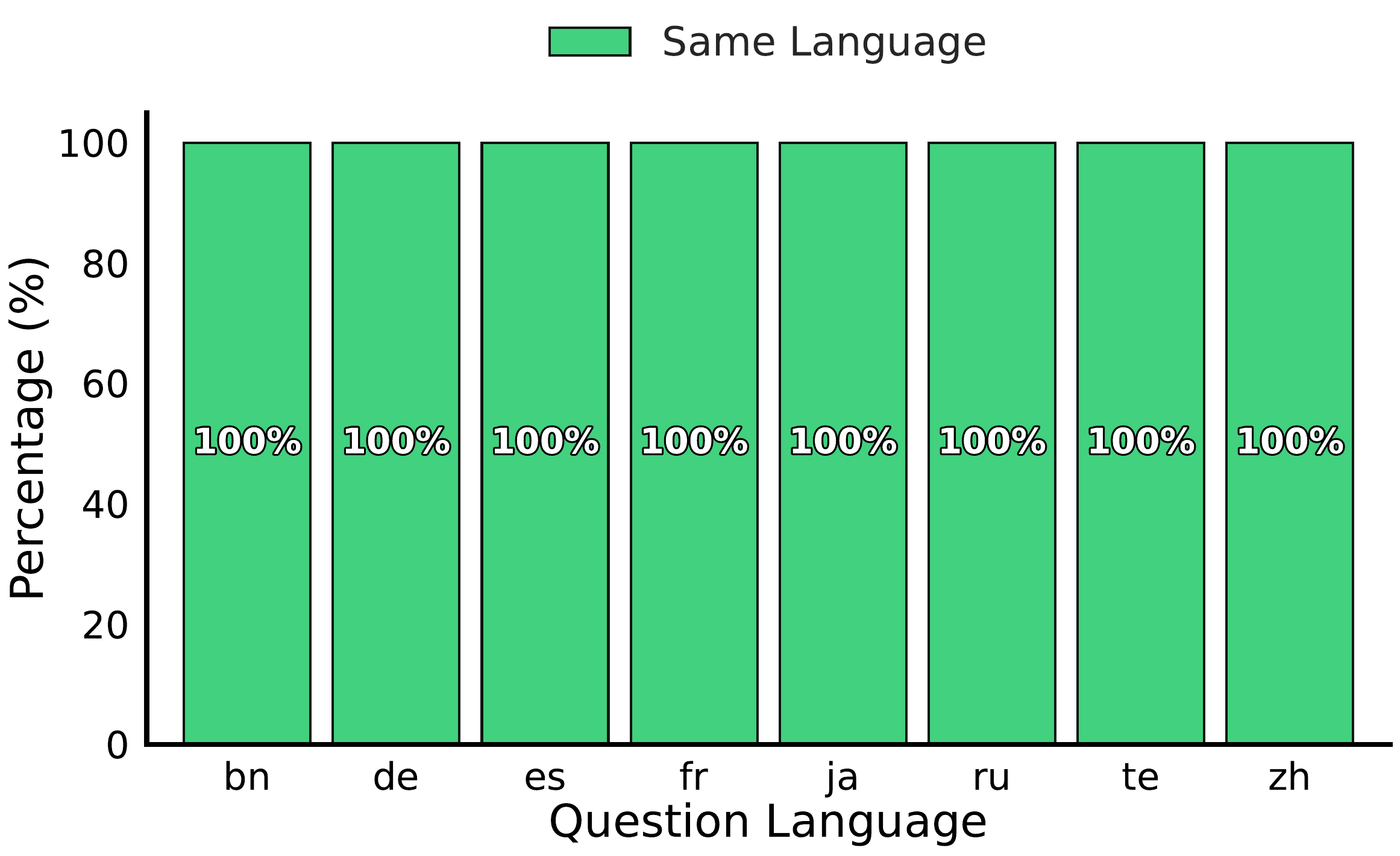}
        \caption{\textbf{Answer language}: the language the model Qwen-3-235B-A22B-Thinking-2507 answers in.}
    \end{subfigure}
    \hfill
    \begin{subfigure}[b]{0.45\textwidth}
        \centering
        \includegraphics[width=\textwidth]{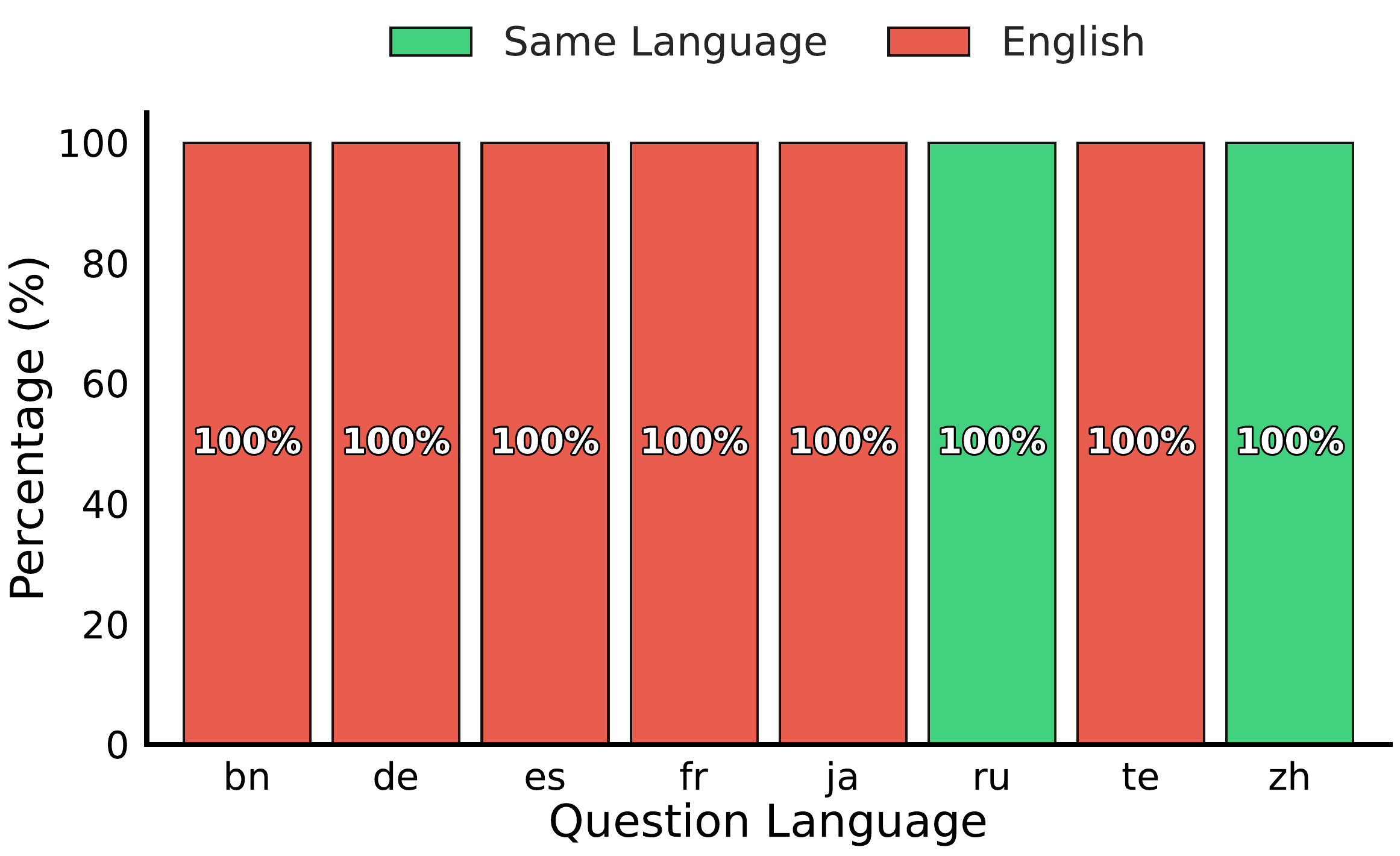}
        \caption{\textbf{Reasoning language}: the language the model Qwen-3-235B-A22B-Thinking-2507 reasons in.}
    \end{subfigure}
\vspace{0.5em} 
    \begin{subfigure}[b]{0.45\textwidth}
        \centering
        \includegraphics[width=\textwidth]{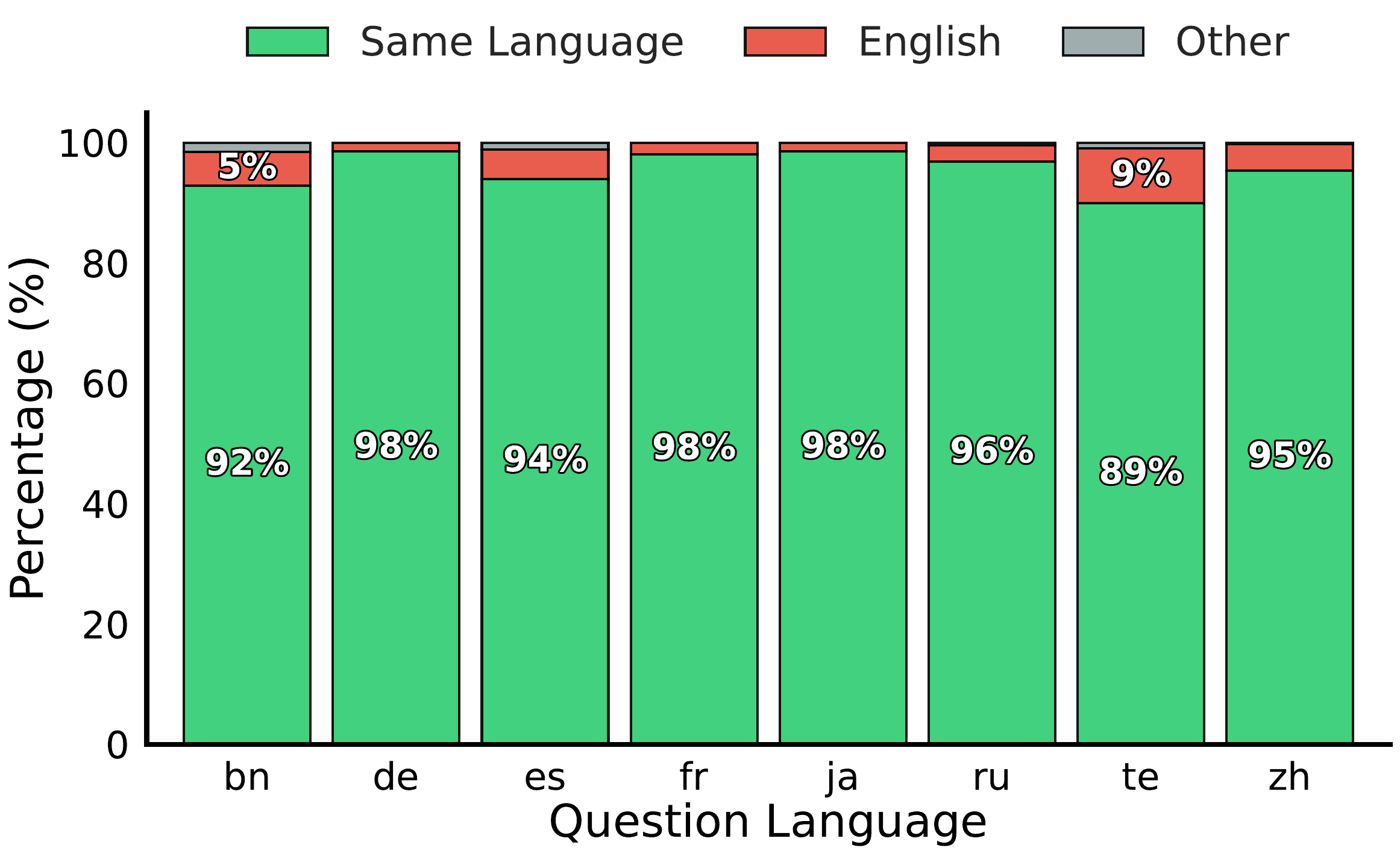}
        \caption{\textbf{Answer language}: the language the model GPT-OSS-20B answers in.}
    \end{subfigure}
    \hfill
    \begin{subfigure}[b]{0.45\textwidth}
        \centering
        \includegraphics[width=\textwidth]{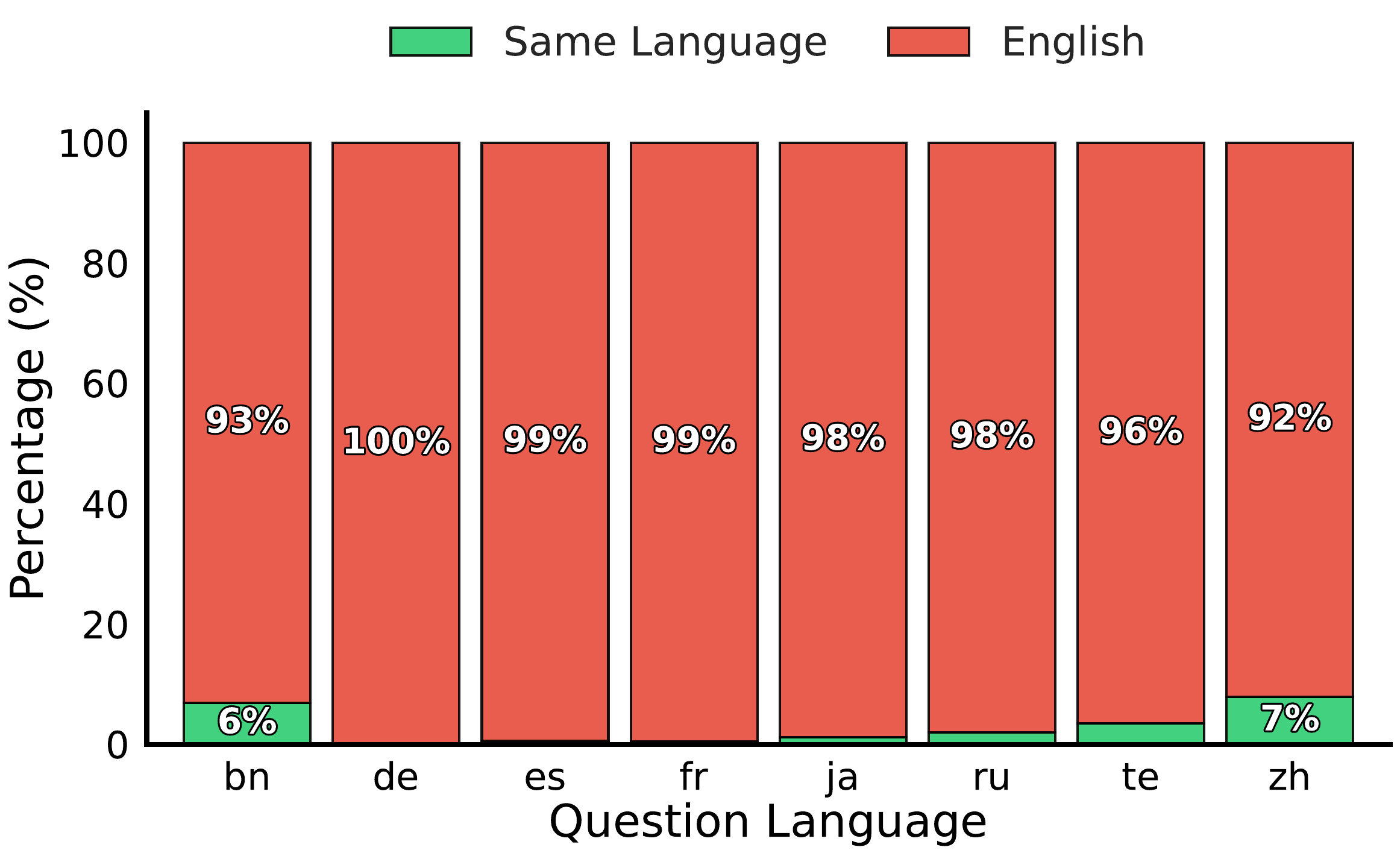}
        \caption{\textbf{Reasoning language}: the language the model GPT-OSS-20B reasons in.}
    \end{subfigure}
\vspace{0.5em} 
        \begin{subfigure}[b]{0.45\textwidth}
        \centering
        \includegraphics[width=\textwidth]{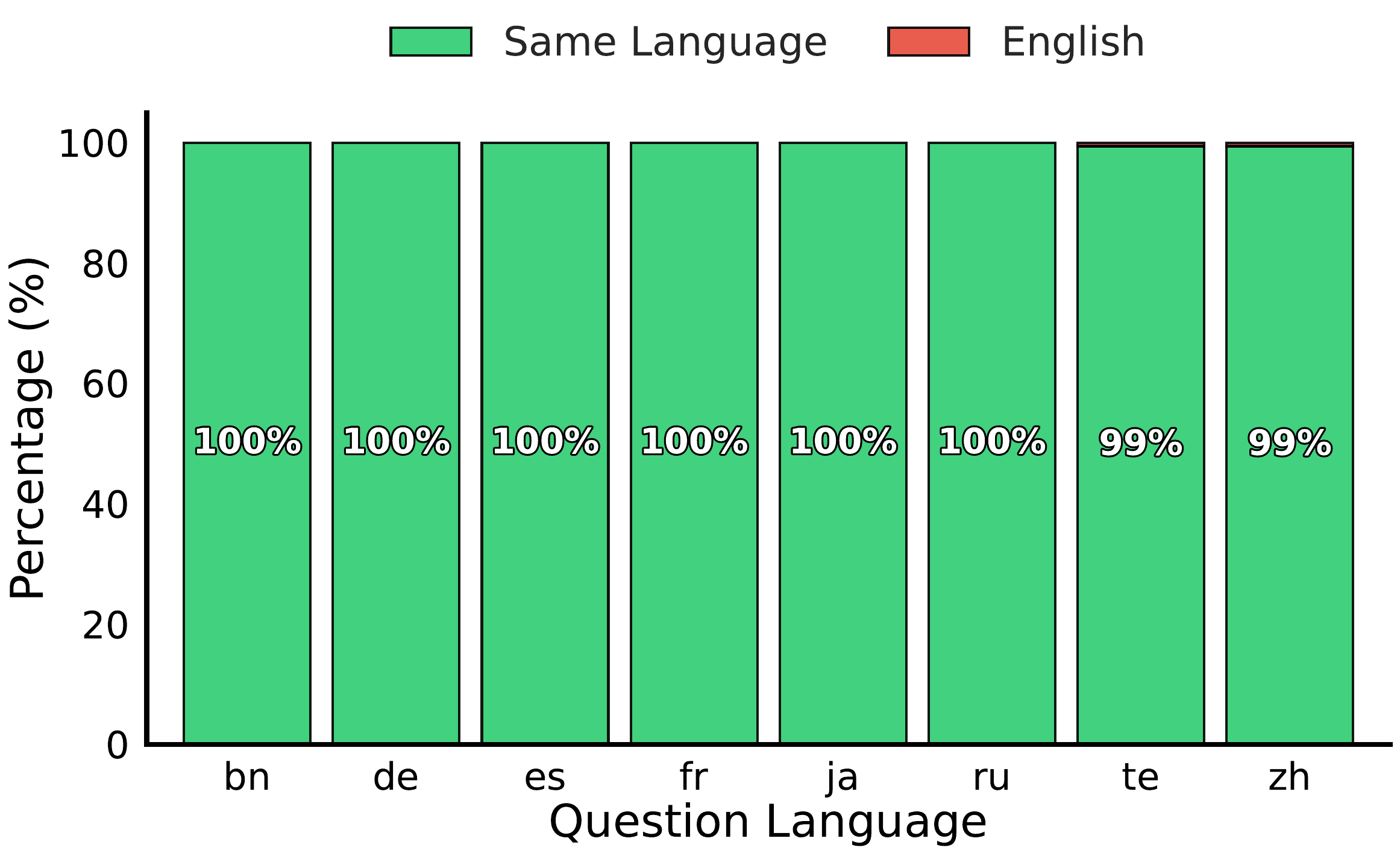}
        \caption{\textbf{Answer language}: the language the model GPT-OSS-120B answers in.}
    \end{subfigure}
    \hfill
    \begin{subfigure}[b]{0.45\textwidth}
        \centering
        \includegraphics[width=\textwidth]{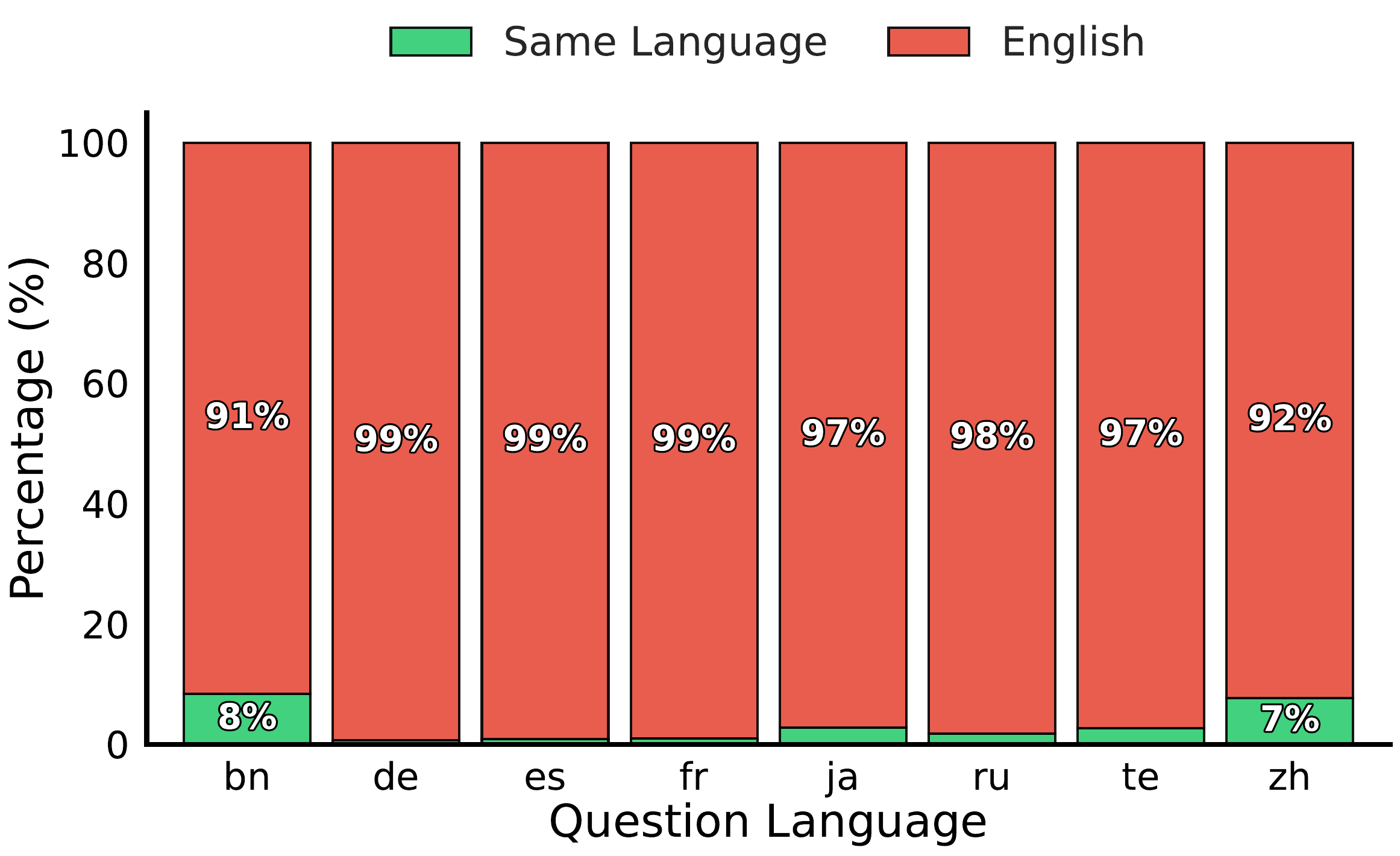}
        \caption{\textbf{Reasoning language}: the language the model GPT-OSS-120B reasons in.}
    \end{subfigure}
    \caption{Qwen-3 and GPT-OSS models reason almost entirely in English on INCLUDE despite answering in target languages. (a-b) Qwen-3-32B answers consistently in the target language but reasons in English 89–100\% of the time. (c-d) Qwen-3-235B-Thinking achieves 100\% target-language answering and 100\% target-language reasoning—a unique pattern among tested models. (e-f) GPT-OSS-20B answers mostly in the target language (89–98\%) but reasons in English 92–100\% of the time. (g-h) GPT-OSS-120B shows similar patterns with 91–99\% English reasoning. The disconnect between answer language and reasoning language explains why INCLUDE performance tracks English knowledge benchmarks.}
    \label{fig:qwen_gpt_include}
\end{figure*}

\begin{figure*}[t!]
    \centering
    \begin{subfigure}[b]{0.48\textwidth}
        \centering
        \includegraphics[width=\textwidth]{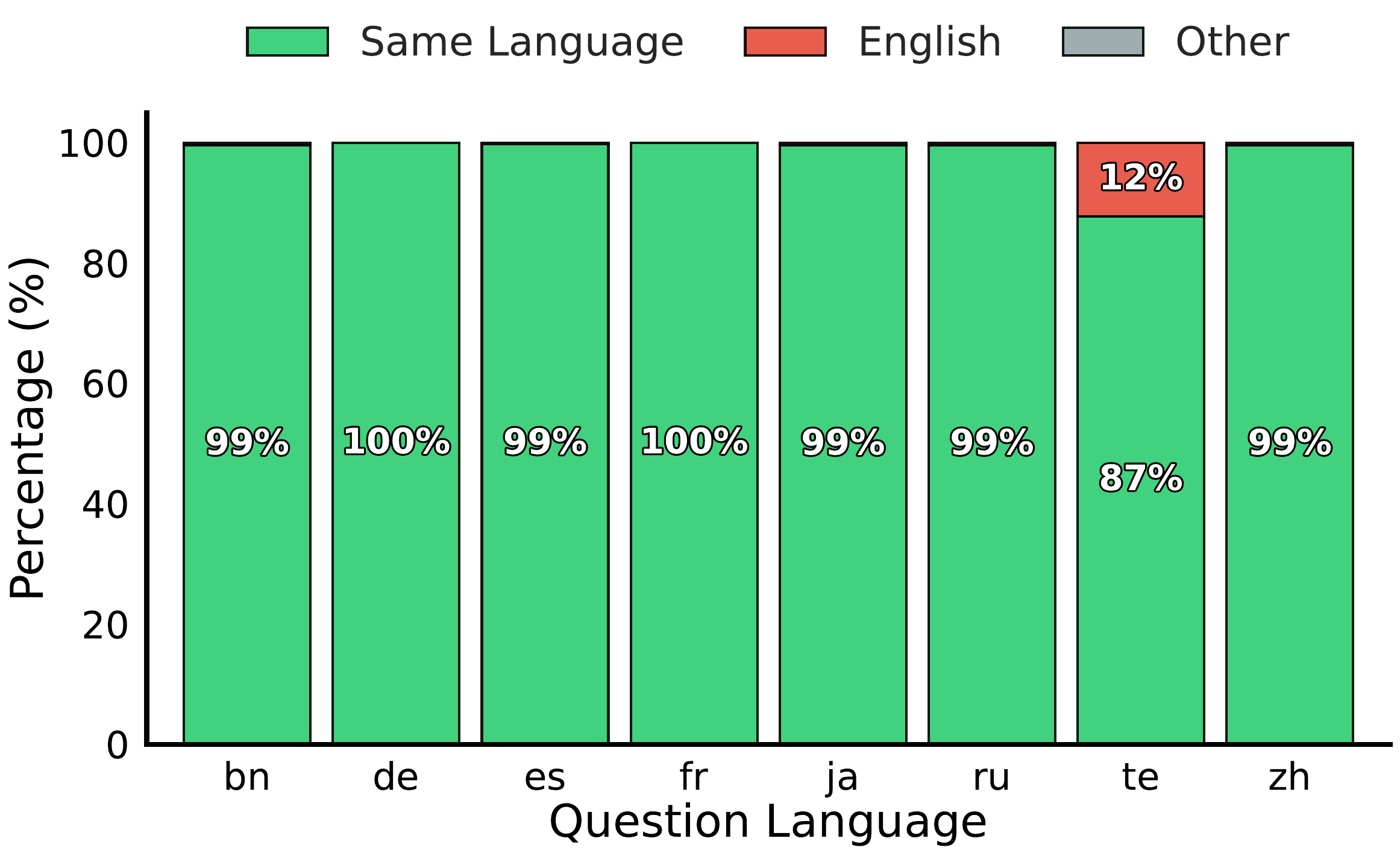}
        \caption{\textbf{Answer language}: the language the model GLM 4.7~\cite{5team2025glm45agenticreasoningcoding} answers in.}
    \end{subfigure}
    \hfill
    \begin{subfigure}[b]{0.48\textwidth}
        \centering
        \includegraphics[width=\textwidth]{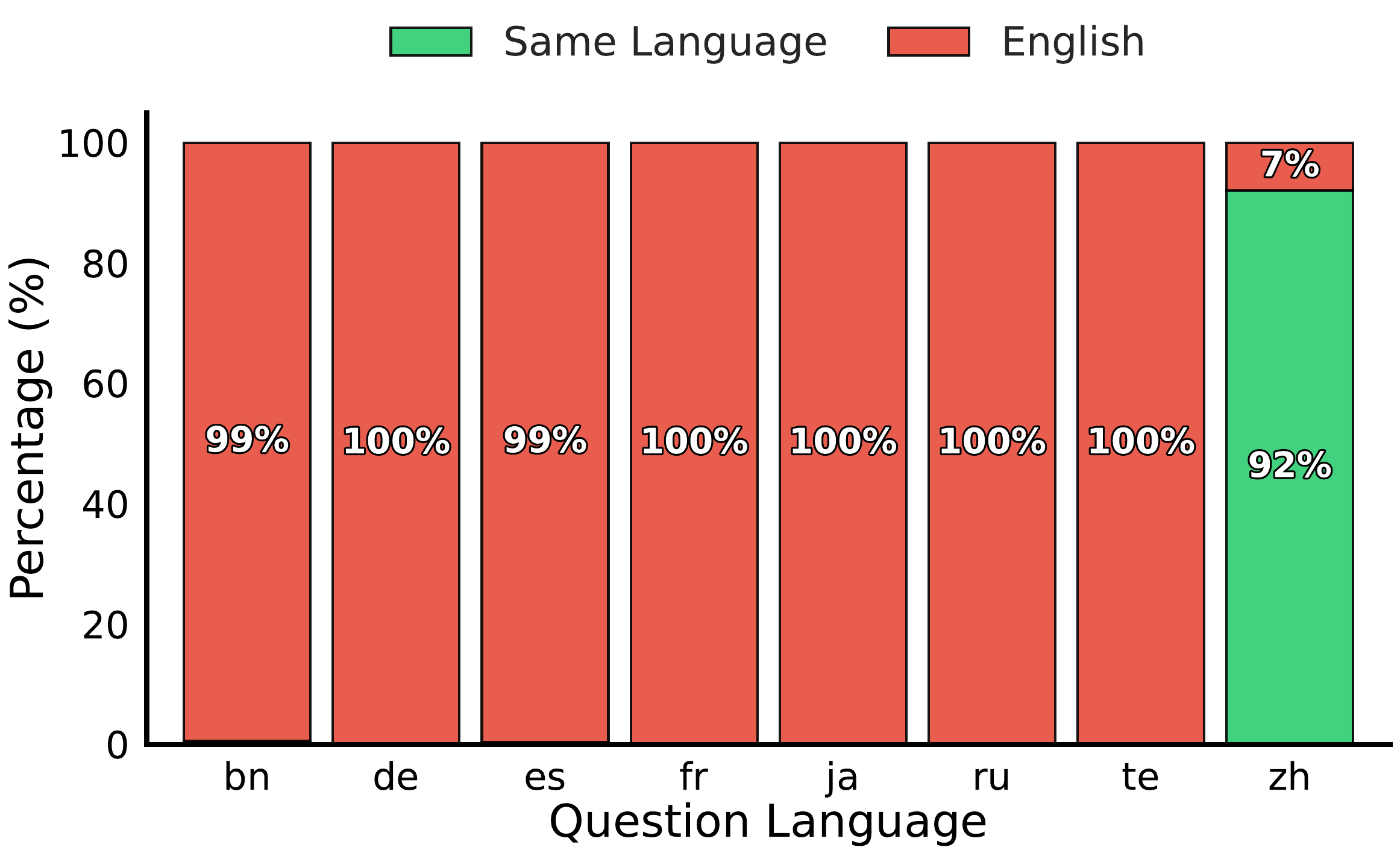}
        \caption{\textbf{Reasoning language}: the language the model GLM 4.7~\cite{5team2025glm45agenticreasoningcoding} reasons in.}
    \end{subfigure}

    \vspace{0.5em} 

    \begin{subfigure}[b]{0.48\textwidth}
        \centering
        \includegraphics[width=\textwidth]{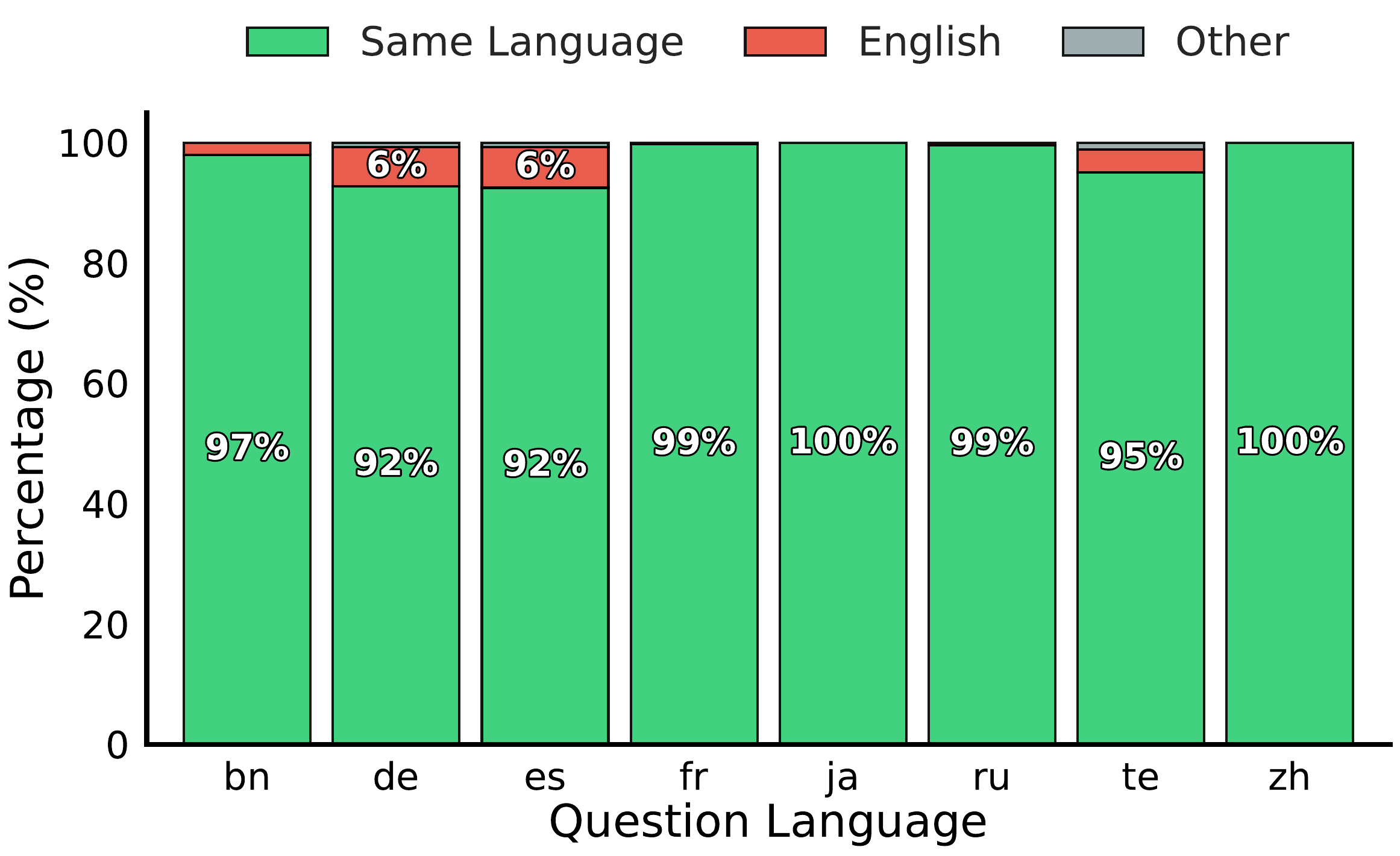}
        \caption{\textbf{Answer language}: the language the model Mimo-V2-Flash answers in.}
    \end{subfigure}
    \hfill
    \begin{subfigure}[b]{0.48\textwidth}
        \centering
        \includegraphics[width=\textwidth]{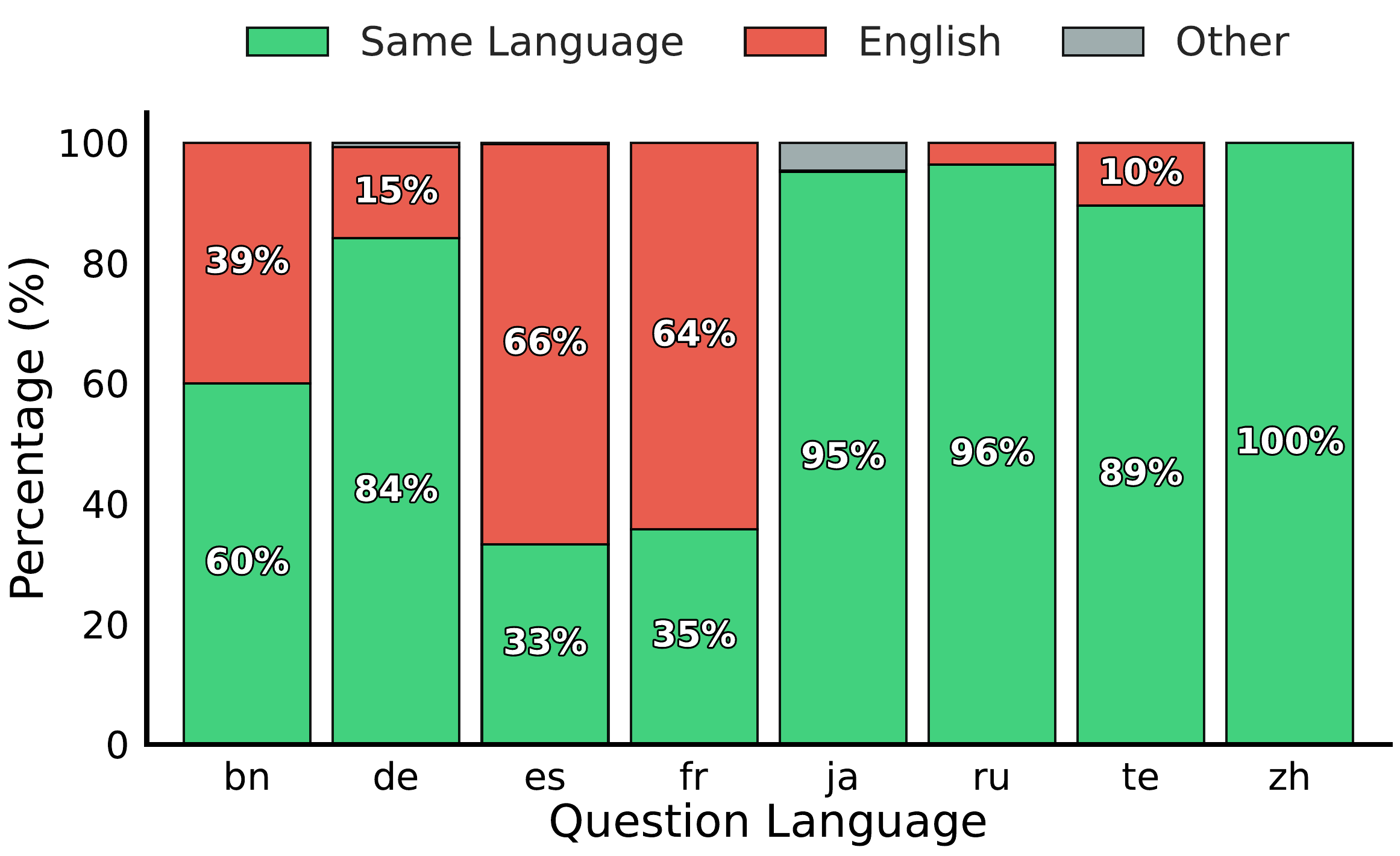}
        \caption{\textbf{Reasoning language}: the language the model Mimo-V2-Flash reasons in.}
    \end{subfigure}

    \caption{GLM-4.7 reasons in English while MiMo-V2-Flash shows mixed reasoning patterns on INCLUDE. (a-b) GLM-4.7 answers in the target language 87–100\% of the time and reasons in English 92–100\% of the time across languages. Russian (87\%) and Telugu (12\%) show the most answer-language variation. (c-d) MiMo-V2-Flash displays highly variable reasoning behavior: it reasons natively 33–100\% of the time depending on the language, with particularly high native reasoning for Telugu (100\%), Bengali (60\%), and Spanish (84\%). This variability makes MiMo-V2-Flash an interesting case study for understanding how reasoning language affects downstream task performance.}
    \label{fig:glm_mimo_include}
\end{figure*}

In this section, we provide results on the languages in which models answer and reason, extending the analysis in Section~\ref{subsec:err_tax} across models. We show results for MT-AIME24 in Figures~\ref{fig:qwen_gpt} and ~\ref{fig:glm_mimo} and Figures~\ref{fig:qwen_gpt_include} and~\ref{fig:glm_mimo_include} for Include to separate two failure modes that can confound multilingual benchmark evaluation. First, some models do not consistently answer in the language of the prompt. For example, Qwen3-32B answers in Swahili only about half of the time when prompted in Swahili. 

Second, an even stronger effect appears in the reasoning traces: most models reason predominantly in English, with only a few partial exceptions, such as Chinese and Russian for some Qwen3 models. Taken together, these results further support our claim that multilingual reasoning and general-knowledge benchmarks often measure English-centered reasoning ability more than genuine multilingual capability.

\label{sec:include_reas}

\clearpage
\section{Robustness of Error Distribution Analysis Across Models}
\label{sec:mt_aime_err}
\begin{figure*}[h]
\vspace{-0.5cm}
    \centering
    \begin{subfigure}[b]{0.48\textwidth}
        \centering
        \includegraphics[width=\textwidth]{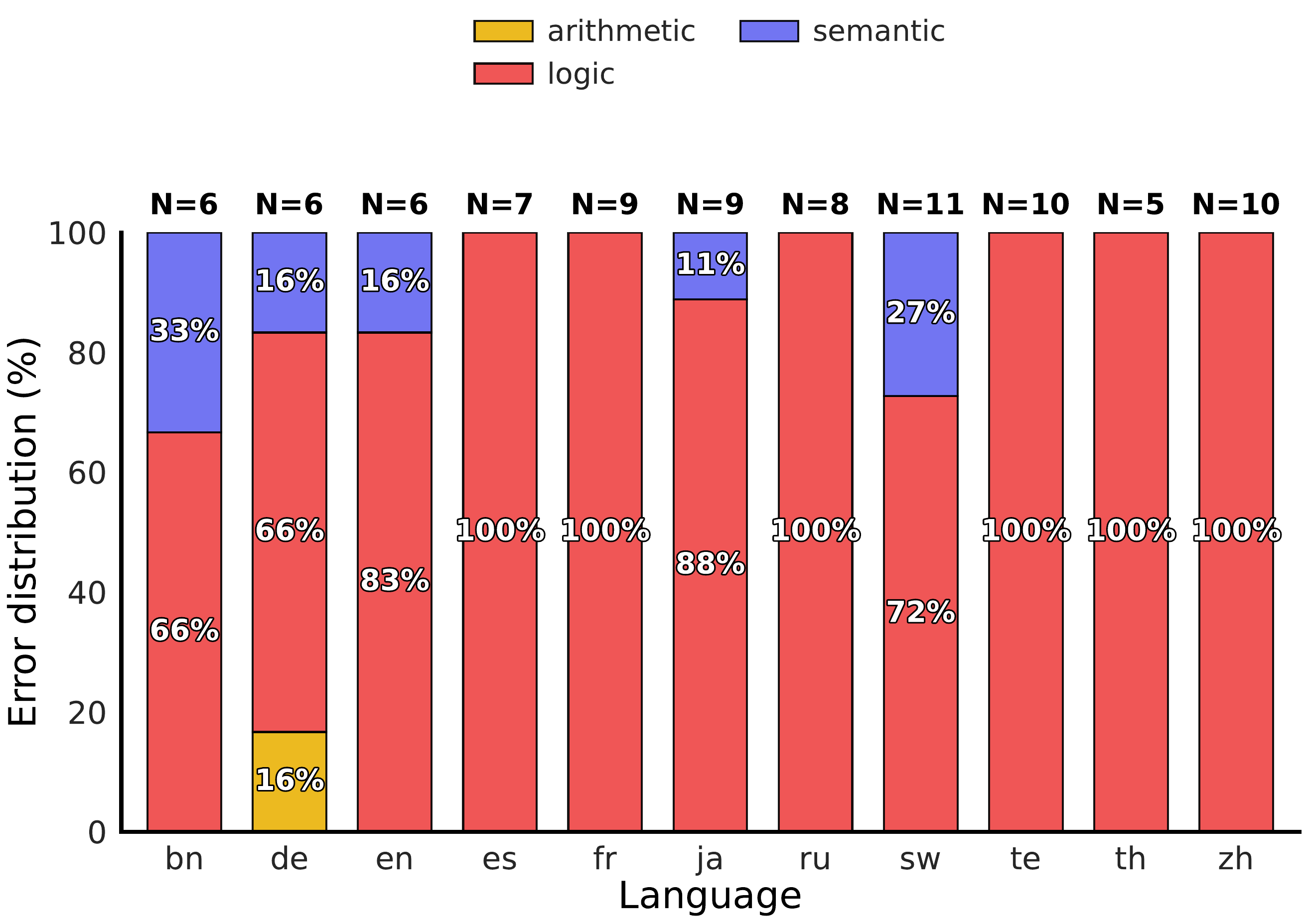}
        \caption{Error distribution for Qwen3-32B}
    \end{subfigure}
    \hfill
    \begin{subfigure}[b]{0.48\textwidth}
        \centering
        \includegraphics[width=\textwidth]{figures/err_dist/mt-aime24_error_distribution_qwen-qwen3-235b-a22b-thinking-2507_0-shot-judge=google-gemini-3-flash-preview.pdf}
        \caption{Error distribution for Qwen3-235B-A22B-Thinking-2507}
    \end{subfigure}

    \vspace{0.5em} 

    \begin{subfigure}[b]{0.48\textwidth}
        \centering
        \includegraphics[width=\textwidth]{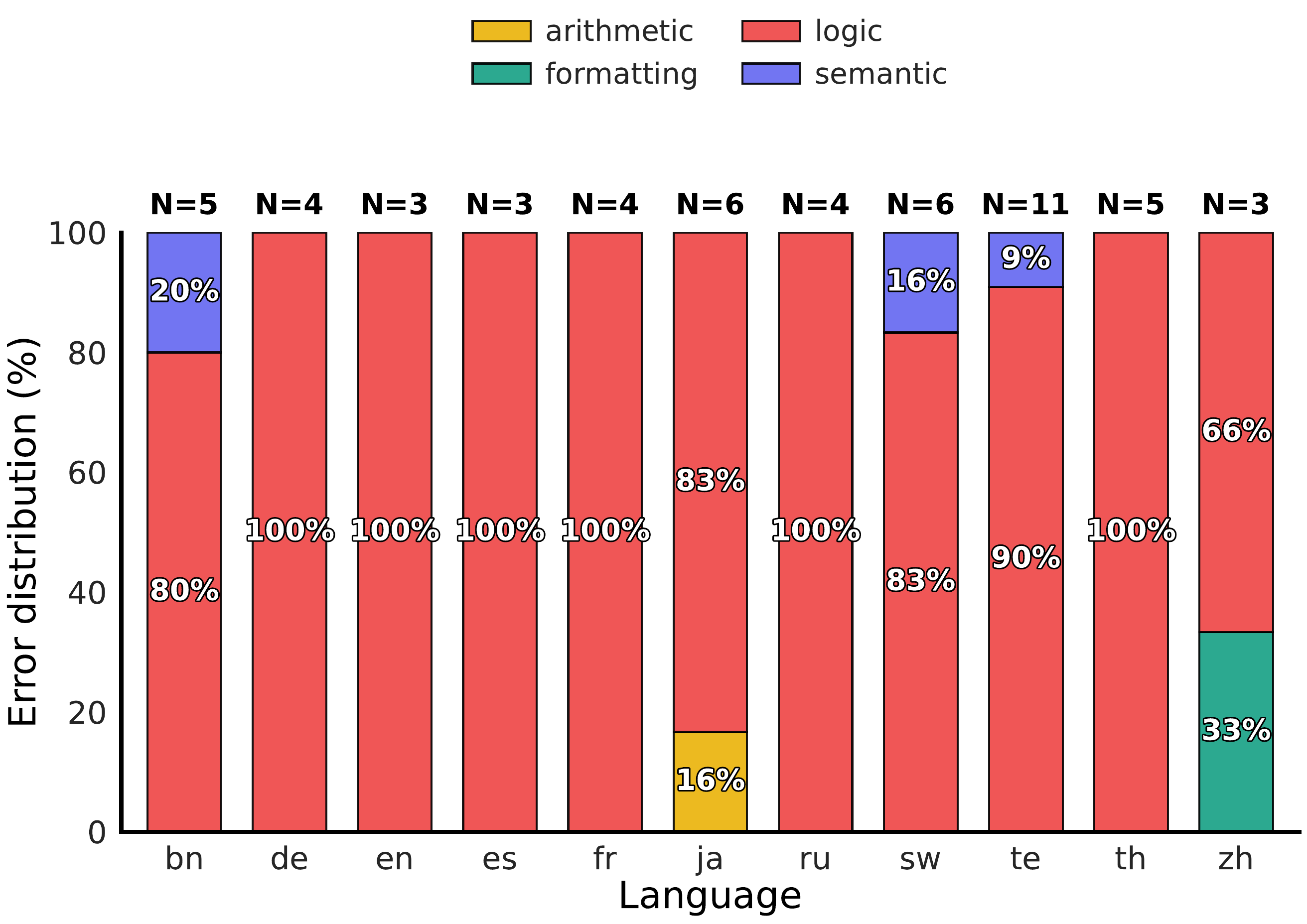}
        \caption{Error distribution for GPT-OSS-120B}
    \end{subfigure}
    \hfill
    \begin{subfigure}[b]{0.48\textwidth}
        \centering
        \includegraphics[width=\textwidth]{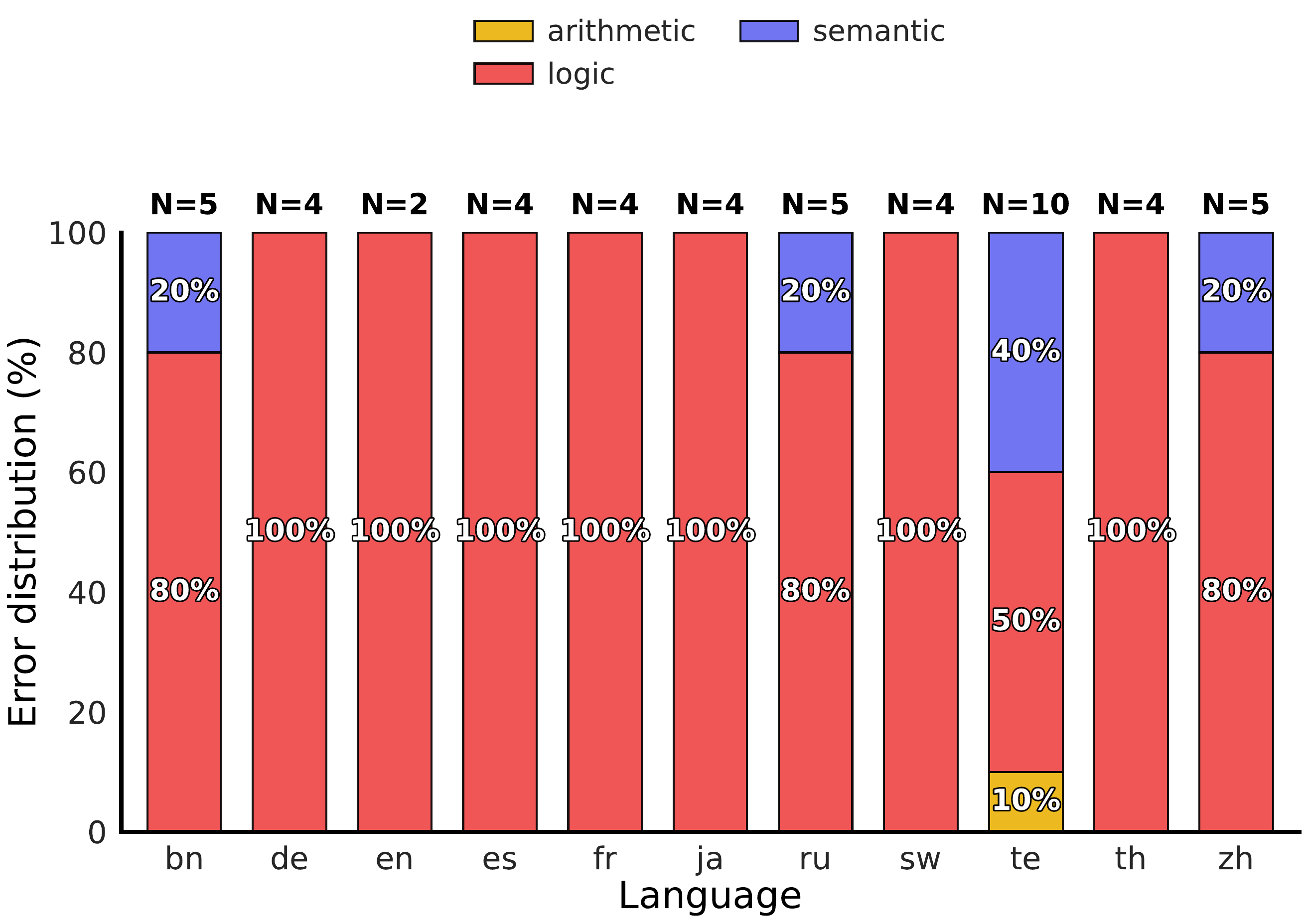}
        \caption{Error distribution for GLM 4.7}
    \end{subfigure}

    \caption{Error analysis across four additional models confirms MT-AIME24 errors are logical, not linguistic. We categorize errors for four models on MT-AIME24 across 11 languages. (a-d) Across all models, errors are predominantly logical (blue) or arithmetic (green) rather than semantic (red). The semantic error rate rarely exceeds 25\% for any language-model combination. For several languages (e.g., French in panels a and c), 100\% of errors stem from reasoning failures. This consistent pattern across diverse model architectures confirms that MT-AIME24 does not effectively measure multilingual comprehension.}
    \label{fig:aime24_err_dist}
\end{figure*}

\begin{figure*}[b]
    \centering
    \begin{subfigure}[b]{0.48\textwidth}
        \centering
        \includegraphics[width=\textwidth]{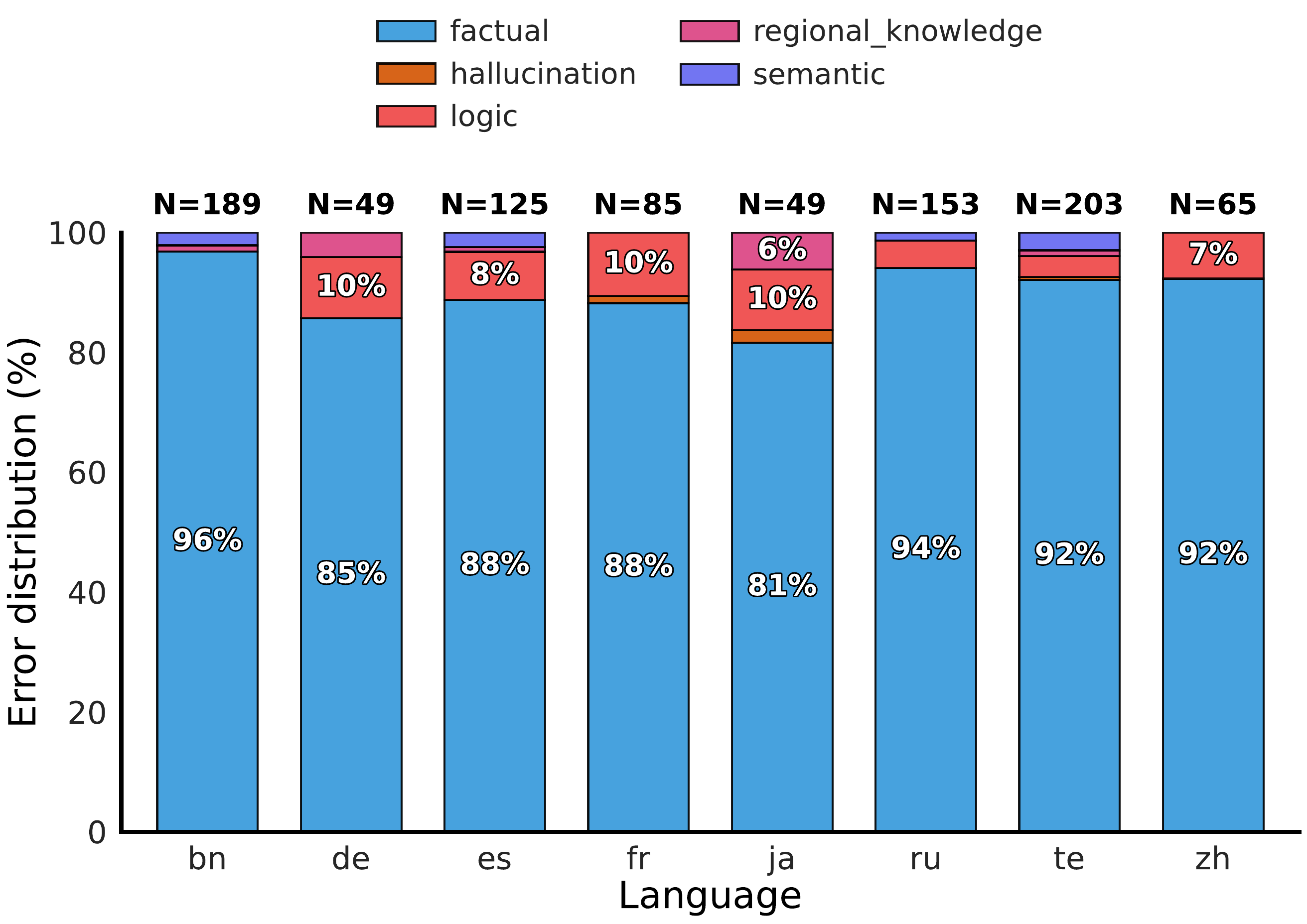}
        \caption{Qwen3-32B}
    \end{subfigure}
    \hfill
    \begin{subfigure}[b]{0.48\textwidth}
        \centering
        \includegraphics[width=\textwidth]{figures/err_dist/include44_error_distribution_qwen-qwen3-235b-a22b-thinking-2507_0-shot-judge=google-gemini-3-flash-preview.pdf}
        \caption{Qwen3-235B-A22B-Thinking}
    \end{subfigure}

    \vspace{0.5em} 

    \begin{subfigure}[b]{0.48\textwidth}
        \centering
        \includegraphics[width=\textwidth]{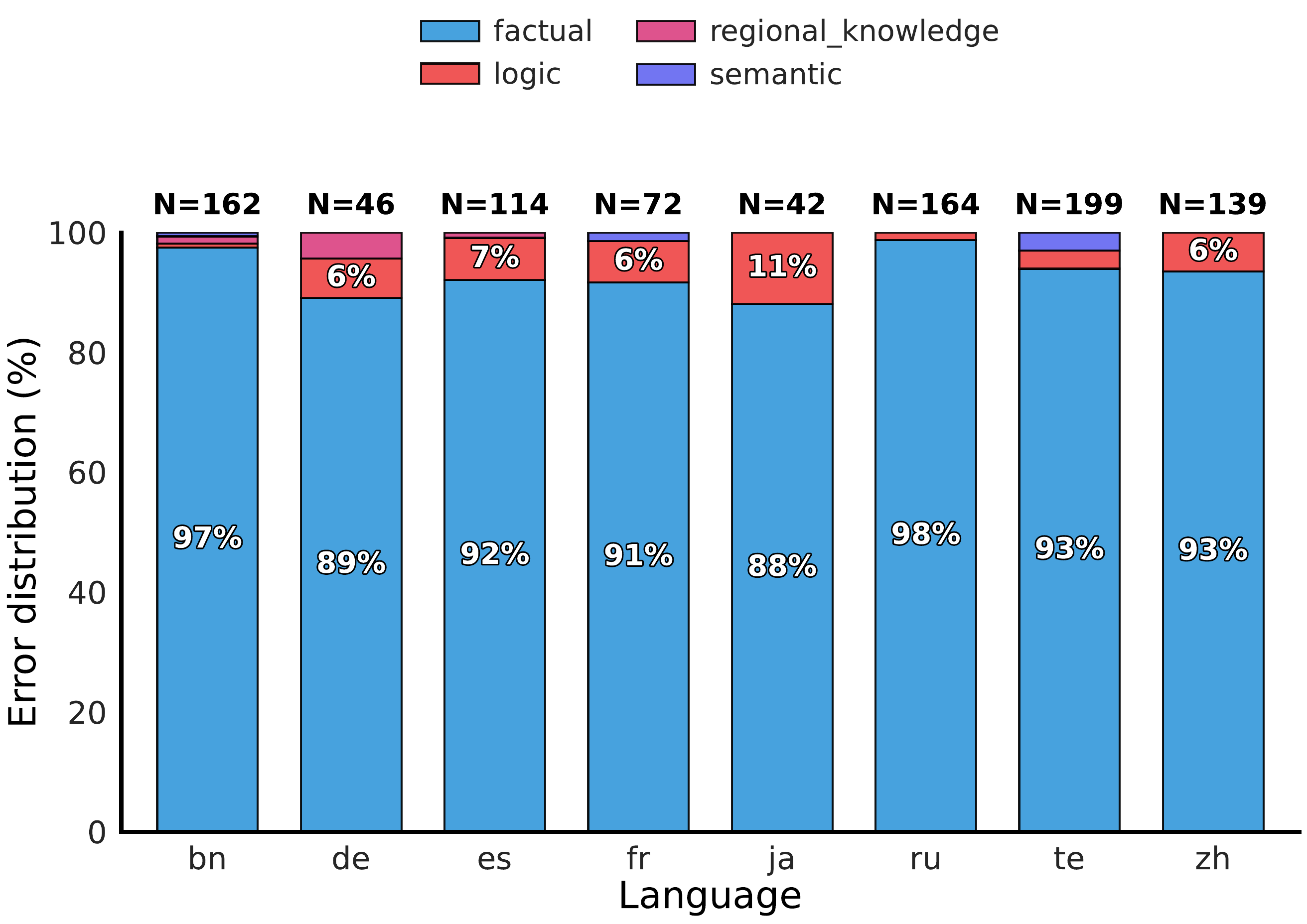}
        \caption{GPT-OSS-120B}
    \end{subfigure}
    \hfill
    \begin{subfigure}[b]{0.48\textwidth}
        \centering
        \includegraphics[width=\textwidth]{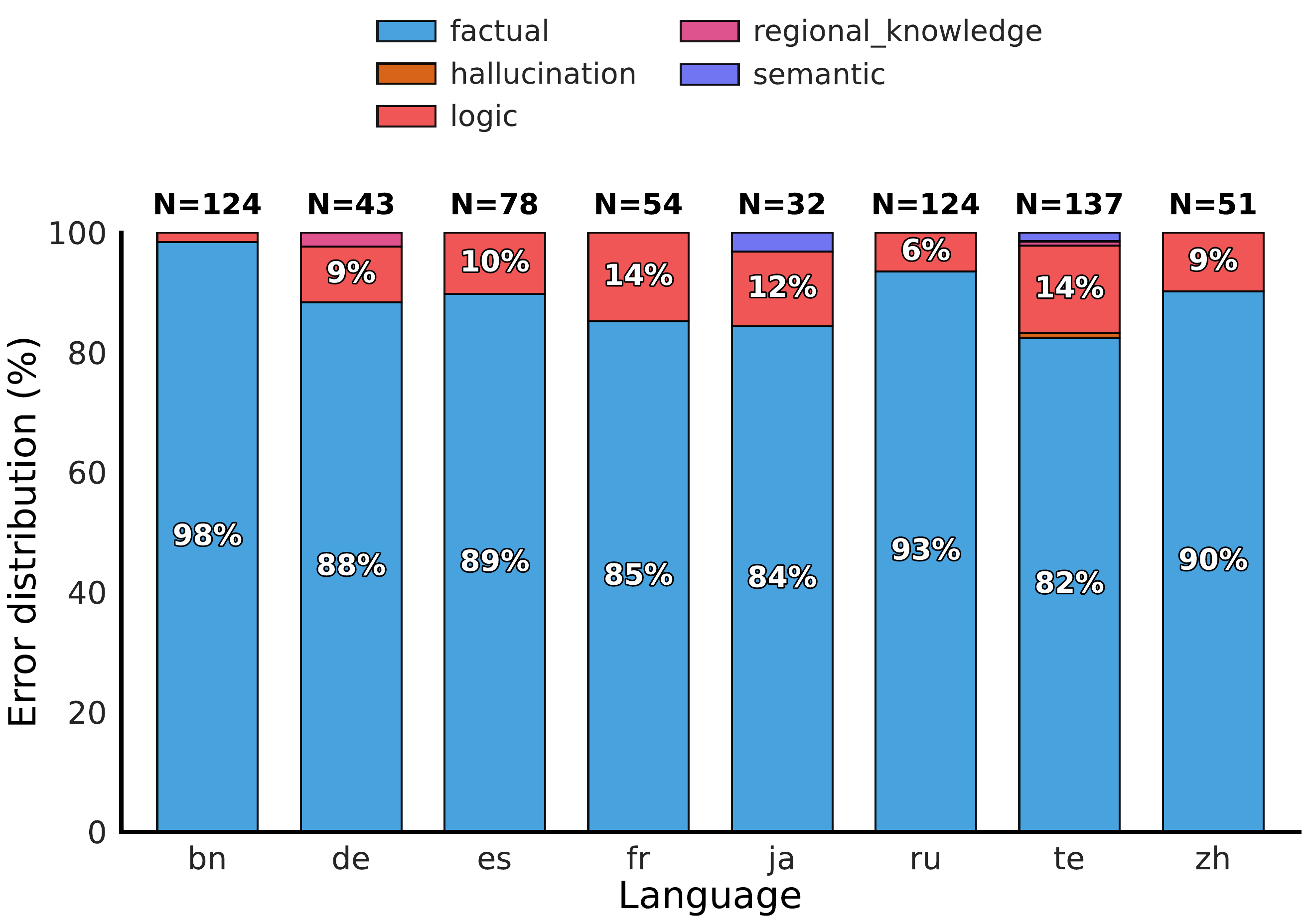}
        \caption{GLM 4.7}
    \end{subfigure}

    \caption{Error analysis on INCLUDE confirms errors are factual and knowledge-based, not linguistic. We categorize errors for four models across eight languages on INCLUDE. (a-d) Across all models, the dominant error type is factual (blue), accounting for 81–98\% of errors depending on language and model. Semantic errors (red) rarely exceed 10\% for any configuration. Regional knowledge gaps (orange) contribute modestly (6–14\% in some cases). Hallucinations appear occasionally but are not the primary failure mode. This consistent pattern confirms that INCLUDE measures factual knowledge coverage rather than multilingual comprehension ability.}
    \label{fig:include_err_dist}
\end{figure*}

We provide additional error distribution analyses for MT-AIME24 in Figure~\ref{fig:aime24_err_dist} and Include in Figure~\ref{fig:include_err_dist} across additional model families. We observe that MT-AIME24 errors are consistently dominated by logical and arithmetic failures rather than semantic misunderstanding, indicating that the benchmark primarily tests mathematical reasoning rather than multilingual comprehension. 

Figure~\ref{fig:include_err_dist} shows a parallel pattern for Include: errors are overwhelmingly factual, with smaller contributions from regional knowledge gaps and hallucinations, while semantic errors due to multilingual misunderstanding remain rare. Together, these results strengthen our conclusion that current multilingual benchmarks mainly reflect reasoning and factual recall, not genuine cross-lingual understanding.

\clearpage

\section{Extended Experimental Details} 
\label{app:additional_details}
\subsection{Dataset Composition}

We provide more details on the composition of our proposed LiT benchmark:

\textit{\textbf{(a) Abstracts}} (20\% samples): We divide this category equally between Humanities (10\%) and STEM (10\%) abstracts. Abstracts are useful benchmark units because they are self-contained, semantically dense, and complete, so errors are easier to detect \citep{isabelle-etal-2017-challenge, kleidermacher2025sciencelanguagesassessingllm}. Humanities abstracts test whether models can preserve argumentative nuance and rhetorical force, while STEM abstracts test terminological precision and mathematical notation, where minor mistakes carry major consequences.

\textit{\textbf{(b) Pragmatics}} (60\% samples): This category tests whether models preserve meaning beyond literal content \citep{park-etal-2024-multiprageval}. We subdivide it into five phenomena: \textit{(i) Core Semantics} (20\% samples): tests preservation of truth conditions and logical entailment \citep{rte_dagan}. \textit{(ii) Discourse Coherence} (17.5\% samples): evaluates maintenance of referential chains, topic continuity, and logical connectives across sentence boundaries \cite{10.3115/1219840.1219858}. \textit{(iii) Implicit Content} (17.5\% samples): probes handling of presuppositions, implicatures, and information that speakers convey without explicitly stating \citep{grice1975logic}. \textit{(iv) Pragmatic Inference} (21.7\% samples): tests understanding of speech acts, speaker intent, and context-dependent meaning \citep{searle1969speech}. \textit{(v) Social Interaction} (23.3\% samples): evaluates preservation of politeness markers, formality levels, and sociolinguistic appropriateness \citep{brown1987politeness}.

\textit{\textbf{(c) Informal}} (20\% samples): This category tests colloquial language, slang, idioms, and register shifts \citep{koehn-knowles-2017-six, fadaee-etal-2018-examining}. Informal text requires preserving tone and social function, not just denotative meaning.

\subsection{Sampling Hyperparameters}

\begin{table}[h]
    \centering
    \footnotesize
    \renewcommand{\arraystretch}{1.2}
    \caption{Hyperparameter details for round-trip translation.}
    \begin{tabular}{l|ccc}
        \toprule
        \textbf{Model} & \textbf{Temp.} & \textbf{Top-p} & \textbf{Effort} \\
        \midrule
        Gemini-3-Flash (No-Thinking) & 1.0 & 0.95 & -- \\
        GLM-4.7 & 1.0 & 0.95 & -- \\
        GLM-5 (Thinking) & 1.0 & 0.95 & -- \\
        GLM-5 (Instruct) & 1.0 & 0.95 & -- \\
        Qwen3-235B (Thinking) & 0.6 & 0.95 & -- \\
        Qwen3-235B (Instruct) & 0.7 & 0.8 & -- \\
        Qwen3-30B (Instruct) & 0.7 & 0.8 & -- \\
        Qwen3.5-397B (Thinking) & 0.6 & 0.95 & -- \\
        Qwen3.5-397B (Instruct) & 0.7 & 0.8 & -- \\
        Qwen3.5-35B (Thinking) & 0.6 & 0.95 & -- \\
        Qwen3.5-35B (Instruct) & 0.7 & 0.8 & -- \\
        Kimi-K2 (Thinking) & 1.0 & 1.0 & -- \\
        Kimi-K2 (Instruct) & 1.0 & 1.0 & -- \\
        DeepSeek-V3.2 (Thinking) & 1.0 & 0.95 & -- \\
        DeepSeek-V3.2 (Instruct) & 1.0 & 0.95 & -- \\
        GPT-OSS-120B & 1.0 & 1.0 & High \\
        MiniMax-M2.5 & 1.0 & 0.95 & -- \\
        MiMo-V2-Flash (Instruct) & 1.0 & 0.95 & -- \\
        Nemotron-3-Nano & 0.6 & 0.95 & -- \\
        Gemma-3-27B (Instruct) & 1.0 & 0.95 & -- \\
        Gemma-4-31B (Instruct) & 1.0 & 0.95 & -- \\
        Gemma-4-31B (Thinking) & 1.0 & 0.95 & -- \\
        \bottomrule
    \end{tabular}
    \label{tab:sampling_params_lit}
\end{table}

\begin{table}[h]
    \centering
    \small
    \renewcommand{\arraystretch}{1.2}
    \setlength{\tabcolsep}{4pt}
    \caption{Hyperparameter details for MT-AIME24 and INCLUDE}
    \begin{tabular}{l|cc|cc|c}
        \toprule
        \multirow{2}{*}{\textbf{Model}} & \multirow{2}{*}{\textbf{Temp.}} & \multirow{2}{*}{\textbf{Top-p}} & \multicolumn{2}{c|}{\textbf{Max Tokens}} & \multirow{2}{*}{\textbf{Effort}} \\
         & & & \textbf{AIME24} & \textbf{Include} & \\
        \midrule
        Gemini-3-Flash & 1.0 & 0.95 & 38,912 & 32,768 & -- \\
        GLM-4.7 & 1.0 & 0.95 & 38,912 & 32,768 & -- \\
        Qwen3-235B (Thinking) & 0.6 & 0.95 & 38,912 & 32,768 & -- \\
        Kimi-K2 (Thinking) & 1.0 & 1.0 & 131,072 & 32,768 & -- \\
        DeepSeek-V3.2-Exp (Think) & 1.0 & 0.95 & 38,912 & 32,768 & -- \\
        DeepSeek-V3.2-Exp & 1.0 & 0.95 & 38,912 & 32,768 & -- \\
        GPT-OSS-120B & 1.0 & 1.0 & 38,912 & 32,768 & High \\
        MiMo-V2-Flash & 1.0 & 0.95 & 38,912 & 32,768 & -- \\
        Nemotron-3-Nano & 0.6 & 0.95 & 38,912 & 32,768 & -- \\
        Gemma-3-27B (Instruct) & 1.0 & 0.95 & 38,912 & 32,768 & -- \\
        Qwen3-235B (Instruct) & 0.7 & 0.8 & 38,912 & 32,768 & -- \\
        Qwen3-30B (Instruct) & 0.7 & 0.8 & 38,912 & 32,768 & -- \\
        \bottomrule
    \end{tabular}
    \label{tab:sampling_params_aime_include}
\end{table}

The hyperparameters used for the model sampling utilizing Openrouter are presented in Tables~\ref{tab:sampling_params_lit} for the LiT benchmark and Table~\ref{tab:sampling_params_aime_include} for MT-AIME24 and Include. We follow the official technical reports for the corresponding models and use the official sampling parameters unless specified otherwise. If no suggested default is officially given, we set the temperature to a default of 1.0 and Top-p sampling to 0.95.

\subsection{Model Prompt}
We provide the translation prompt used in our evaluation, for reproducibility of our pipeline:

\begin{systemprompt}
You are a professional translator.

Task: Translate the SOURCE TEXT from \{src\_lang\} into \{target\_lang\}.

Instructions:
\begin{itemize}
    \setlength\itemsep{0em}
    \item Use natural, idiomatic \{target\_lang\}, avoid unnatural word-for-word translation.
    \item Preserve meaning, tone, and register. Do not add, omit, or summarize.
    \item Output the translated text in \{target\_lang\}. Do not include any additional text.
\end{itemize}

SOURCE TEXT: "\{text\}"
\end{systemprompt}

\subsection{Judge Prompt} 
We provide the judge prompt used in our  evaluation, for reproducibility of our pipeline:
\clearpage

\begin{systemprompt}
You are an annotator and expert linguist. Your task is to evaluate the back-translation given the original text and output a score, classification of machine translation quality, and a list of issues found.

Score measures back-translation quality on a continuous scale from 0 to 100, where a score of zero means "little-to-no meaning preserved" and score of one hundred means "perfect meaning preserved".

The classification of the quality of machine translation must be into one of following 10 categories:
\begin{itemize}
    \setlength\itemsep{0em}
    \item '1-nonsense': The text is gibberish, in the wrong language, or completely unrelated to the source.
    \item '2-severe distortion': Contains unrecognizable fragments. The core meaning is often lost or dangerously misleading.
    \item '3-failed gist': The topic is broadly correct, but the translation is mostly misleading or incomprehensible due to severe errors.
    \item '4-unreliable': Meaning is often preserved but significant meaning errors (critical mistranslations) are present.
    \item '5-machine-like': The meaning is roughly preserved (no critical errors), but the phrasing is overly literal ("translationese"), awkward, or grammatically poor.
    \item '6-understandable but Flawed': Meaning is preserved. Grammar is mostly functional but contains distracting errors or very unnatural stiffness.
    \item '7-good': Accurate meaning. Grammatically correct with only minor, non-impeding errors (e.g., wrong punctuation, slight awkwardness).
    \item '8-very good': Fluent and accurate. No grammatical errors. However, it may miss minor nuances of tone or style found in the Reference.
    \item '9-excellent': Native-level fluency. Captures the exact meaning and tone. Indistinguishable from professional human translation.
    \item '10-perfect': Flawless. Captures distinct cultural nuances, idioms, and subtext perfectly. Equivalent to the reference.
\end{itemize}

The list of issues is a comprehensive list of errors based on the following MQM Core dimensions: Accuracy, Fluency, and Terminology and Style/Locale.
\begin{enumerate}
    \item Accuracy: (Mistranslation, Omission, Addition, Untranslated). Does the target text accurately reflect the source meaning?
    \item Fluency: (Grammar, Spelling, Punctuation, Unintelligible). Is the target text linguistically correct and natural?
    \item Terminology: (Inconsistent, Wrong Term). Does it adhere to domain standards?
    \item Style/Locale: Does it follow local formats (dates, currencies) and cultural norms? Does the translation match the required formality/register (e.g., formal vs. casual)?
\end{enumerate}

The three severity categories are:
\begin{itemize}
    \item 'minor': Has a limited impact on accuracy, stylistic quality, consistency, fluency, clarity, or general appeal of the content.
    \item 'major': Seriously affects the understandability, reliability, or usability of the content for its intended purpose. For example, it causes significant loss or change in meaning or because the error appears in a highly visible or important part of text.
    \item 'critical': Hallucination, completely changes meaning or catastrophic failure rendering the sentence unusable or poses serious reputational harm.
\end{itemize}
Issues is a list of tuples of each issue being a tuple of (severity category, issue description).

Output Format: You must output a single valid JSON object. Do not include markdown formatting (like ```json) or conversational text. The JSON must follow this schema: 
\{"score": <0-100>, "classification": "<one of the 10 quality categories>", "issues": [\{"severity":"< one of three severity categories>", "description":"<issue description>"\} , ...]\}

Original text: "\{original\}"\\
Back-translation: "\{translation\}"
\end{systemprompt}

\end{document}